\documentclass{article}
\usepackage[preprint]{neurips_2020}

\usepackage[utf8]{inputenc} 
\usepackage[T1]{fontenc}    
\usepackage{hyperref}       
\usepackage{url}            
\usepackage{booktabs}       
\usepackage{amsfonts}       
\usepackage{nicefrac}       
\usepackage{microtype}      
\usepackage{xcolor}
\usepackage{graphicx}
\usepackage{subfigure}
\usepackage{tabulary}
\usepackage{wrapfig}
\usepackage[font=small]{caption}
\usepackage{amsbsy,amssymb,amsmath,amsthm}
\usepackage{paralist,physics}

\usepackage{tikz}
\usepackage{tkz-graph}
\tikzstyle{vertex}=[circle, draw, inner sep=0pt, minimum size=6pt]
\usetikzlibrary{arrows}

\newcommand{\vertexp}{\node[vertex,fill=green]}
\newcommand{\vertexn}{\node[vertex,fill=cyan]}
\newcommand{\vertexinput}{\node[vertex,fill=lightgray]}

\newtheorem{theorem}{Theorem}
\newtheorem{lemma}[theorem]{Lemma}

\newtheorem{condition}[theorem]{Condition}

\newcommand{\secref}[1]{Section~\ref{#1}}
\newcommand{\figref}[1]{Figure~\ref{#1}}

\newcommand{\thmref}[1]{Theorem~\ref{#1}}

\newcommand{\lemref}[1]{Lemma~\ref{#1}}

\newcommand{\condref}[1]{Condition~\ref{#1}}

\DeclareMathOperator{\arcsinh}{arcsinh}
\DeclareMathOperator{\diag}{diag}

\DeclareMathOperator*{\argmin}{argmin}

\newcommand{\x}{\mathbf{x}}
\newcommand{\w}{\mathbf{w}}
\renewcommand{\r}{\mathbf{r}}
\renewcommand{\u}{\mathbf{u}}
\newcommand{\bR}{\mathbb{R}}
\newcommand{\remove}[1]{{}}
\newcommand{\innerprod}[2]{{\langle #1,#2\rangle}}
\newcommand{\ie}{\textit{i.e.,}~}
\newcommand{\eg}{\textit{e.g.,}~}

\title{Implicit Bias in Deep Linear Classification: Initialization Scale vs Training Accuracy}

\author{
Edward Moroshko \\
\texttt{edward.moroshko@gmail.com} \\
 Technion
\And
Suriya Gunasekar \\
\texttt{suriya@ttic.edu} \\
 Microsoft Research
\AND
Blake Woodworth \\
\texttt{blake@ttic.edu} \\
 Toyota Technological Institute at Chicago
\And
Jason D. Lee \\
\texttt{jasonlee@princeton.edu} \\
 Princeton University
\AND
Nathan Srebro \\
\texttt{nati@ttic.edu} \\
 Toyota Technological Institute at Chicago
\And
Daniel Soudry \\
\texttt{daniel.soudry@gmail.com} \\
 Technion
}

\begin{document}

\maketitle

\begin{abstract}
We provide a detailed asymptotic study of gradient flow trajectories and their implicit optimization bias when minimizing the exponential loss over ``diagonal linear networks”. This is the simplest model displaying a transition between ``kernel'' and non-kernel (``rich'' or ``active'') regimes.  We show how the transition is controlled by the relationship between the initialization scale and how accurately we minimize the training loss.  Our results indicate that some limit behaviors of gradient descent only kick in at ridiculous training accuracies (well beyond $10^{-100}$). Moreover, the implicit bias at reasonable initialization scales and training accuracies is more complex and not captured by these limits.

\end{abstract}

\section{Introduction}

The optimization trajectory, and in particular the ``implicit bias'' determining which predictor the optimization algorithm leads to, plays a crucial role in learning with massively under-determined models, including deep networks, where many zero-error predictors are possible \citep[\eg][]{telgarsky2013margins,neyshabur2015search,zhang2017understanding,neyshabur2017implicit}.  Indeed, in several models we now understand how rich and natural implicit bias, often inducing sparsity of some form, can arise when training a multi-layer network with gradient descent \citep{telgarsky2013margins,gunasekar2017implicit,li2018algorithmic,gunasekar2018implicit,ji2019gradient,arora2019implicit,nacson2019lexicographic,lyu2020gradient}. This includes low $\ell_1$ norm \citep{woodworth2020kernel}, sparsity in the frequency domain \citep{gunasekar2018implicit}, low nuclear norm \citep{gunasekar2017implicit,li2018algorithmic}, low rank \citep{arora2019implicit,razin2020implicit}, and low higher order total variations \citep{chizat2020implicit}.  A different line of works focuses on how, in a certain regime, the optimization trajectory of neural networks, and hence also the implicit bias, stays near the initialization and mimics that of a kernelized linear predictor (with the kernel given by the tangent kernel) \citep{li2018learning,du2018gradient,jacot2018neural,chizat2019lazy,allen2019convergence,du2019gradient,zou2018stochastic,allen2019learning,arora2019fine,cao2019generalization}.  In such a ``kernel regime'' the implicit bias corresponds to minimizing the norm in some Reproducing Kernel Hilbert Space (RKHS), and cannot yield the rich sparsity-inducing\footnote{Sparsity in an implicit space can also be understood as feature search or ``adaptation''.  E.g.~eigenvalue sparsity is equivalent to finding features which are linear combinations of input features, and sparsity in the space of ReLU units \cite[\eg][]{savarese2019infinite} corresponds to finding new non-linear features.} inductive biases discussed above, and is perhaps not as ``adaptive'' as one might hope.  It is therefore important to understand when and how learning is in the kernel regime, what hyper-parameters (\eg initialization, width, etc.) control the transition out and away from the kernel regime, and how the implicit bias (and thus the inductive bias driving learning) changes in different regimes and as a function of different hyper-parameters.

Initial work identified the {\bf width} as a relevant hyper-parameter, where the kernel regime is reached when the width grows towards infinity \citep{li2018learning,du2018gradient,jacot2018neural,allen2019convergence,du2019gradient,zou2018stochastic}.  But subsequent work by \citet{chizat2019lazy} pointed to the {\bf initialization scale} as the relevant hyper-parameter, showing that models of any width enter the kernel regime as the scale of initialization tends towards infinity.  Follow-up work by \citet{woodworth2020kernel} studied this transition in detail for regression with  ``diagonal linear networks'' (see \secref{sec:diagnet}), showing how it is controlled by the interactions of {\bf width}, {\bf scale of initialization} and also {\bf depth}.  They obtained an exact expression for the implicit bias in terms of these hyper-parameters, showing how it transitions from $\ell_2$ (kernel) implicit bias when the width or initialization go to infinity, to $\ell_1$ (``rich'' or ``active'' regime) for infinitesimal initialization, showing how a width of $k$ has an equivalent effect to increasing initialization scale by a factor of $\sqrt{k}$.  Studying the effect of initialization scale can therefore also be understood as studying the effect of width.

\begin{figure}
\begin{minipage}[l]{0.6\textwidth}
\centering
\begin{tabular}{|c|c|c|}
\hline
   &   \small $\alpha < \infty$ &   $\small \alpha\rightarrow \infty$ \\
\hline
    $\small \epsilon > 0$ &  &   \small Kernel \citep{chizat2019lazy}\\
\hline
    $\small \epsilon\rightarrow 0$ &   \small Rich \citep{lyu2020gradient} & \small  \textbf{This Work} \\
\hline
\end{tabular}
\end{minipage}
\begin{minipage}[r]{0.4\textwidth}
\begin{tikzpicture}[scale=1]
\draw[->,ultra thick] (0,0)--(2.5,0);
\draw[->,ultra thick] (0,0)--(0,2.5);
\node at (1.25,-0.3) {\footnotesize{Initialization $\alpha$}};
\node[rotate=90] at (-0.4,1.25) {\footnotesize{Accuracy $\log(1/\epsilon)$}};
\node[align=right] at (2,0.73) {  Kernel};
\node[align=right] at (2,0.36) {  Regime};
\node[align=left] at (0.7,2.43) {  Rich};
\node[align=left] at (0.7,2.06) {  Regime};
\node[align=center] at (2,2) {\huge ?};
\end{tikzpicture}
\centering
\end{minipage}
\caption{Kernel and rich limits.}
\label{fig:phase_diagram}
\end{figure}
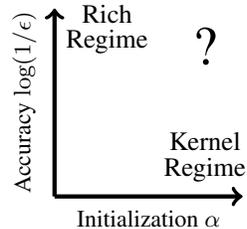

In this paper, we explore the transition and implicit bias for classification (as opposed to regression) problems, and highlight the effect of another hyper-parameter: {\bf training accuracy}.  A core distinction  is that in classification with an exponentially decaying loss (such as the cross-entropy/logistic loss), the ``rich regime'' (\eg $\ell_1$) implicit bias is attained with {\em any finite initialization}, and not only with infinitesimal initialization as in regression \citep{nacson2019lexicographic,lyu2020gradient}, and the scale of initialization does {\em not} effect the asymptotic implicit bias \citep{telgarsky2013margins,soudry2018implicit,ji2019implicit,gunasekar2018characterizing}. This is in seeming contradiction to the fact that even for classification problems, the kernel regime {\em is} attained when the scale of initialization grows \citep{chizat2019lazy}.  How can one reconcile the implicit bias not depending on the scale of initialization, with the kernel regime (and thus RKHS implicit bias) being reached as a function of initialization scale?  The answer is discussed in detail in \secref{sec:kernel-rich-background} and depicted in \figref{fig:phase_diagram}:  with the decaying losses used in classification, we can never get the loss to zero, only drive it arbitrarily close to zero.  If we train infinitely long, then indeed we will eventually reach ``rich'' bias even for arbitrarily large initialization.  On the other hand, if we only consider optimizing to within some arbitrarily high accuracy (\eg to loss $10^{-10}$), we will find ourselves in the kernel regime as the initialization scale grows.  But how do these hyper-parameters interact?  Where does this transition happen?  How accurately do we need to optimize to exit the ``kernel'' regime?  To get the ``rich'' implicit bias?  And what happens in between?  What happens if both initialization scale and training accuracy go to infinity? And how is this affected by the depth?

To answer these questions quantitatively, we consider a concrete and simple model, and study classification using diagonal linear networks.   This is arguably the simplest model that displays such transitions\footnote{Our guiding principal is that when studying a new or not-yet-understood phenomena, we should first carefully and fully understand it in the simplest model that shows it, so as not to get distracted by possible confounders, and to enable a detailed analytic understanding.}, and as such has already been used to understand non-RKHS implict bias and dynamics \citep{woodworth2020kernel,gissin2020}.

We consider minimizing the exp-loss of a $D$-layer diagonal linear network on separable data, starting from initialization at scale $\alpha$ and optimizing until the training loss reaches a value of $\epsilon$.  In this case, the two extreme regimes depicted in \figref{fig:phase_diagram} correspond to implicit biases given by the $\ell_2$ norm in the kernel regime and by the $\ell_{2/D}$ quasi-norm when $\epsilon\rightarrow 0$ for fixed $\alpha$.  We consider $\alpha\rightarrow \infty$ and optimizing to within $\epsilon(\alpha)\rightarrow 0$ and ask what happens for different joint behaviours of $\alpha$ and $\epsilon$.
\begin{compactitem}
\item We identify $\epsilon(\alpha)=\exp(-\Theta(\alpha^D))$ as the boundary of the kernel regime.  When $\epsilon(\alpha)=\exp(-o\left(\alpha^D\right))$, the optimization trajectory follows that of a kernel machine, and the implicit bias is given by the $\ell_2$ norm.  But when $\epsilon(\alpha)=\exp(-\Omega\left(\alpha^D\right))$, the trajectory deviates from this kernel behaviour.
\item Under additional condition (concerning stability of the support vectors), we characterize the behaviour at the transition regime, when $\epsilon(\alpha)=\exp(-\Theta(\alpha^D))$, and show that at this scaling the implicit bias is given by \citeauthor{woodworth2020kernel}'s $Q^D$ regularizer, which interpolates between the $\ell_1$-norm and the $\ell_2$-norm.  This indicates roughly the following optimization trajectory for large enough initialization scale and under the support vector stability condition: we will first pass through the minimum $Q^D_\infty=\ell_2$-norm predictor (or more accurately, max margin w.r.t.~the $\ell_2$-norm), then traverse the path of minimizers of $Q^D_\mu$, for $\mu\in(\infty,0)$, until we reach the minimum $Q^D_0=\ell_1$-norm predictor.  We confirm such behaviour in simulations, as well as deviations from it when the condition does not hold.
\item As suggested by our asymptotic theory, simulations in \secref{sec:simulations} show that even at moderate initialization scales, extreme training accuracy is needed in order to reach the asymptotic behaviour where the implicit bias is well understood.  Rather, we see that even in tiny problems, at moderate and reasonable initialization scales and training accuracies, the implicit bias behaves rather different from both the ``kernel'' and ``rich'' extremes, suggesting that further work is necessary to understand the effective implicit bias in realistic settings.
\end{compactitem}

\paragraph{Notation} For vectors $\mathbf{z}, \mathbf{v}$, we denote by $\mathbf{z}^{k}$, $\exp(\mathbf{z})$, and $\log(\mathbf{z})$ the element-wise $k$th power, exponential, and natural logarithm, respectively;  $\mathbf{z}\circ\mathbf{v}$ denotes element-wise multiplication;  and $\mathbf{z}\propto\mathbf{v}$ implies that $\mathbf{z}=\gamma\mathbf{v}$ for a positive scalar $\gamma$. $\mathbf{1}$ denotes the vector of all ones.

\section{Kernel and rich regimes in classification}\label{sec:kernel-rich-background}

We consider models as mappings $f:\bR^p\times\mathcal{X}\to\bR$ from trainable parameters $\u\in\bR^p$ and input $\x\in\mathcal{X}$ to predictions. We denote by $F(\u):\x\mapsto f(\u,\x)$ the function implemented by the parameters $\u$.  We will focus on models that are $D$-homogeneous in $\u$ for some $D>0$, \ie such that $\forall c>0, F(c\u)=c^DF(\u)$.  This includes depth-$D$ linear and ReLU networks.

We consider minimizing $\mathcal{L}(\u)=\frac{1}{N}\sum_{n=1}^N\ell(f(\u,\x_n),y_n)$ for a given dataset $\{(\x_n,y_n): n=1,2,\ldots N\}$ where $\ell:\bR\times\mathcal{Y}\rightarrow\bR$ is a loss function.  We will be mostly focus on binary classification problems, where $y_n \in \{-1,1\}$, and with the exp-loss $\ell(\hat{y},y)=\exp(-\hat{y}y)$, which has the same tail behaviour and thus similar asymptotic properties as the logistic or cross-entropy loss \citep[\eg][]{telgarsky2013margins,soudry2018implicit,lyu2020gradient}.  All our results and discussion refer to the exp-loss unless explicitly stated otherwise.  We are concerned with understanding the trajectory of gradient descent, which we consider at the limit of infintesimal stepsize, yielding the gradient flow dynamics,
\begin{equation}
    \dot{\u}(t)=-\nabla\mathcal{L}(\u(t))~, 
\label{eq:gf}
\end{equation}
where here and throughout $\dot{\u}=\dv{\u}{t}$.

Along the gradient flow path, $\mathcal{L}(\u(t))$ is monotonically decreasing and we consider cases where the loss is indeed minimized, \ie converges to $0$ for $t\rightarrow\infty$. However, if we stop the optimization trajectory at a large but finite $t$, which is what we do in practice, we optimize to some positive training loss $\epsilon(t)=\mathcal{L}(\u(t))$.
We define $\tilde{\gamma}(t)=-\log(\epsilon(t))$ as the {\em training accuracy}. $\tilde{\gamma}(t)$ can also be interpreted as the number of digits of the precision representing the training loss.
This is related to the prediction 
margin $\gamma(t)=\min_n y_n\x_n^\top \w(t)$ as $\gamma(t)\le \tilde{\gamma}(t)\le \gamma(t)+\log(N)$ and was introduced as \textit{smoothed margin} in \citet{lyu2020gradient}.

For classification problems, we consider {\bf separable data}, \ie $\exists \u_{\ast}\in\mathbb{R}^{p} \;:\forall_n\,y_{n} f(\u_{\ast},\x_n)>0$, and so $\mathcal{L}(\gamma \u_\ast)\stackrel{\gamma\longrightarrow\infty}{\longrightarrow}0$.  But especially in high dimensions, there are many such separating predictors.  If $\mathcal{L}(\u(t))\rightarrow 0$, which of these does $\u(t)$ converge to?  Of course $\u(t)$ does not converge, since to approach zero error it must {\em diverge}.  Therefore, instead of the limit of the parameters, we study the limit of the \textit{decision boundary} of the resulting classifier, which is given by $F\left(\frac{\u(t)}{\|\u(t)\|}\right)$.

\paragraph{Kernel Regime}  When the gradients $\nabla_\u f(\u,\x_n)$ do not change much during optimization, then $\u(t)$ behaves as if optimizing over a linearized model $\u$: $\bar{f}(\u,\x) = f(\u(0),\x)+\innerprod{\u}{\phi(\x)}$ where $\phi(\x)=\nabla_\u f(\u(0),\x)$ is the feature map corresponding to the Tangent Kernel at initialization $K(\x,\x')=\innerprod{\phi(\x)}{\phi(\x')}$ \citep{jacot2018neural,allen2019convergence,du2019gradient}.  Consider the trajectory $\bar{\u}(t)$ of gradient flow $\dot{\bar{\u}}(t)=-\nabla\bar{\mathcal{L}}(\bar{\u}(t))$ on the loss of this linearized  model $\bar{\mathcal{L}}(\u)=\frac{1}{N}\sum_{n=1}^N \ell(\bar{f}(\u,\x_n),y_n)$. \citet{chizat2019lazy} showed that any $D$-homogeneous model enters the kernel regime (\ie behaves like a linearized model) when the scale of the initialization is large:

\begin{theorem} [Adapted from Theorem 2.2 in \citet{chizat2019lazy}] \label{thm:kernel-finite-t}
For any fixed time horizon $T>0$, and any $\u_0$ such that $F(\u_0)=0$, and for the exp-loss, consider the two gradient flow trajectories $\u(t)$ and $\bar{\u}(t)$, respectively, both initialized with $\u(0)=\bar{\u}(0)=\alpha \u_0$, for $\alpha>0$.  Then $\lim_{\alpha\rightarrow\infty} \sup_{t\in[0,T]} \left\lVert F(\u(t))-\bar{F}(\bar{\u}(t))\right\rVert=0$.
\end{theorem}
For a linear model like $\bar{f}$, the gradient flow $\bar{\u}$ converges in direction to the maximum margin solution in the corresponding RKHS norm \citep{soudry2018implicit}. Combining this with \thmref{thm:kernel-finite-t}, we have
\begin{equation}
    \lim_{t\to\infty}\lim_{\alpha\to\infty}F\left(\frac{\u(t)}{\|\u(t)\|}\right)\propto\argmin_{f:\mathcal{X}\to\bR} \|f\|_K\;\text{s.t. }y_nf(\x_n)\ge 1,
\label{eq:kernel-1}
\end{equation}
where recall that $K$ is the Tangent Kernel at the initialization and $\norm{f}_K$ is the RKHS norm with respect to this kernel. 
It is important to highlight the crucial difference here compared to the corresponding statement for the squared loss \citep[Theorem 2.3]{chizat2019lazy}.  For the squared loss we have that $\lim_{\alpha\rightarrow\infty} \sup_{t\in[0,\infty)} \left\lVert F(\u(t))-\bar{F}(\bar{\u}(t))\right\rVert=0$, i.e.~the entire optimization trajectory converges uniformly to $\bar{\u}(t)$.  But for the exp-loss, Theorem \ref{thm:kernel-finite-t} only ensures convergence for prefixes of the path, up to finite time horizons $T$.  The order of limits in \eqref{eq:kernel-1} is thus crucial, while for the square loss the order of limits can be reversed.

\paragraph{Rich regime} On the other hand, for or any finite initialization $\alpha \u_0$, the limit direction of gradient flow, when optimized indefinitely, gives rise to a different limit solution \citep{gunasekar2018characterizing,nacson2019lexicographic,lyu2020gradient}:
\begin{theorem}[Paraphrasing Theorem 4.4. in \citet{lyu2020gradient}]\label{thm:rich_limit} Assume that the gradient flow trajectory in \eqref{eq:gf} minimizes the loss, i.e., $\mathcal{L}(\u(t))\to 0$. 
Then, any limit point of $\left\{\frac{\u(t)}{\|\u(t)\|}:t>0\right\}$ is along the direction of a KKT point of the following constrained optimization problem:
\begin{equation}
\min_\u \|\u\|_2\;\;\text{ s.t. }y_n f(\u,\x_n)\ge 1.
\label{eq:l2param}
\end{equation}
\end{theorem}

Compared to the kernel regime in \eqref{eq:kernel-1},  \thmref{thm:rich_limit} suggests\footnote{Theorem~\ref{thm:rich_limit} is suggestive of $\mathcal{R}(f)$ in \eqref{eq:induced-l2param} as the implicit induced bias in rich regime. However, although global minimizers of \eqref{eq:l2param} and the RHS of \eqref{eq:induced-l2param} are equivalent, the same is not the case for stationary points. For the special cases of certain linear networks and the infinite width univariate ReLU network, stronger results for convergence in direction to the KKT points of \eqref{eq:induced-l2param} can be shown  \citep{gunasekar2018implicit,ji2019gradient,chizat2020implicit}.}  that
\begin{equation}
\begin{split}
\lim\limits_{\alpha\to\infty}\lim\limits_{t\to\infty}F\left(\frac{\u(t)}{\|\u(t)\|}\right) \propto \text{ stationary points of }&\min_{f:\mathcal{X}\to\bR} \mathcal{R}(f)\;\;\text{ s.t. }y_n f(\x_n)\ge 1,\\
&\text{ where }\mathcal{R}(f)=\min_\u \|\u\|_2\;\text{ s.t. }F(\u)=f.
\end{split}
\label{eq:induced-l2param}
\end{equation}
To understand this double limit, note that Theorem \ref{thm:rich_limit} ensures convergence for every $\alpha$ separately, and so also as we take $\alpha\rightarrow\infty$.  For neural networks including linear networks, $\mathcal{R}(f)$ captures rich and often sparsity inducing inductive biases (\eg nuclear norm, higher order total variations, $\ell_p$ bridge penalty for $p\le1$) that are not captured by RKHS norms \citep{gunasekar2017implicit,ji2019implicit,savarese2019infinite,ongie2019function,gunasekar2018implicit}.

Contrasting \eqref{eq:kernel-1} and \eqref{eq:induced-l2param} we see that if both the initialization scale $\alpha$ and the optimization time $t$ go to infinity, the order in which we take the limits is crucial in determining the implicit bias, matching the depiction in Figure \ref{fig:phase_diagram}.  Roughly speaking, if $\alpha\to\infty$ first (\ie faster) and then $t\to\infty$, we end up in the kernel regime, but if $t\rightarrow\infty$ first (\ie faster), we can end up with rich implicit bias corresponding to $\mathcal{R}$. The main question we ask is: \textit{where the transition from kernel and rich regime happens, and what is the implicit bias when $\alpha\to\infty$ and $t\rightarrow\infty$ together?}

Since the time $t$ for gradient flow does not directly correspond to actual ``runtime'', and is perhaps less directly meaningful, we instead consider the optimization trajectory in terms of the training loss $\epsilon(t)=\mathcal{L}(\u(t))$. For some loss tolerance parameter $\epsilon$, we  follow the gradient flow trajectory until time $t$ such that $\mathcal{L}(t)=\epsilon$ and ask \textit{what is the implicit bias when $\alpha\to\infty$ and $\epsilon\rightarrow 0$ together?}

\section{Diagonal linear network of depth D} \label{sec:diagnet}
In the remainder of the paper we focus on depth-$D$ {\em diagonal linear networks}.  This is a $D$-homogeneous model with parameters $\u=\left[\begin{array}{c}
\u_{+}\\ \u_{-} \end{array}\right]\in\mathbb{R}^{2d}$ specified by:
\begin{equation}\label{eq:diagNN}
f(\u,\x) = \innerprod{\u_{+}^{D}-\u_{-}^{D}}{\x}
\end{equation}
where recall that the exponentiation is element-wise.

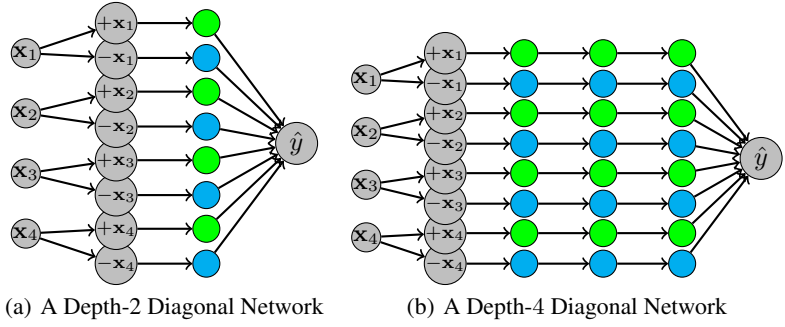
\begin{figure}
\centering
\subfigure[A Depth-2 Diagonal Network]{
    \begin{tikzpicture}[scale=1.6]
        \vertexinput(p004) at (-0.75,0.25) {\small $\x_4$};
        \vertexinput(p005) at (-0.75,0.75) {\small $\x_3$};
        \vertexinput(p006) at (-0.75,1.25) {\small $\x_2$};
        \vertexinput(p007) at (-0.75,1.75) {\small $\x_1$};
        \vertexinput(pp04) at (0,0) {\scriptsize $-\x_4$};
        \vertexinput(pp05) at (0,0.29) {\scriptsize $+\x_4$};
        \vertexinput(pp06) at (0,0.57) {\scriptsize $-\x_3$};
        \vertexinput(pp07) at (0,0.86) {\scriptsize $+\x_3$};
        \vertexinput(pn04) at (0,1.14) {\scriptsize $-\x_2$};
        \vertexinput(pn05) at (0,1.43) {\scriptsize $+\x_2$};
        \vertexinput(pn06) at (0,1.71) {\scriptsize $-\x_1$};
        \vertexinput(pn07) at (0,2) {\scriptsize $+\x_1$};

        \vertexn(p4) at (0.75,0) {$\phantom{+}$};
        \vertexn(p5) at (0.75,0.57) {$\phantom{+}$};
        \vertexn(p6) at (0.75,1.14) {$\phantom{+}$};
        \vertexn(p7) at (0.75,1.71) {$\phantom{+}$};
        \vertexp(p41) at (0.75,0.29) {$\phantom{+}$};
        \vertexp(p51) at (0.75,0.86) {$\phantom{+}$};
        \vertexp(p61) at (0.75,1.43) {$\phantom{+}$};
        \vertexp(p71) at (0.75,2) {$\phantom{+}$};

        \vertexinput(p15) at (1.5,1) {\rule[-5pt]{0pt}{15pt}$\hat{y}$};
    \tikzset{EdgeStyle/.style={->}}
        \Edge(p004)(pp04)
        \Edge(p004)(pp05)
        \Edge(p005)(pp06)
        \Edge(p005)(pp07)
        \Edge(p006)(pn04)
        \Edge(p006)(pn05)
        \Edge(p007)(pn06)
        \Edge(p007)(pn07)
        \Edge(pp04)(p4)
        \Edge(pp05)(p41)
        \Edge(pp06)(p5)
        \Edge(pp07)(p51)
        \Edge(pn04)(p6)
        \Edge(pn05)(p61)
        \Edge(pn06)(p7)
        \Edge(pn07)(p71)

        \Edge(p4)(p15)
        \Edge(p5)(p15)
        \Edge(p6)(p15)
        \Edge(p7)(p15)
        \Edge(p41)(p15)
        \Edge(p51)(p15)
        \Edge(p61)(p15)
        \Edge(p71)(p15)
    \end{tikzpicture}
}
~
\subfigure[A Depth-$4$ Diagonal Network]{
    \begin{tikzpicture}[scale=1.4]
        \vertexinput(p004) at (-0.75,0.25) {\small $\x_4$};
        \vertexinput(p005) at (-0.75,0.75) {\small $\x_3$};
        \vertexinput(p006) at (-0.75,1.25) {\small $\x_2$};
        \vertexinput(p007) at (-0.75,1.75) {\small $\x_1$};
        \vertexinput(pp04) at (0,0) {\scriptsize $-\x_4$};
        \vertexinput(pp05) at (0,0.29) {\scriptsize $+\x_4$};
        \vertexinput(pp06) at (0,0.57) {\scriptsize $-\x_3$};
        \vertexinput(pp07) at (0,0.86) {\scriptsize $+\x_3$};
        \vertexinput(pn04) at (0,1.14) {\scriptsize $-\x_2$};
        \vertexinput(pn05) at (0,1.43) {\scriptsize $+\x_2$};
        \vertexinput(pn06) at (0,1.71) {\scriptsize $-\x_1$};
        \vertexinput(pn07) at (0,2) {\scriptsize $+\x_1$};

        \vertexn(p4) at (0.75,0) {$\phantom{+}$};
        \vertexn(p5) at (0.75,0.57) {$\phantom{+}$};
        \vertexn(p6) at (0.75,1.14) {$\phantom{+}$};
        \vertexn(p7) at (0.75,1.71) {$\phantom{+}$};
        \vertexp(p41) at (0.75,0.29) {$\phantom{+}$};
        \vertexp(p51) at (0.75,0.86) {$\phantom{+}$};
        \vertexp(p61) at (0.75,1.43) {$\phantom{+}$};
        \vertexp(p71) at (0.75,2) {$\phantom{+}$};
        \vertexn(p4a) at (1.5,0) {$\phantom{+}$};
        \vertexn(p5a) at (1.5,0.57) {$\phantom{+}$};
        \vertexn(p6a) at (1.5,1.14) {$\phantom{+}$};
        \vertexn(p7a) at (1.5,1.71) {$\phantom{+}$};
        \vertexp(p41a) at (1.5,0.29) {$\phantom{+}$};
        \vertexp(p51a) at (1.5,0.86) {$\phantom{+}$};
        \vertexp(p61a) at (1.5,1.43) {$\phantom{+}$};
        \vertexp(p71a) at (1.5,2) {$\phantom{+}$};
        \vertexn(p4b) at (2.25,0) {$\phantom{+}$};
        \vertexn(p5b) at (2.25,0.57) {$\phantom{+}$};
        \vertexn(p6b) at (2.25,1.14) {$\phantom{+}$};
        \vertexn(p7b) at (2.25,1.71) {$\phantom{+}$};
        \vertexp(p41b) at (2.25,0.29) {$\phantom{+}$};
        \vertexp(p51b) at (2.25,0.86) {$\phantom{+}$};
        \vertexp(p61b) at (2.25,1.43) {$\phantom{+}$};
        \vertexp(p71b) at (2.25,2) {$\phantom{+}$};

        \vertexinput(p15) at (3,1) {\rule[-5pt]{0pt}{15pt}$\hat{y}$};
    \tikzset{EdgeStyle/.style={->}}
        \Edge(p004)(pp04)
        \Edge(p004)(pp05)
        \Edge(p005)(pp06)
        \Edge(p005)(pp07)
        \Edge(p006)(pn04)
        \Edge(p006)(pn05)
        \Edge(p007)(pn06)
        \Edge(p007)(pn07)
        \Edge(pp04)(p4)
        \Edge(pp05)(p41)
        \Edge(pp06)(p5)
        \Edge(pp07)(p51)
        \Edge(pn04)(p6)
        \Edge(pn05)(p61)
        \Edge(pn06)(p7)
        \Edge(pn07)(p71)

        \Edge(p4)(p4a)
        \Edge(p5)(p5a)
        \Edge(p6)(p6a)
        \Edge(p7)(p7a)
        \Edge(p41)(p41a)
        \Edge(p51)(p51a)
        \Edge(p61)(p61a)
        \Edge(p71)(p71a)
        \Edge(p4a)(p4b)
        \Edge(p5a)(p5b)
        \Edge(p6a)(p6b)
        \Edge(p7a)(p7b)
        \Edge(p41a)(p41b)
        \Edge(p51a)(p51b)
        \Edge(p61a)(p61b)
        \Edge(p71a)(p71b)
        \Edge(p4b)(p15)
        \Edge(p5b)(p15)
        \Edge(p6b)(p15)
        \Edge(p7b)(p15)
        \Edge(p41b)(p15)
        \Edge(p51b)(p15)
        \Edge(p61b)(p15)
        \Edge(p71b)(p15)
    \end{tikzpicture}
}
\vspace{-8pt}
\small \caption{\small Diagonal linear networks.
}\vspace{-8pt}
\label{fig:diagonal_networks}
\end{figure}

As depicted in Figure \ref{fig:diagonal_networks}, the model can be though of as a depth-$D$  network, with $D-1$ hidden linear layers (i.e.~the output is a weighted sum of the inputs), each consisting of $2d$ units, with the first hidden layer connected to the $d$ inputs and their negations (depicted in the figure as another fixed layer), each unit in subsequent hidden layers connected to only a single unit in the preceding hidden layer, and the single output unit connected to all units in the final hidden layer.  That is, the weight matrix at each layer $i=1..D$ is a {\em diagonal} matrix $\diag(\u^i)$.  This presentation has $2dD$ parameters, as every layer has a different weight matrix.  However, it is easy to verify that if we initialize all layers to the same weight matrix, i.e.~with $\u^i=\u$, then the weight matrices will remain equal to each other throughout training, and so we can just use the $2d$ parameters in $\u\in\mathbb{R}^{2d}$ and take the weight matrix in every layer to be $\diag(\u)$, recovering the model \eqref{eq:diagNN}.

Since all operations in a linear neural net are linear, the model just implements a linear mapping from the input $\x$ to the output, and can therefor be viewed as an alternate parametrization of linear predictors.  That is, the functions $F(\u):\mathcal{X}\rightarrow\bR$ implemented by the model is a linear predictor $F(\u)\in\mathcal{X}^*$ (since $\mathcal{X}=\bR^d$, we also take $\mathcal{X}^*=\bR^d$), and we can write $f(\u,\x)=\innerprod{F(\u)}{\x}$, where for diagonal linear nets $F(\u)=\u_{+}^{D}-\u_{-}^{D}$.  In particular, for a trajectory $\u(t)$ in parameter space, we can also describe the corresponding trajectory $\w(t)=F(\u(t))$ of linear predictors.

The reason for using both the input features and their negation, and thus $2d$ units per layer, instead of a simpler model with $\u\in\bR^d$ and $F(\u)=\u^D$ is two-fold: first, this allows the model to capture mixed sign linear predictors even with even depth.  Second, this allows for scaling the parameters while still initializing at the zero predictor.  In particular, we will consider initializing to $\u(0)=\alpha \mathbf{1}$, which for the diagonal neural net model \eqref{eq:diagNN} corresponds to $\w(0)=F(\u(0))=0$ regardless of the scale of $\alpha$.  Such {\em unbiased initialization} was suggested by \citet{chizat2019lazy} in order to avoid scaling a bias term when scaling the initialization.

\citet{woodworth2020kernel} provided a detailed study of diagonal linear net regression using the square loss $\ell(\hat{y},y)=(\hat{y}-y)^2$. 
They showed that for an underdetermined problem (i.e.~with multiple zero-error solutions), for any finite $\alpha$ the gradient flow trajectory with squared loss and initialization $\u(0)=\alpha\mathbf{1}$ converges to a zero-error (interpolating) solution minimizing the penalty $Q^D_{\alpha^D}$, where $Q^D_\mu$ is defined as:
\begin{equation}
\begin{split}
    Q_\mu^D(\w)=\sum_{i=1}^d q^D\left(\frac{\w_i}{\mu}\right), \text{where }&q^D(z)=\left\{\begin{array}{ll}2-\sqrt{4+z^{2}}+z\cdot\arcsinh\left(\frac{z}{2}\right)&\text{for }D=2\\\int_{0}^{z}h_{D}^{-1}\left(s\right)ds&\text{for }D>2\end{array}\right.\\
    \text{and } & h_{D}\left(s\right)=\left(1-s\right)^{-\frac{D}{D-2}}-\left(1+s\right)^{-\frac{D}{D-2}}.
\end{split}
\label{eq:Q_function}
\end{equation}
For all $D\geq 2$, $Q_\mu^D$ with $\mu=\alpha^D$ interpolates between the $\ell_1$ norm as $\alpha\to 0$, which corresponds to the rich limit previously shown by   \citet{gunasekar2017implicit} and \citet{arora2019implicit}, and the $\ell_2$ norm as $\alpha\to\infty$, which is the RKHS norm defined by the tangent kernel at initialization.

Can we identify a similar transition behavior between kernel and rich regimes with exponential loss?

\section{Theoretical Analysis: Between the Kernel and Rich Regimes}
\label{sec:analysis}

We are now ready to state our main result, which describe the limit behaviour of gradient flow for classification with linear diagonal networks and the exp-loss, on separable data, and where the initialization scale $\alpha$ and the training accuracy $1/\epsilon$ go to infinity together.

First, we establish that if the data is separable, even though the objective $\mathcal{L}(\u)$ is non-convex,  gradient flow will minimize it, i.e.~we will have $\mathcal{L}\left(\u(t)\right)\rightarrow 0$.  In particular, gradient flow will lead to a separating predictor.  We furthermore obtain a quantitative bound on how fast the training error decreases as a function of the initialization scale $\alpha$, the depth $D$ and the $\ell_2$ separation margin of the data,  $\gamma_{2}=\max_{\norm{\w}_2=1}\min_{n}y_n\x_n^\top\w$: 
\begin{lemma}\label{lemma:loss_bound}
For $D\geq 2$, any fixed $\alpha$, and  $\forall t$,
$\mathcal{L}\left(\u(t)\right)\leq\frac{1}{1+2D^{2}\alpha^{2D-2}\gamma_{2}^{2}t}$.
\end{lemma}

The proof appears in Appendix \ref{loss_bound_proof}.
We now turn to ask which separating classifier we'd get to, if optimizing to within training accuracy $\epsilon$ and initialization of scale $\alpha$, in terms of the relationship between these two quantities.  To do so we consider different mapping $\epsilon(\alpha)$ between the initialization scale and training accuracy, such that $\epsilon:\bR_{++}\to(0,1]$ is strictly monotonic and $\lim_{\alpha\to\infty}\epsilon(\alpha)=0$.
We call $\epsilon(\alpha)$ the {\em stopping accuracy function}.
For each $\alpha$, we follow the gradient flow trajectory until time $T_{\alpha}$ such that $\mathcal{L}\left(\u(T_\alpha)\right)=\epsilon(\alpha)$ and denote $\hat{\w}_{\alpha}=\frac{\w\left(T_{\alpha}\right)}{\gamma\left(T_{\alpha}\right)}$.  We are interested in characterizing the limit point $\hat{\w}=\lim_{\alpha\rightarrow\infty}\hat{\w}_{\alpha}$
for different stopping accuracy functions $\epsilon\left(\alpha\right)$, assuming this limit exists.

\paragraph{The Kernel Regime}
We start with showing that if $\epsilon(\alpha)$ goes to zero slowly enough, namely if $\log 1/\epsilon$ is sublinear in $\alpha^D$, then with large initialization we obtain the $\ell_2$ bias of the kernel regime:

\begin{theorem}\label{thm:l2_limit}
For $D\geq 2$, if $\epsilon(\alpha)=\exp\big(-o(\alpha^D)\big)$
then
$\hat{\w}=\argmin_{\w} \|\w\|_2\;\text{ s.t. }\forall n,  y_n\w^{\top}\x_n\ge 1$.
\end{theorem}

The proof for $D=2$ appears in Appendix \ref{l2_limit_proof} and the proof for $D>2$ appears in Appendix \ref{l2_limit_proof_D}.

\paragraph{Escaping the kernel regime}

Theorem~\ref{thm:l2_limit} shows that escaping the kernel regime requires optimizing to higher accuracy, such that $\log 1/\epsilon$ is at least linear in the initialization scale $\alpha^D$.  Complementing Theorem~\ref{thm:l2_limit}, we show a converse: that indeed the linear scaling $\log 1/\epsilon=\Theta(\alpha^D)$ is the transition point out of kernel regime, and once $\log 1/\epsilon=\Omega(\alpha^D)$ we no longer obtain the kernel $\ell_2$ bias.

To show this, we first state a condition about the stability of support vectors\footnote{Data point $(\x_k,y_k)$ is a support vector at time $t$ if $y_k\x^{\top}_k\w(t)=\min_n y_n\x^{\top}_n\w(t)=\gamma(t)$.}. Given a dataset $S=\{(\x_n,y_n): n=1,2,\ldots N\}$ and a stopping accuracy function $\epsilon(\alpha)$:
\begin{condition}[Stability condition]\label{assmp:intermediate_depthD}
For all $k\in\left[N\right]$ such that $y_k\x_{k}^{\top}\hat{\w}>1$, and large enough $\alpha$,
there exists
$\epsilon^{\star}(\alpha)=\exp\left(-o(\alpha^D)\right)$
and $\rho_{0}>1$ such that
$\forall t$ with $\mathcal{L}(t)\in\left[\epsilon\left(\alpha\right),\epsilon^{\star}\left(\alpha\right)\right]:\frac{y_k\x_{k}^{\top}\w\left(t\right)}{\gamma(t)}\geq\rho_{0}$.
\end{condition}

We say that \condref{assmp:intermediate_depthD} holds {\em uniformly} if given a dataset $S$ it holds for {\em all} stopping functions $\epsilon(\alpha)=\exp(-\Omega(\alpha^D))$.

For the linearized model \condref{assmp:intermediate_depthD} holds almost surely 
(uniformly), see details in Appendix \ref{linearized_condition}. Therefore, if \condref{assmp:intermediate_depthD} does not hold, it follows that
we exit the kernel regime.

 If \condref{assmp:intermediate_depthD} does hold we show in  \thmref{thm:Q_alpha_D} that when $\epsilon=\exp\left(-\Theta(\alpha^D)\right)$ we will be in the intermediate regime, leading to max-margin solution with respect to  $Q^D$ function in \eqref{eq:Q_function}, and again deviate from the kernel regime. 

\begin{theorem}\label{thm:Q_alpha_D}
Under \condref{assmp:intermediate_depthD}, for $D\geq 2$, if
$\lim\limits_{\alpha\rightarrow\infty}\frac{\alpha^{D}}{\log\left(1/\epsilon(\alpha)\right)}=\mu>0$,
then for $Q_{\mu}^{D}$ as defined in \eqref{eq:Q_function},
$\hat{\w}=\argmin_{\w}Q_{\mu}^{D}\left(\w\right)\;\text{ s.t. }\forall n,\,y_n\x_{n}^{\top}\w\geq1
$.
\end{theorem}

To prove \thmref{thm:Q_alpha_D} we show that the following KKT conditions hold: 
\begin{align*}
\exists \mathbf{\nu}\in\mathbb{R}_{\geq0}^{N}\;\text{ s.t. }\quad\nabla Q_{\mu}^D\left(\hat{\w}\right)  =\sum_{n=1}^N \nu_n y_n\x_n,\quad
\forall n:\,y_n\x_{n}^{\top}\hat{\w}  \geq1,\quad
\forall n:\,\nu_{n}\left(y_n\x_{n}^{\top}\hat{\w}-1\right)  =0.
\end{align*}
The result then follows from convexity of $Q_{\mu}^D\left(\w\right)$.

The proof for $D=2$ appears in Appendix \ref{q_alpha_proof} and the proof for $D>2$ appears in Appendix \ref{q_alpha_D_proof}.
\paragraph{The Rich Limit}

Above we saw that $\log 1/\epsilon$ being linear in $\alpha$ is enough to leave the kernel regime, i.e.~with this training accuracy and beyond the trajectory no longer behaves as if we were training a kernel machine.  We also saw that under \condref{assmp:intermediate_depthD}, when $\log 1/\epsilon$ is exactly linear in $\alpha$, we are in a sense in an ``transition'' regime, with bias given by the $Q^D_\mu$ penalty which interpolates between $\ell_2$ (kernel) and $\ell_1$ (the rich limit for $D=2$).  Next we see that once the accuracy $\log 1/\epsilon$ is superlinear, and again under \condref{assmp:intermediate_depthD}, we are firmly at the rich limit:
\begin{theorem}\label{thm:l1_limit_D}
Under \condref{assmp:intermediate_depthD}, for $D\geq2$ if
$\epsilon(\alpha)=\exp\left(-\omega(\alpha^D)\right)$
then
$\hat{\w}=\argmin_{\w}\left\Vert \w\right\Vert _{1}\,\,\,\,\text{s.t.}\,\,\forall n:\,y_n\x_{n}^{\top}\w\geq1
$.

For $D=2$ the result holds also with a weaker condition, when $\epsilon^{\star}\left(\alpha\right)$ in \condref{assmp:intermediate_depthD} is replaced with $\epsilon^{\star}\left(\alpha\right)=\exp\left[-o\left(\alpha^{2}\log\frac{\log(1/\epsilon(\alpha))}{\alpha^2}\right)\right]$.
\end{theorem}

The proof for $D=2$ appears in Appendix \ref{L1_limit_2_proof} and the proof for $D>2$ appears in Appendix \ref{L1_limit_D_proof}.

For $D>2$ we know from \thmref{thm:rich_limit} that the implicit bias in the rich limit is given by an $\ell_{2/D}$ quasi norm penalty, and not by the $\ell_1$ penalty as in \thmref{thm:l1_limit_D}.  It follows that when \condref{assmp:intermediate_depthD} holds, the max $\ell_1$-margin predictor must also be a first order stationary point of the $\ell_{2/D}$ max margin problem.  As we demonstrate in \secref{sec:simulations}, this is certainly not always the case, and for many problems the max $\ell_1$-margin is not a stationary point for the $\ell_{2/D}$ max margin problem---in those cases \condref{assmp:intermediate_depthD} does not hold.  It might well be possible to show that a super-linear scaling is sufficient to reach the rich limit, be it the max $\ell_1$-margin for depth two, or the max $\ell_{2/D}$-margin (or a stationary point for this non-convex criteria) for higher depth, and we hope future work will address this issue.

\paragraph{Role of depth} From \thmref{thm:Q_alpha_D} we have that asymptotically $\epsilon(\alpha)=\exp(-\alpha^D/\mu)$. \citet{woodworth2020kernel} analyzed the $Q^D$ function and concluded that in order to have $\delta$ approximation to $\ell_1$ limit (achieved for $\mu\to 0$) we need to have $\mu=\exp(-1/\delta)$ for $D=2$, and $1/\mu=\mathrm{poly}(1/\delta)$ for $D>2$. We conclude that in order to have $\delta$ approximation to $\ell_1$ limit we need the training accuracy to be $\epsilon=\exp(-\alpha^D \exp(1/\delta))$ for $D=2$, and $\epsilon=\exp(-\alpha^D \mathrm{poly}(1/\delta))$ for $D>2$. Thus, depth can mitigate the need to train to extreme accuracy. We confirm such behaviour in simulations.

\section{Numerical Simulations and Discussion}
\label{sec:simulations}

We numerically study optimization trajectories to see whether we can observe the asymptotic phenomena studied at finite initialization and accuracy.  We focus on low dimensional problems, where we can plot the trajectory in the space of predictors. In all our simulations we employ the Normalized GD algorithm, where the gradient is normalized by the loss itself, to accelerate convergence \citep{DBLP:conf/aistats/NacsonLGSSS19}. In all runs the learning rate was small enough to ensure gradient flow-like dynamics (always below $10^{-3}$).

\begin{figure*}[t!]
\centering
    \quad\quad\subfigure[]{\label{fig:sphere1}\includegraphics[width=0.35\textwidth]{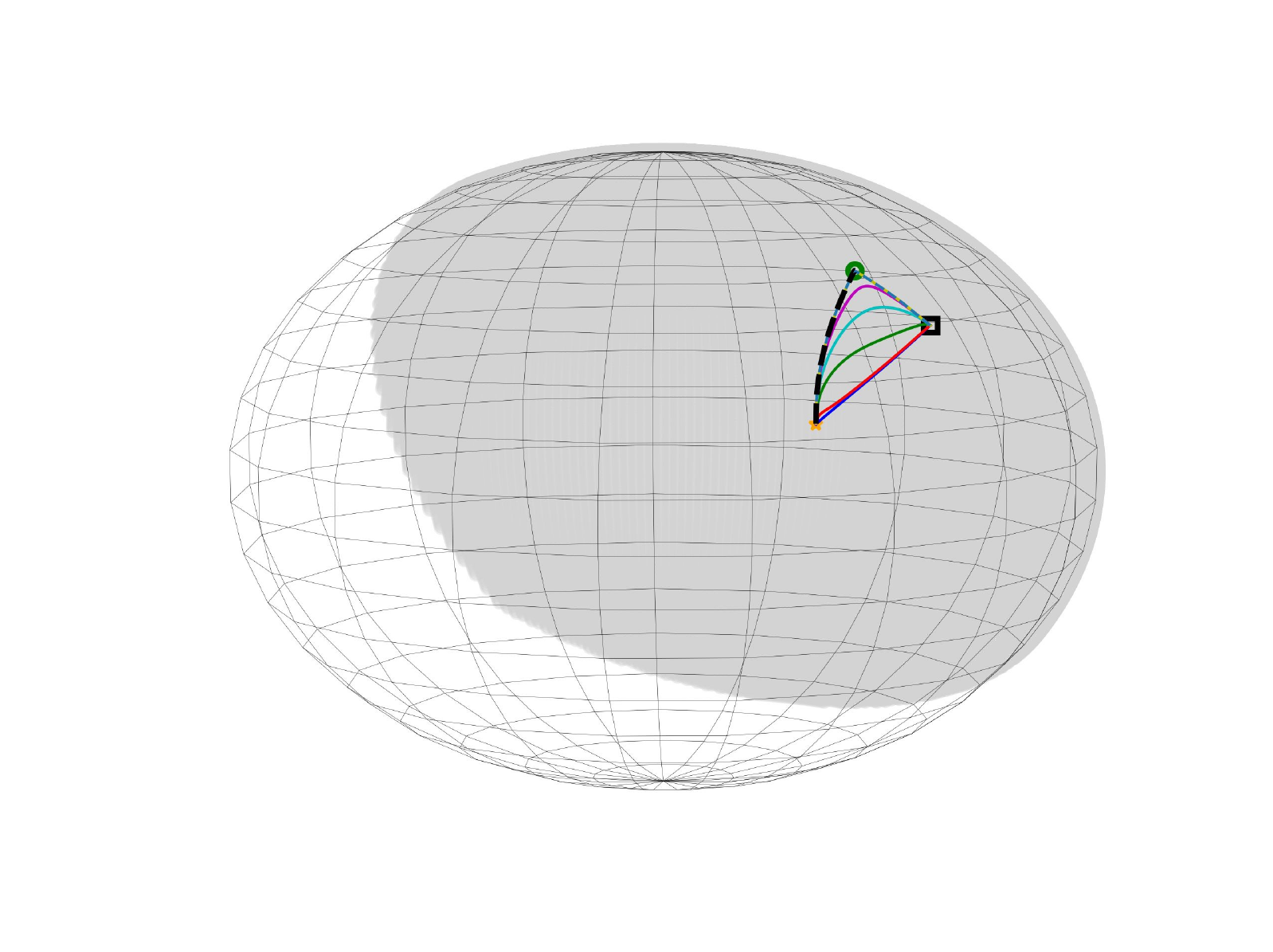}}	\hfill
	\subfigure[\scriptsize Data: $(0.3,1.5,1),(1.5,3,1),(1,2.5,1)$]{\label{fig:uniqueL1}\includegraphics[width=0.47\textwidth]{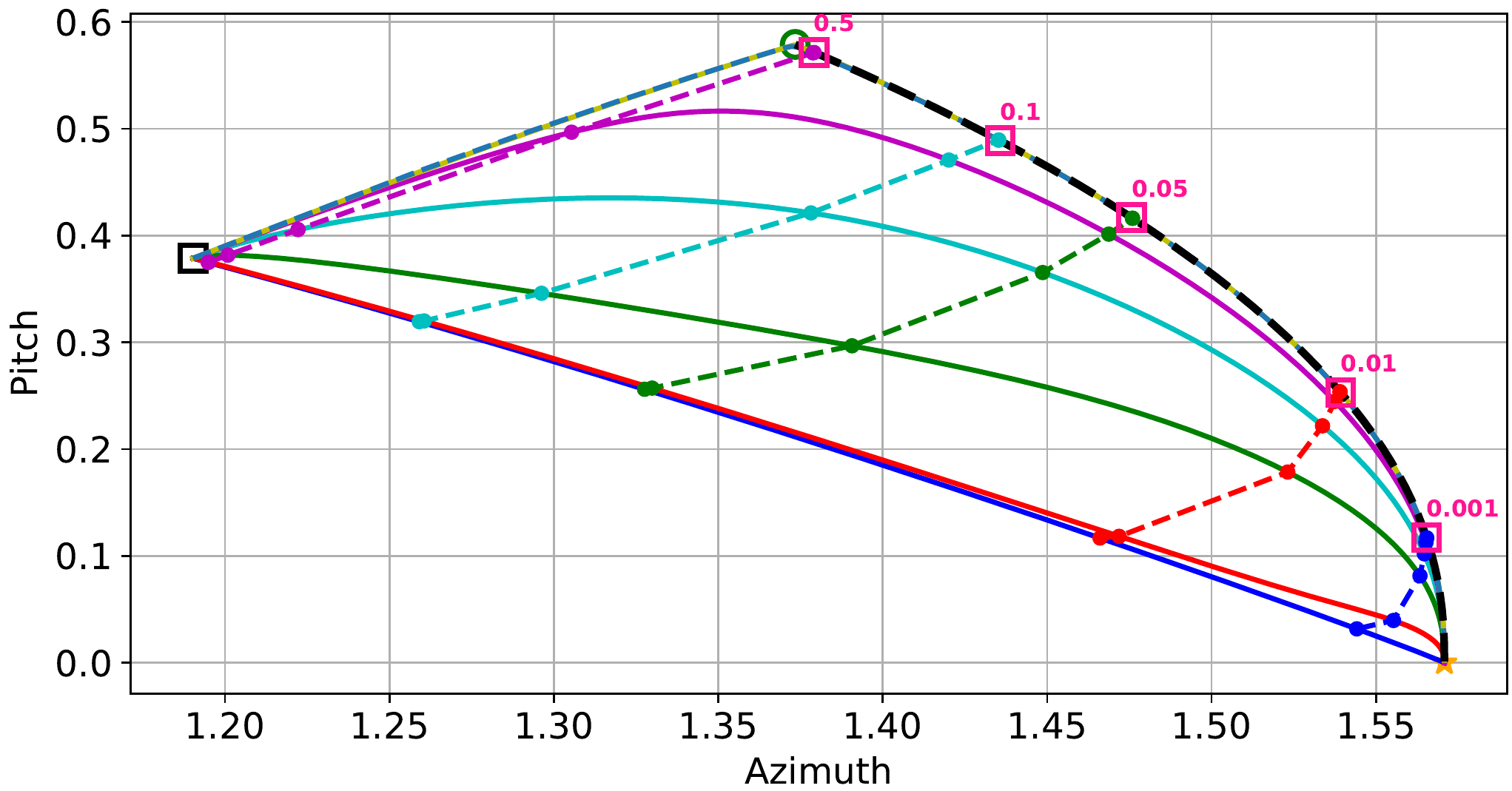}}   \\
	\subfigure[\scriptsize Data: $(0.5,1,1),(1,1.5,1),(1.5,2,0.5)$]{\label{fig:manyL1}\includegraphics[width=0.47\textwidth]{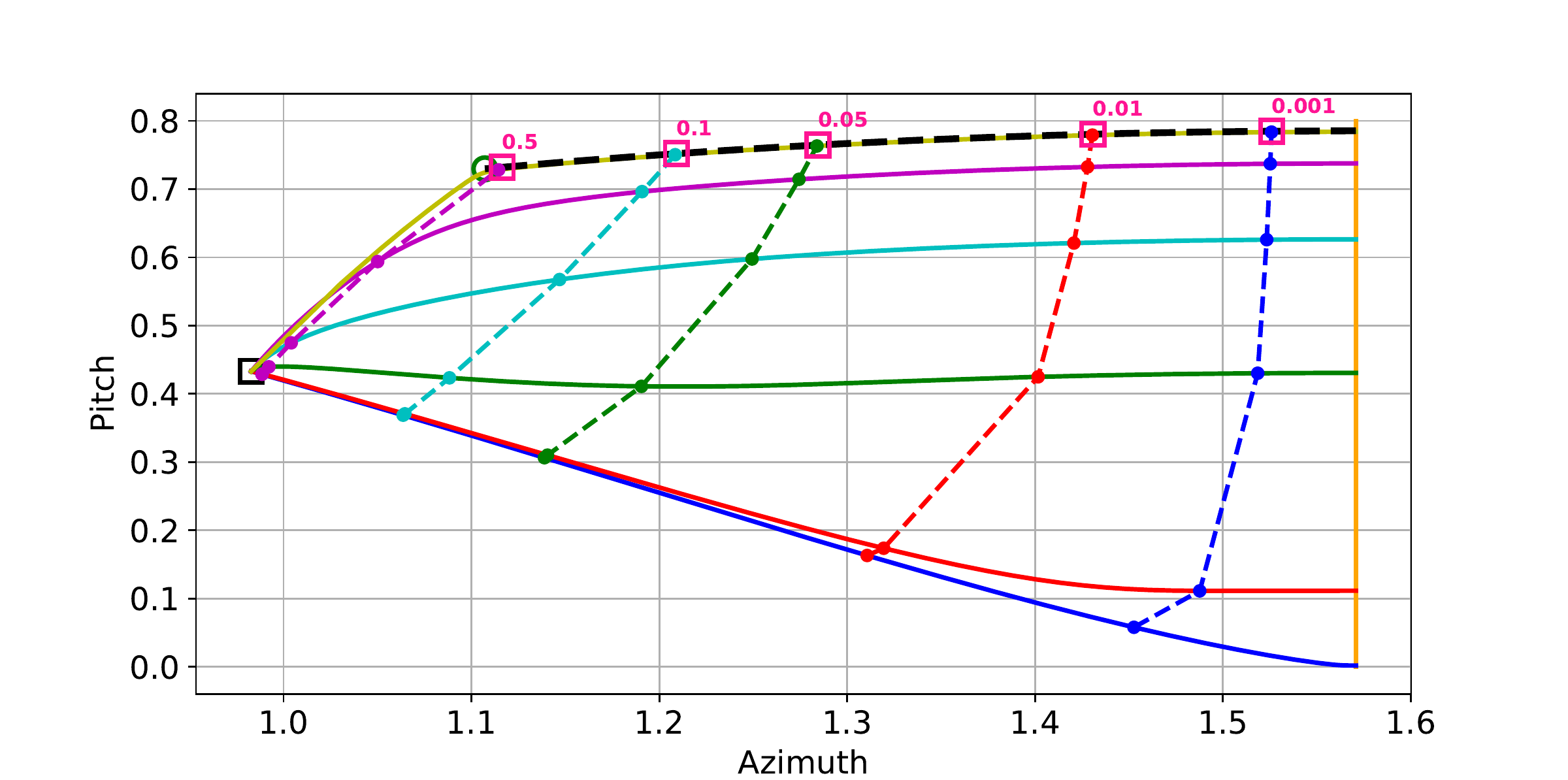}}\hfill
	\subfigure[\scriptsize Data: $(3,1,1),(2.7,2,1.5),(4.5,2.6,0.5)$]{\label{fig:noQ}\includegraphics[width=0.47\textwidth]{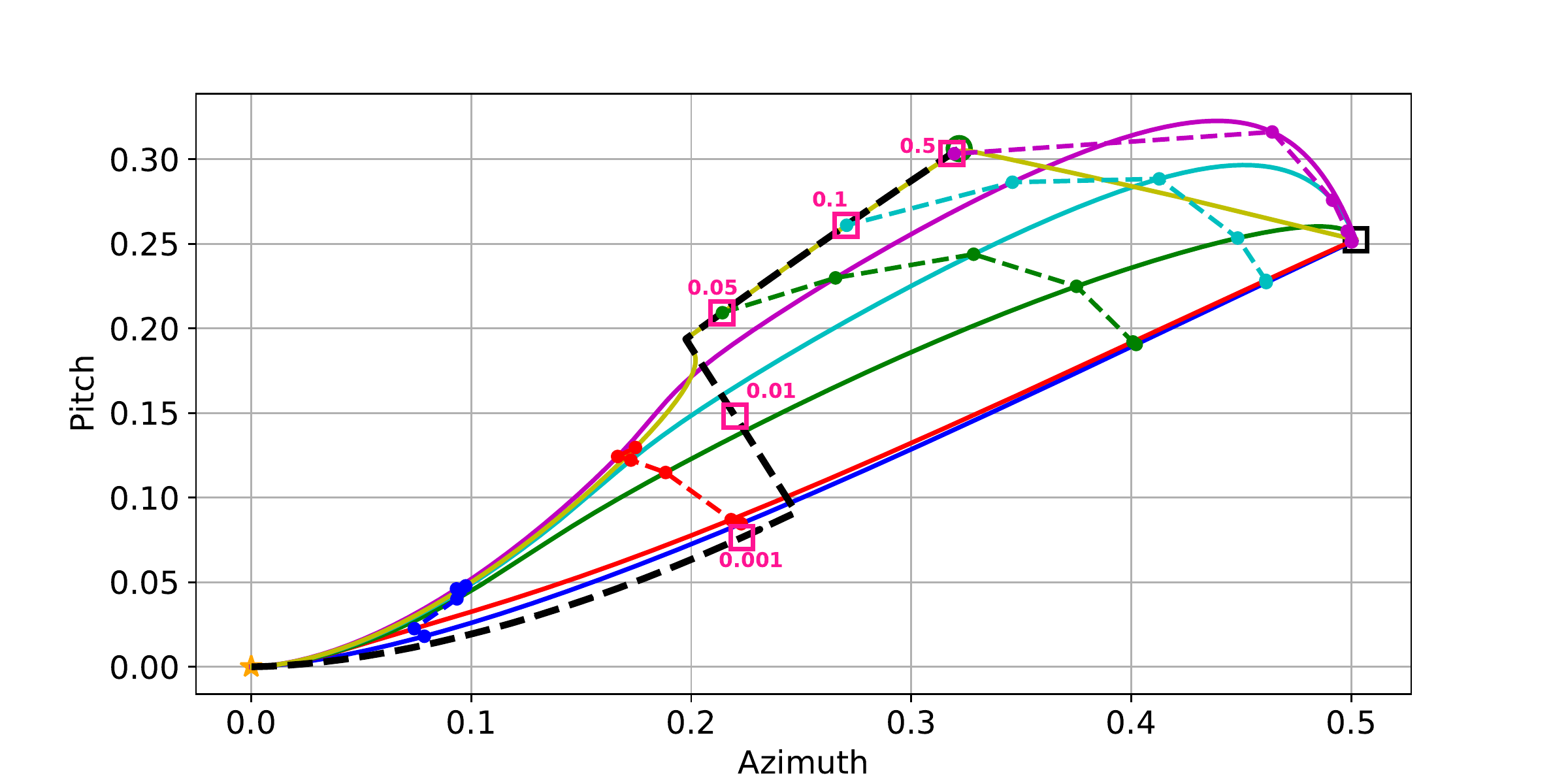}}
	\includegraphics[width=0.95\textwidth]{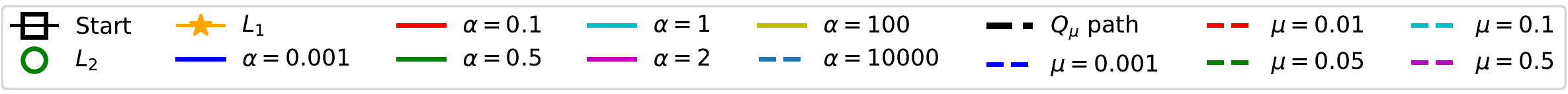}
\small \caption{\small Optimization trajectories for $3$ simple datasets in depth $2$ linear diagonal network (b-d). Each point in Azimuth-Pitch plane represents a normalized classifier $\nicefrac{\w}{\norm{\w}_2}$. The curves corresponding to $\alpha$ in the legend are the entire gradient flow trajectories initialized with the respective $\alpha$. The curves corresponding to $\mu$ are end points of gradient flow trajectories for different $\alpha$ with stopping criteria set as $\epsilon(\alpha)=\exp(-\alpha^2/\mu)$. The pink squares represent the directions along the $Q_{\mu}$ max-margin path for the appropriate $\mu$ marked near the square. The dynamics in (b) takes place on a small part of the sphere as shown in (a), where the grey area represents all separating directions. 
\label{fig:depth2_dim3}}\vspace{-12pt}
\end{figure*}

\begin{figure*}[t!]
\centering
	\subfigure[\scriptsize Data:$(0.6,0.7,0.8),(0.7,0.6,0.6),(1.0,0.5,0.5)$]{\label{fig:data1_D3_d3_paths}\includegraphics[width=0.45\textwidth]{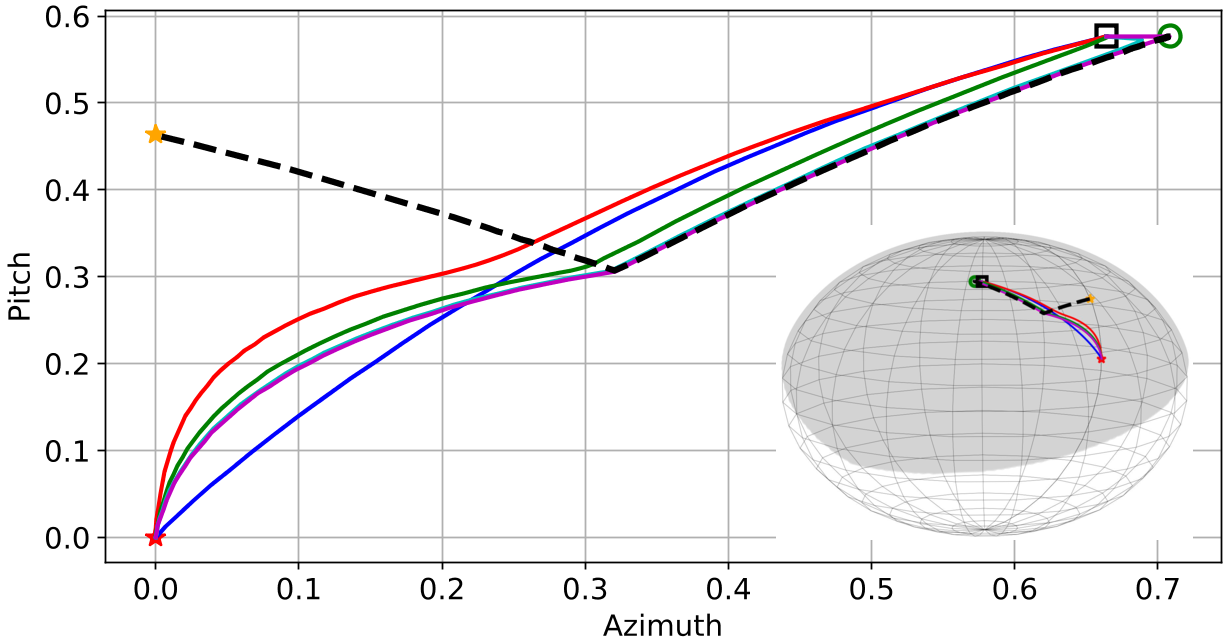}}
    \hfill
	\subfigure[\scriptsize Data:$(0.6,0.7,0.1),(0.4,0.6,0.6),(1.0,0.5,0.5)$]{\label{fig:data2_D3_d3_paths}\includegraphics[width=0.45\textwidth]{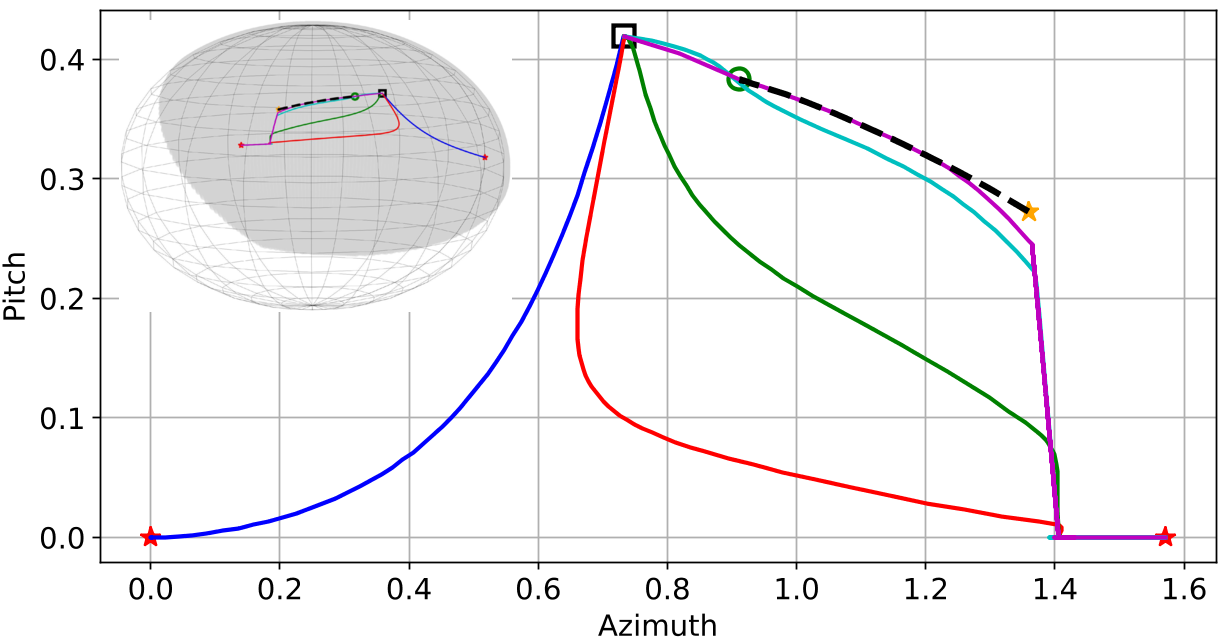}}
	\includegraphics[width=0.42\textwidth]{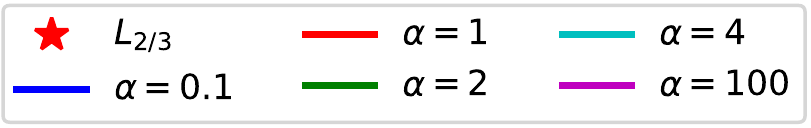}\quad\quad\quad\quad\quad
	\includegraphics[width=0.42\textwidth]{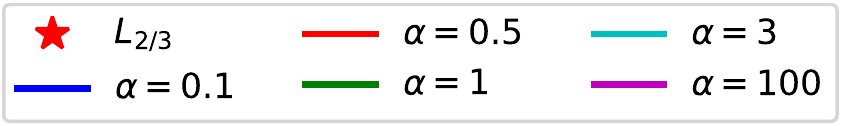}
	\small\caption{ \small Optimization trajectories for  $D=3$. The legend for $\ell_1$, $\ell_2$, start and $Q_\mu$ path is the same as in \figref{fig:depth2_dim3}
	\label{fig:depth3_dim3}}\vspace{-20pt}
\end{figure*}

\paragraph{Gradient flow trajectories} In \figref{fig:depth2_dim3} we plot trajectories for training depth $D=2$ diagonal linear networks in dimension $d=3$, on several constructed datasets, each consisting of three points.  The trajectory $\w(t)$ in this case is in $\bR^3$, and so the corresponding binary predictor given by the normalization $\nicefrac{\w(t)}{\norm{\w(t)}_2}$ lies on the sphere (panel (a)).  We zoom in on a small section of the sphere, and plot the trajectory of $\nicefrac{\w(t)}{\norm{\w(t)}_2}$ --- the axes correspond to coordinates on the sphere (given as azimuth and pitch). The first  step taken by gradient flow will be to the predictor proportional to the average of the data, $\frac{1}{N}\sum_n y_n \x_n$, and we denote this as the ``start'' point. The grey area in the sphere represents classifiers separating the data (with zero misclassification error), and thus directions where the loss can be driven to zero.  The question of ``implicit bias'' is which of these classifiers the trajectory will converge to.  With infinitesimal (small) stepsizes, the trajectories always remain inside this area (\ie just finding a separating direction is easy), and in a sense the entire optimization trajectory is driven by the implicit bias.

Panel \ref{fig:uniqueL1} corresponds to a simple situation, with a unique $\ell_1$-max-margin solution, and where the support vectors for the $\ell_2$-max-margin and $\ell_1$-max-margin are the same (although the solutions are not the same!), and so the support vectors do not change throughout optimization and \condref{assmp:intermediate_depthD} holds uniformly.  For large initialization scales ($\alpha=100$ and $\alpha=10000$, which are indistinguishable here), the trajectory behaves as the asymptotic theory tells us: from the starting point (average of the data), we first go to the $\ell_2$-max-margin solution (green circle, and recall that this is also the $Q^2_\infty$-max-margin solution), and then follow the path of $Q^2_\mu$-max-margin predictors for $\mu$ going from $\infty$ to zero (this path is indicated by the dashed black lines in the plots), finally reaching the $\ell_1$-max-margin predictor (orange star, and this is also the $Q^2_0$-max-margin solution).  For smaller initialization scales, we still always reach the same endpoint as $\epsilon\to0$ (as assured by the theory), but instead of first visiting the $\ell_2$-max-margin solution and traversing the $Q^2_\mu$ path, we head more directly to the $\ell_1$-max-margin predictor.  This can be thought of as the effect of initialization on the implicit bias and kernel regime transition: with small initialization we will never see the kernel regime, and go directly to the ``rich'' limit, but with large initialization we will initially remain in the kernel regime (heading to the $\ell_2$-max-margin), and then, only when the optimization becomes very accurate, escape it gradually.

To see the relative effect of scale and training accuracy, and following our theory, we plot for different  values of $\mu$, the different points along trajectories with initialization $\alpha$ such that we fix the stopping criteria as $\epsilon(\alpha)=\exp(-\alpha^2/\mu)$ (dashed cross-lines in the plots).  Our theory indicates that for any value of $\mu$, as $\alpha\to\infty$, the dashes line would converge to the $Q^2_\mu$-max-margin (a specific point on the $Q^2_\mu$ path), and this is indeed confirmed in Panel \ref{fig:uniqueL1}, where the dashed lines converge to pink squares, which correspond to points along the $Q^2_\mu$ path for the appropriate $\mu$ values. The clear correspondence between points with the same relationship $\mu$ between initialization and accuracy confirms that also at relatively small initialization scales, this parameter is the relevant relationship between them.

From the value $\mu$ we can also extract the actual training accuracy.  E.g., we see that for a relatively large initialization scale $\alpha=100$, escaping the kernel regime (getting to the first pink square just removed from the $\ell_2$-max-margin solution, with $\mu=0.5)$, requires optimizing to loss $\epsilon=\exp(-10^4/0.5)\approx 10^{-8700}$, while getting close to the asymptotic limit of the trajectory (the last pink square, with $\mu=0.001$) requires $\epsilon\approx 10^{-4\cdot10^6}$ (that's four millions digits of precision).  Even with reasonable initialization at scale $\alpha=1$, getting to this limit requires $\epsilon\approx 10^{-434}$.

Panel (c) displays the trajectories for another dataset where \condref{assmp:intermediate_depthD} holds uniformly, but
the $\ell_1$-max-margin solution is not unique. Although will always (eventually) converge to a $\ell_1$-max-margin solution, which one we converge to, and thus the implicit bias, {\em does} depends on the initialization.
Panel (d) shows a situation where the support vectors for the $\ell_2$-max-margin and $\ell_1$-max-margin solutions are different, and so change during optimization, and \condref{assmp:intermediate_depthD} does not hold uniformly. The condition does hold for $\mu > \mu_0$ where $\mu_0\approx 0.04$, but does not hold otherwise. In this case, even for very large initialization scales $\alpha$, the trajectory does not follow the $Q^2_\mu$ path entirely.  It is interesting to note that the $Q^2_\mu$ is not smooth, which perhaps makes it difficult to follow. Around the kink in the path (at $\mu=\mu_0$), it does seem that larger initialization scales are able to follow it a bit longer, and so perhaps with huge initialization (beyond our simulation ability), the trajectory {\em does} follow the $Q^2_\mu$ path.  Whether the trajectory follows the path for sufficiently large initialization scales even when \condref{assmp:intermediate_depthD} fails, or whether we can characterize its limit behaviour otherwise, remains an open question.

In \figref{fig:depth3_dim3} we show optimization trajectories for depth 3 linear diagonal network where \condref{assmp:intermediate_depthD} does not hold uniformly. As we can observe, in both cases the trajectory for large $\alpha$ will first go to $\ell_2$ max-margin, then stay near the $Q_\mu$ path, until the trajectory starts to deviate in a direction of $\ell_{2/3}$ max-margin solution, whereas the $Q_\mu$ path continues to the $\ell_1$ max-margin point. In \figref{fig:data2_D3_d3_paths} we observe that there is a local minimum point at (0,0) and for small $\alpha$ the trajectory converges to it. Further discussion about convergence to local minima in high dimension appears in Appendix \ref{app:simulations}.

\paragraph{Initialization Scale vs Training Accuracy}
\label{sec:scale_vs_accuracy}

\begin{figure*}[t!]
\centering
	\subfigure[Excess $\ell_1$ norm for training accuracy $\tilde{\gamma}$. ]{\label{fig:D1}\includegraphics[width=0.4\textwidth]{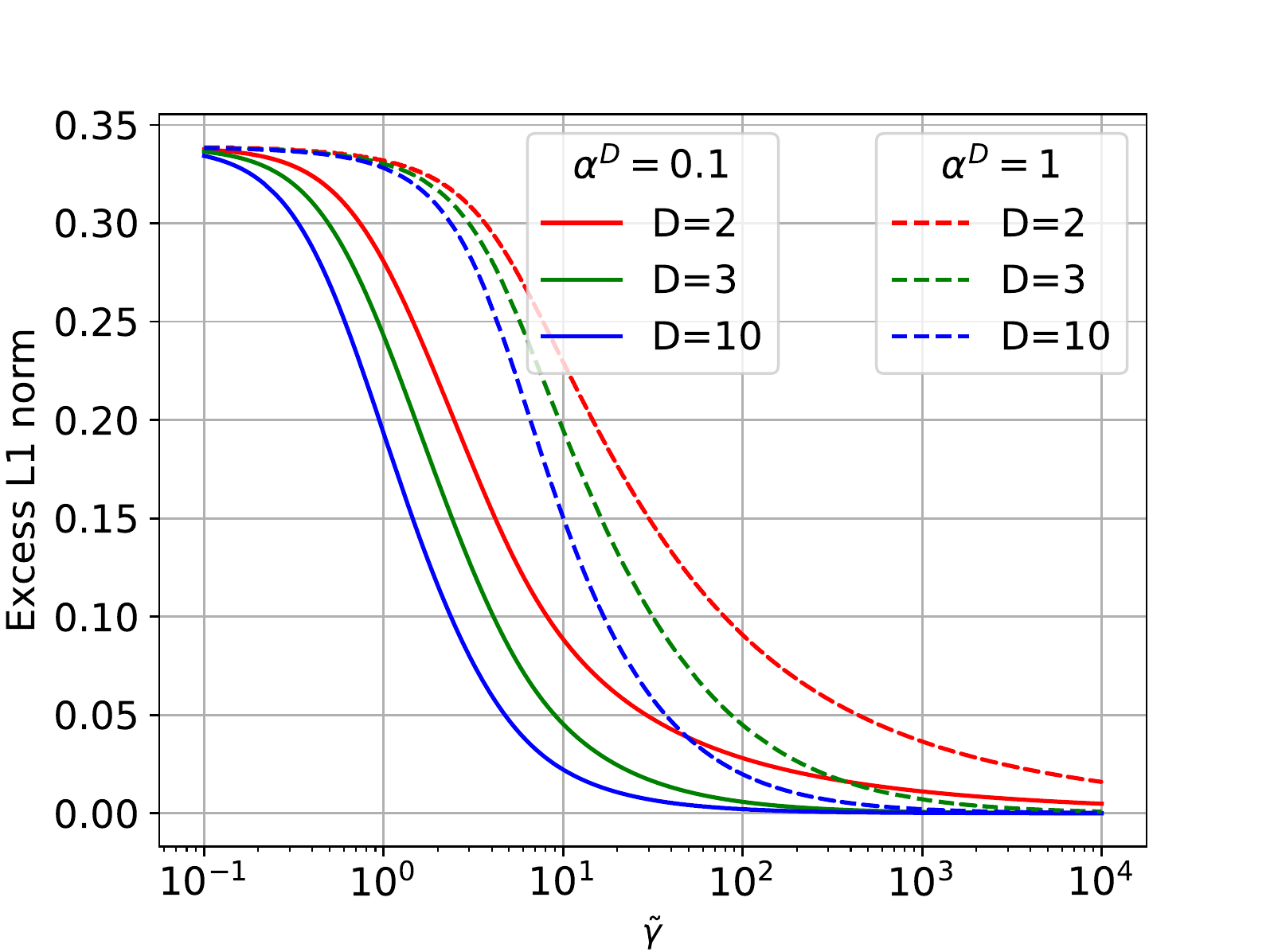}}\quad\quad
	\subfigure[Accuracy vs Initialization]{\label{fig:D2}\includegraphics[width=0.4\textwidth]{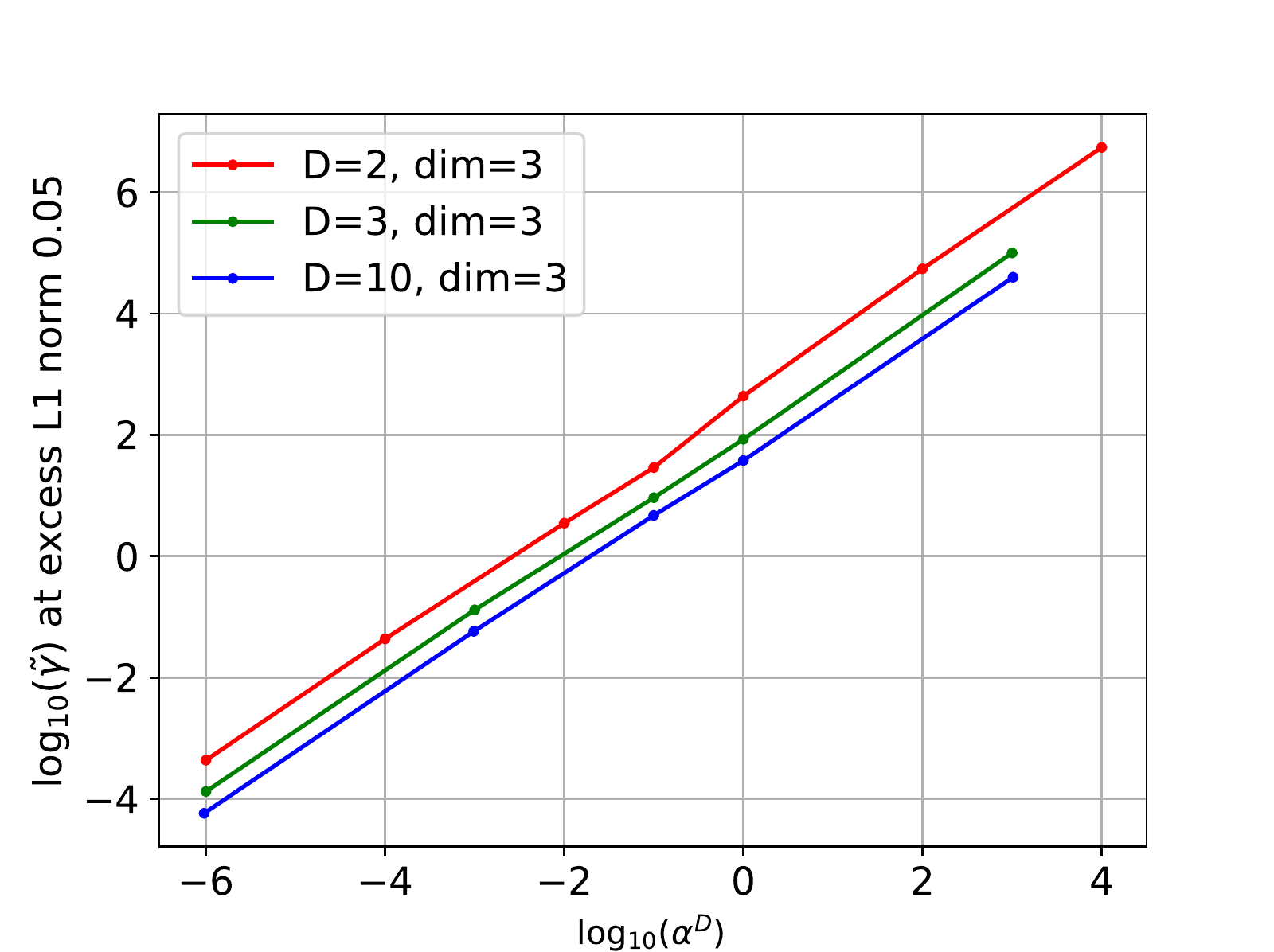}}
	\caption{ In (a) we plot the excess $\ell_1$ norm, defined as $\Vert \w(t)\Vert_1/\Vert \w_{\ell_1}\Vert_1-1$ where $\w_{\ell_1}$ is the $\ell_1$ max-margin (minimum norm) solution, as a function of $\tilde{\gamma}$. For a fixed excess $\ell_1$ norm of 0.05, in (b) we plot how long we need to optimize, given some initialization scale, to obtain the $5\%$ closeness to $\ell_1$ max-margin solution.}
	\label{fig:scale_accuracy}
	\vspace{-5pt}
\end{figure*}

Using the same dataset from \figref{fig:uniqueL1} we examine the question: given some initialization scale, how long we need to optimize to be in the rich regime?
In \figref{fig:D1} we demonstrate how the initialization and depth affect the  convergence rate to the rich regime. Specifically, we chose two reasonable initialization scales $\alpha^D$, namely $0.1$ and $1$, and show their convergence to the $\ell_1$ max-margin solution as a function of the training accuracy  $\tilde{\gamma}$ (recall that $\tilde{\gamma}=\log(1/\epsilon)$), for depths $D=2,3,10$. We chose this dataset so that the rich regime is the same for all depths (i.e., the minimum $\ell_1$ norm solution corresponds also to the minimum $\ell_{2/3}$ and $\ell_{2/10}$ quasi-norm solutions).

As previously discussed, even on such a small dataset (with three datapoints) we are required to optimize to an incredibly high precision --- in order to converge near the rich regime.
Notably, the situation is improved and we converge faster to the rich regime when the initialization scale is smaller and/or the depth is larger.
Unfortunately, taking the initialization scale to $0$ will increase the time needed to escape the vicinity of the saddle point $\u=0$. For example, \citet{shamir2018exponential} showed that exiting the initialization may take exponential-in-depth time.

In \figref{fig:D2} we examine the relative scale between (log) accuracy and (log)  initialization needed to obtain $5\%$ closeness to $\ell_1$ max-margin. Based on the asymptotic result in Theorem \ref{thm:Q_alpha_D}, we expect that $\tilde{\gamma}\propto\alpha^D$ or $\log(\tilde{\gamma}) = a\log(\alpha^D) + b$, for some constants $a=1$ and $b\in\mathbb{R}$. And indeed, in \figref{fig:D2} we obtain $a=1$, as expected. Note that although our theoretical results are valid for $\alpha\rightarrow\infty$, we obtain the same accuracy vs initialization rate also  for small $\alpha$.
Moreover, the intercept of the lines decreases when increasing the depth $D$, which matches the observed behavior on \figref{fig:D1} and to the discussion about the effect of depth in \secref{sec:analysis}.

\section*{Broader Impact}

The goal of this work is to shed light on the implicit bias hidden in the training process of deep networks. These results may enable a better understanding of how hyperparameters select the types of solutions that deep networks converge to, which in turn affect their final generalization performance and hidden biases. This could lead to better performance guarantees or to improved training algorithms which quickly converge to beneficial types of biases. Eventually, we believe progress on these fronts can transform deep learning from the current nascent “alchemy” age (where all the “knobs and levers” of the model and the training algorithm are tuned mostly heuristically during research and development), to a more mature field (like “chemistry”), which can be seamlessly integrated in many real world applications that require high performance, safety, and fair decisions.

Our guiding principal is that when studying a new or not-yet-understood phenomena, we should first study it in the simplest model that shows it, so as not to get distracted by possible confounders, and to enable a detailed analytic understanding,  \eg when understanding or teaching many statistical issues, we would typically start with linear regression, understand the phenomena there, and \emph{then} move on to more complex models.  In the specific case here, one of the few models where we have an analytic handle on the implicit bias in the ``rich'' regime are linear diagonal networks, and it would be very optimistic to hope to get a detailed analytic description of the more complex phenomena we study in models where we can't even understand the endpoint.

\paragraph{Acknowledgements}
The research of DS was supported by the Israel Science Foundation (grant No. 31/1031), and by the Taub Foundation. This work was partially done while SG, JDL, NS, and DS were visiting the Simons Institute for the Theory of Computing. BW is supported by a Google Research PhD fellowship.
JDL acknowledges support of the ARO under MURI Award W911NF-11-1-0303,  the Sloan Research Fellowship, and NSF CCF 2002272.

\bibliographystyle{plainnat}
\bibliography{bib}

\newpage
\appendix
{\centering\Large\bfseries
	Appendix
}

\section{Outline}

This appendix is organized as follows:
In \secref{proof_pre} we provide preliminaries and notations used in the proofs.
In \secref{auxiliary_lemmas} we prove auxiliary lemmas that characterize the dynamics of $\w\left(t\right)$ and bound the norm of $\w\left(t\right)$.
In \secref{loss_bound_proof} we prove the loss bound in \lemref{lemma:loss_bound}.
In \secref{linearized_condition} we prove that \condref{assmp:intermediate_depthD} holds for the linearized model.
In the proofs we distinguish between the case $D=2$ and $D>2$ since the dynamics is different. In \secref{proofs_D2} we prove the results for $D=2$ and in \secref{proofs_D} we prove the results for $D>2$. Finally, in \secref{app:simulations} we provide additional simulation results and implementation details.

\section{Preliminaries and Notations for Proofs}
\label{proof_pre}
To simplify notation in the proofs, without loss of generality we assume that $\forall n:y_n=1$, as equivalently we can re-define $y_n\x_n$ as $\x_n$.

\paragraph{Path parametrization:} In the proofs we parameterize the optimization path in terms of $\tilde{\gamma}$. 
Recall that $\tilde{\gamma}=-\log\epsilon$ 
and $\tilde{\gamma}(t)$ is monotonically increasing along the gradient flow path starting from $\tilde{\gamma}(0)=-\log\epsilon(0)=0$. Accordingly, the stopping criteria is $\tilde{\gamma}(\alpha)=\tilde{\gamma}(T_\alpha)=-\log\epsilon(\alpha)$. We also overload notation and denote $\w(\tilde{\gamma}')=\w(t_{\tilde{\gamma}'})$ and $\gamma(\tilde{\gamma}')=\gamma(t_{\tilde{\gamma}'})$ where $t_{\tilde{\gamma}'}$ is the unique $t$ such that $\tilde{\gamma}(t)=\tilde{\gamma}'$.
Moreover, in this appendix we restate the conditions and theorems in terms of $\tilde{\gamma}$ rather than $\epsilon$.

\paragraph{Notation:}
We use the following notations:
\begin{itemize}
    \item $X=\left[\x_{1},\ldots,\x_{N}\right]\in\mathbb{R}^{d\times N}$ denotes the data matrix.
    \item $\tilde{X}=\left[\tilde{\x}_{1},\dots,\tilde{\x}_{N}\right]\in\mathbb{R}^{2d\times N}$ denotes the augmented data matrix where $\tilde{\x}_{n}=\left[\begin{array}{c}\x_{n}\\-\x_{n}\end{array}\right]\in\mathbb{R}^{2d}$.
    \item $\bar{x}_{i}=\sum_{n=1}^{N}\left|x_{n,i}\right|$ where $x_{n,i}$ is the coordinate $i$ of $\x_{n}$. Also $\bar{x}=\max_{i}\left(\bar{x}_{i}\right)$.
    \item $x_{\max}=\max_{n}\left\Vert \x_{n}\right\Vert _{2}$.
    \item For some vector $\mathbf{z}$ we denote by $\diag(\mathbf{z})$ the diagonal matrix with diagonal $\mathbf{z}$, and $[\mathbf{z}]_i$ is the $i$ coordinate.
    \item The $\ell_2$ margin at time $t$ is $\gamma_{2}\left(t\right)=\frac{\min_{n}\left(\x_{n}^{\top}\w\left(t\right)\right)}{\left\Vert \w\left(t\right)\right\Vert _{2}}$. Recall that $\gamma_{2}=\max_{\norm{\w}_2=1}\min_{n}\x_n^\top\w$.
    \item $\partial^{\circ}$ denotes the local sub-differential (Clarle's
        sub-differential) operator defined as
        \[
        \partial^{\circ}h\left(\mathbf{z}\right)=\textrm{conv}\left\{ \mathbf{v}:\exists \mathbf{z}_{k}\,\,\,\textrm{s.t.}\,\,\mathbf{z}_{k}\rightarrow \mathbf{z}\,\,\,\textrm{and}\,\,\,\nabla h\left(\mathbf{z}_{k}\right)\rightarrow \mathbf{v}\right\} \,.
        \]
        Specifically, for $h\left(\mathbf{z}\right)=\left\Vert \mathbf{z}\right\Vert _{1}$:
        \[
        \partial^{\circ}\left\Vert \mathbf{z}\right\Vert _{1}=\left\{ \mathbf{v}\in\mathbb{R}^{d}:\forall i=1,...,d:\,\,\,-1\leq v_i\leq1\,\,\,\textrm{and}\,\,\,z_i\neq0\Rightarrow v_i=\mathrm{sign}\left(z_i\right)\right\} \,.
        \]
    \item We denote $\r\left(t\right)=\frac{1}{N}\exp\left(-X^{\top}\w\left(t\right)\right)$.
    Note that $\left\Vert \r\left(t\right)\right\Vert _{1}=\mathcal{L}\left(t\right)=\exp\left(-\tilde{\gamma}\left(t\right)\right)$.
    \item We denote $A\left(t\right)=\diag\left(4\sqrt{\w^{2}\left(t\right)+4\alpha^{4}\mathbf{1}}\right)$. This matrix is used in the proofs for $D=2$.
    \item For $D>2$ let:
    \begin{equation}
    h_{D}\left(z\right)=\left(1-z\right)^{-\frac{D}{D-2}}-\left(1+z\right)^{-\frac{D}{D-2}}\,\,\,\,\,,\,z\in\left(-1,1\right)~.
    \label{eq: h_D}
    \end{equation}
     Note that $h_{D}\left(z\right)$ is monotonically increasing, where
    $h_{D}\left(z\right)\overset{z\rightarrow-1}{\rightarrow}-\infty$
    and $h_{D}\left(z\right)\overset{z\rightarrow1}{\rightarrow}\infty$,
    and thus the inverse $h_{D}^{-1}$ is well defined, $h_{D}^{-1}:\left(-\infty,\infty\right)\rightarrow\left(-1,1\right)$.
    In addition, it is easy to verify that for $z\in\left(-1,1\right)$
    \begin{equation}
    h_{D}^{'}\left(z\right)\doteq\frac{dh_{D}\left(z\right)}{dz}\geq\frac{2D}{D-2}\label{eq:h_deriv}
    \end{equation}
    and
    \begin{equation}
    \lim_{z\rightarrow0}\frac{h_{D}\left(z\right)}{z}=\frac{2D}{D-2}\,.\label{eq: h_0}
    \end{equation}
    \item We denote $A_{D}\left(t\right)=\diag\left(\alpha^{2D-2}D\left(D-2\right)h_{D}^{'}\left(h_{D}^{-1}\left(\frac{\w\left(t\right)}{\alpha^{D}}\right)\right)\right)$. This matrix is used in the proofs for $D>2$.
\end{itemize}

\paragraph{Useful inequalities:} From the definitions of $\mathcal{L}(t)$ and $\tilde{\gamma}(t)$ we have that
\[
\mathcal{L}\left(t\right)=\frac{1}{N}\sum_{n=1}^{N}\exp\left(-\mathbf{x}_{n}^{\top}\mathbf{w}\left(t\right)\right)=\exp\left(-\tilde{\gamma}\left(t\right)\right)
\]
and thus
\begin{align}
&\frac{1}{N}\exp\left(-\gamma\left(t\right)\right)\leq\frac{1}{N}\sum_{n=1}^{N}\exp\left(-\mathbf{x}_{n}^{\top}\mathbf{w}\left(t\right)\right)\leq\frac{1}{N}N\exp\left(-\gamma\left(t\right)\right)\nonumber\\
&\Rightarrow\frac{1}{N}\exp\left(-\gamma\left(t\right)\right)\leq\exp\left(-\tilde{\gamma}\left(t\right)\right)\leq\exp\left(-\gamma\left(t\right)\right)\nonumber\\
&\Rightarrow\gamma\left(t\right)\leq\tilde{\gamma}\left(t\right)\leq\gamma\left(t\right)+\log\left(N\right)~.
\label{tilda}
\end{align}

From eq. \eqref{tilda} we have that $\lim_{t\rightarrow\infty}\frac{\tilde{\gamma}(t)}{\gamma(t)}=1$ and thus
\begin{align}
\lim_{\alpha\rightarrow\infty}\frac{\tilde{\gamma}(T_{\alpha})}{\gamma(T_{\alpha})}=1~.
\label{ratio1}
\end{align}

In addition, using $\mathbf{x}_{n}^{\top}\mathbf{w}\left(t\right)\leq x_{\max}\left\Vert \mathbf{w}\left(t\right)\right\Vert _{2}$ we derive a lower bound on $\|\w(t)\|_2$ as following:
\begin{align*}
\mathcal{L}\left(t\right) & =\frac{1}{N}\sum_{n=1}^{N}\exp\left(-\mathbf{x}_{n}^{\top}\mathbf{w}\left(t\right)\right)\\
 & \geq\frac{1}{N}N\exp\left(-x_{\max}\left\Vert \mathbf{w}\left(t\right)\right\Vert _{2}\right)\\
 & =\exp\left(-x_{\max}\left\Vert \mathbf{w}\left(t\right)\right\Vert _{2}\right)
\end{align*}
and thus
\[
x_{\max}\left\Vert \mathbf{w}\left(t\right)\right\Vert _{2}\geq\log\frac{1}{\mathcal{L}\left(t\right)}=\tilde{\gamma}\left(t\right)
\]
\begin{equation}
\Rightarrow\left\Vert \mathbf{w}\left(t\right)\right\Vert _{2}\geq\frac{\tilde{\gamma}\left(t\right)}{x_{\max}}~.
\label{eq: w_lower_bound}
\end{equation}

\paragraph{Conditions:} We restate \condref{assmp:intermediate_depthD} in terms of $\tilde{\gamma}$, i.e. we substitute $\epsilon=\exp(-\tilde{\gamma})$. We consider two cases:
\begin{condition}\label{assmp:intermediate_depthD_app}
For all $k\in\left[N\right]$ such that $\x_{k}^{\top}\hat{\w}>1$, and large enough $\alpha$,
there exists
$\tilde{\gamma}^{\star}\left(\alpha\right)=o\left(\alpha^{D}\right)$
and $\rho_{0}>1$ such that
$\forall\tilde{\gamma}\in\left[\tilde{\gamma}^{\star}\left(\alpha\right),\tilde{\gamma}\left(\alpha\right)\right]:\frac{\x_{k}^{\top}\w\left(\tilde{\gamma}\right)}{\gamma(\tilde{\gamma})}\geq\rho_{0}$.
\end{condition}

\begin{condition}\label{assmp:intermediate_depthD_app_richD2}
For all $k\in\left[N\right]$ such that $\x_{k}^{\top}\hat{\w}>1$, and large enough $\alpha$, there exists
$\tilde{\gamma}^{\star}\left(\alpha\right)=o\left(\alpha^{2}\log\frac{\tilde{\gamma}(\alpha)}{\alpha^2}\right)$
and $\rho_{0}>1$ such that
$\forall\tilde{\gamma}\in\left[\tilde{\gamma}^{\star}\left(\alpha\right),\tilde{\gamma}\left(\alpha\right)\right]:\frac{\x_{k}^{\top}\w\left(\tilde{\gamma}\right)}{\gamma(\tilde{\gamma})}\geq\rho_{0}$.
\end{condition}

We prove the intermediate regime for $D\geq2$ and the rich regime for $D>2$ under \condref{assmp:intermediate_depthD_app}.
To prove the rich regime for $D=2$ the weaker \condref{assmp:intermediate_depthD_app_richD2} will suffice.

\section{Auxiliary lemmas}
\label{auxiliary_lemmas}

\subsection{The case $D=2$}

\begin{lemma}
For $D=2$ and all $t$,
\begin{align}
\w\left(t\right)=2\alpha^{2}\sinh\left(4X\int_{0}^{t}\r(s)ds\right)
\label{w_sol}
\end{align}
and
\begin{align}
\frac{d\w(t)}{dt}=\frac{4}{N}\sqrt{\w^{2}\left(t\right)+4\alpha^{4}\mathbf{1}}\circ X\exp\left(-X^{\top}\w\left(t\right)\right)=A(t)X\r(t)
\label{w_ode}
\end{align}
where $A\left(t\right)=\diag\left(4\sqrt{\w^{2}\left(t\right)+4\alpha^{4}\mathbf{1}}\right)$.
\label{lem:w_ode}
\end{lemma}

\proof
The gradient flow dynamics in the parameters space is given by
\begin{align}
\dot{\u}\left(t\right) & =-\nabla_{\u}\mathcal{L}\left(\u(t)\right)=\frac{2}{N}\u\left(t\right)\circ\tilde{X}\exp\left(-\tilde{X}^{\top}\u^{2}\left(t\right)\right)\label{eq: u_flow}~.
\end{align}
It is easy to verify that the solution to eq. (\ref{eq: u_flow})
can be written as
\begin{equation}
\!\u\left(t\right)\!=\u\left(0\right)\,\circ\,\exp\left(\frac{2}{N}\tilde{X}\!\int_{0}^{t}\exp\left(-\tilde{X}^{\top}\u^{2}\left(s\right)\right)ds\right)\!=\alpha\exp\left(\frac{2}{N}\tilde{X}\!\int_{0}^{t}\exp\left(-\tilde{X}^{\top}\u^{2}\left(s\right)\right)ds\right).\label{eq: u_sol}
\end{equation}
From (\ref{eq: u_sol}) and $\w=\u_{+}^{2}-\u_{-}^{2}$ we get eq. (\ref{w_sol}). Taking
the derivative of eq. (\ref{w_sol}) we have
\begin{equation}
\dot{\w}\left(t\right)=\frac{8}{N}\alpha^{2}\cosh\left(4X\int_{0}^{t}\r(s)ds\right)\circ X\exp\left(-X^{\top}\w\left(t\right)\right).\label{eq: w_der}
\end{equation}
By combining eqs. (\ref{w_sol}) and (\ref{eq: w_der}) we get
\[
\dot{\w}\left(t\right)=\frac{8}{N}\alpha^{2}\cosh\left(\arcsinh\left(\frac{\w\left(t\right)}{2\alpha^{2}}\right)\right)\circ X\exp\left(-X^{\top}\w\left(t\right)\right).
\]
Since $\cosh\left(\arcsinh\left(x\right)\right)=\sqrt{x^{2}+1}$
we get eq. (\ref{w_ode}).
\qed

\begin{lemma}
\label{lem:w_upper_bound}
For $D=2$ and all $t$,
\[
\left\Vert \w\left(t\right)\right\Vert _{\infty}\leq2\alpha^{2}\sinh\left(\frac{\bar{x}}{2\gamma_{2}^{2}\alpha^{2}}\tilde{\gamma}\left(t\right)\right)\,.
\]
\end{lemma}

\proof
Note that
\[
\frac{d\mathcal{L}\left(t\right)}{dt}=\left(\nabla_{\w}\mathcal{L}\left(t\right)\right)^{\top}\frac{d\w\left(t\right)}{dt}=-\left(X\r\left(t\right)\right)^{\top}A\left(t\right)X\r\left(t\right)
\]
\[
\Rightarrow\frac{d\tilde{\gamma}\left(t\right)}{dt}=-\frac{1}{\mathcal{L}\left(t\right)}\frac{d\mathcal{L}\left(t\right)}{dt}=\frac{\left(X\r\left(t\right)\right)^{\top}A\left(t\right)X\r\left(t\right)}{\left\Vert \r\left(t\right)\right\Vert _{1}}\,.
\]
 From $A_{i,i}\left(t\right)\geq8\alpha^{2}$ we have
\begin{equation}
\frac{d\tilde{\gamma}\left(t\right)}{dt}\geq\frac{8\alpha^{2}\left\Vert X\r\left(t\right)\right\Vert _{2}^{2}}{\left\Vert \r\left(t\right)\right\Vert _{1}}\,.\label{eq:gamma_tilda_bound}
\end{equation}

From Lemma 2 of \cite{DBLP:conf/aistats/NacsonLGSSS19} we have that
\begin{equation}
\left\Vert X\r\left(t\right)\right\Vert _{2}\geq\gamma_{2}\left\Vert \r\left(t\right)\right\Vert _{1}\,.\label{eq:xr_bound}
\end{equation}
Combining eqs. (\ref{eq:gamma_tilda_bound}) and (\ref{eq:xr_bound})
we get
\begin{equation}
\frac{d\tilde{\gamma}\left(t\right)}{dt}\geq8\alpha^{2}\gamma_{2}^{2}\left\Vert \r\left(t\right)\right\Vert _{1}=8\alpha^{2}\gamma_{2}^{2}\exp\left(-\tilde{\gamma}\left(t\right)\right)\label{eq:gamma_tilda_bound1}~.
\end{equation}
We employ the dynamics equation $\dot{\w}\left(t\right)=\frac{4}{N}\sqrt{\w^{2}\left(t\right)+4\alpha^{4}\mathbf{1}}\circ X\exp\left(-X^{\top}\w\left(t\right)\right)$
and change variables $t\rightarrow\tilde{\gamma}\left(t\right)$.
Using eq. (\ref{eq:gamma_tilda_bound1}) we get that
\[
\left|\frac{dw_{i}\left(\tilde{\gamma}\right)}{d\tilde{\gamma}}\right|=\left|\frac{dw_{i}\left(t\right)}{dt}\frac{dt}{d\tilde{\gamma}}\right|\leq\left(\sqrt{w_{i}^{2}\left(\tilde{\gamma}\right)+4\alpha^{4}}\right)\left|\left[X\exp\left(-X^{\top}\w\left(\tilde{\gamma}\right)\right)\right]_{i}\right|\frac{1}{2N\alpha^{2}\gamma_{2}^{2}\exp\left(-\tilde{\gamma}\right)}\,.
\]
Using $\exp\left(-\x_{k}^{\top}\w\left(\tilde{\gamma}\right)\right)\leq N\exp\left(-\tilde{\gamma}\right)$
which follows from eq. \eqref{tilda} we get
\begin{align*}
\left|\frac{dw_{i}\left(\tilde{\gamma}\right)}{d\tilde{\gamma}}\right| & \leq\frac{\bar{x}_{i}}{2\alpha^{2}\gamma_{2}^{2}}\sqrt{w_{i}^{2}\left(\tilde{\gamma}\right)+4\alpha^{4}}
\end{align*}
and by the Grönwall's inequality we get the desired bound
\begin{align*}
\left|w_{i}\left(\tilde{\gamma}\right)\right|  \leq2\alpha^{2}\sinh\left(\frac{\bar{x}_{i}}{2\gamma_{2}^{2}\alpha^{2}}\tilde{\gamma}\right)
  \leq2\alpha^{2}\sinh\left(\frac{\bar{x}}{2\gamma_{2}^{2}\alpha^{2}}\tilde{\gamma}\right)\,.
\end{align*}
\qed

\subsection{The case $D>2$}

\begin{lemma}
\label{lem:w_D_dynamics}
For $D>2$ and all $t$,
\[
\w\left(t\right)=\alpha^{D}h_{D}\left(\alpha^{D-2}D\left(D-2\right)X\int_{0}^{t}\r\left(s\right)ds\right)
\]
and
\[
\frac{d\w\left(t\right)}{dt}=A_{D}\left(t\right)X\r\left(t\right)
\]
 where $A_{D}\left(t\right)=\diag\left(\alpha^{2D-2}D\left(D-2\right)h_{D}^{'}\left(h_{D}^{-1}\left(\frac{\w\left(t\right)}{\alpha^{D}}\right)\right)\right)$.
 \end{lemma}
\proof
The gradient flow dynamics in the parameters space is given by
\begin{align}
\dot{\u}\left(t\right) & =-\nabla_{\u}\mathcal{L}\left(\u(t)\right)=\frac{D}{N}\u^{D-1}\left(t\right)\circ\tilde{X}\exp\left(-\tilde{X}^{\top}\u^{D}\left(t\right)\right)\,.\label{eq: u_flow-1}
\end{align}
It is easy to verify that the solution to eq. (\ref{eq: u_flow-1})
is
\begin{align}
\u\left(t\right) & =\left(\u^{2-D}\left(0\right)-\frac{D\left(D-2\right)}{N}\tilde{X}\int_{0}^{t}\exp\left(-\tilde{X}^{\top}\u^{D}\left(s\right)\right)ds\right)^{-\frac{1}{D-2}}\nonumber \\
 & =\left(\alpha^{2-D}\mathbf{1}-\frac{D\left(D-2\right)}{N}\tilde{X}\int_{0}^{t}\exp\left(-\tilde{X}^{\top}\u^{D}\left(s\right)\right)ds\right)^{-\frac{1}{D-2}}\nonumber \\
 & =\alpha\left(\mathbf{1}-\frac{\alpha^{D-2}D\left(D-2\right)}{N}\tilde{X}\int_{0}^{t}\exp\left(-\tilde{X}^{\top}\u^{D}\left(s\right)\right)ds\right)^{-\frac{1}{D-2}}\,.\label{eq: u_sol-1}
\end{align}
From eq. (\ref{eq: u_sol-1}) and $\w=\u_{+}^{D}-\u_{-}^{D}$ we get
\begin{align}
\w\left(t\right)=\alpha^{D}\Bigg[\left(\mathbf{1}-\alpha^{D-2}D\left(D-2\right)X\int_{0}^{t}\r\left(s\right)ds\right)^{-\frac{D}{D-2}}\nonumber \\
-\left(\mathbf{1}+\alpha^{D-2}D\left(D-2\right)X\int_{0}^{t}\r\left(s\right)ds\right)^{-\frac{D}{D-2}}\Bigg]\,.
\label{eq: w_sol-1}
\end{align}
 As $u_{i}\left(t\right)\geq0$ for all $i$ (because $u_{i}\left(0\right)=\alpha>0$;
the gradient flow dynamics are continuous; and $u_{i}\left(t\right)=0\Rightarrow\dot{u}_{i}\left(t\right)=0$)
we get from eq. (\ref{eq: u_sol-1}) that
\begin{equation}
-1\leq\frac{\alpha^{D-2}D\left(D-2\right)}{N}X\int_{0}^{t}\exp\left(-\tilde{X}^{\top}\u^{D}\left(s\right)\right)ds\leq1\,.\label{eq:domain_bounds}
\end{equation}
Therefore we can write eq. (\ref{eq: w_sol-1}) as
\begin{equation}
\w\left(t\right)=\alpha^{D}h_{D}\left(\alpha^{D-2}D\left(D-2\right)X\int_{0}^{t}\r\left(s\right)ds\right)\label{eq:w_D}
\end{equation}
\[
\Rightarrow\alpha^{D-2}D\left(D-2\right)X\int_{0}^{t}\r\left(s\right)ds=h_{D}^{-1}\left(\frac{\w\left(t\right)}{\alpha^{D}}\right)\,.
\]
Taking the derivative of eq. (\ref{eq:w_D}) we get
\begin{align}
\dot{\w}\left(t\right) & =\alpha^{D}h_{D}^{'}\left(\alpha^{D-2}D\left(D-2\right)X\int_{0}^{t}\r\left(s\right)ds\right)\circ\left(\alpha^{D-2}D\left(D-2\right)X\r\left(t\right)\right)\nonumber \\
 & =\alpha^{2D-2}D\left(D-2\right)h_{D}^{'}\left(h_{D}^{-1}\left(\frac{\w\left(t\right)}{\alpha^{D}}\right)\right)\circ\left(X\r\left(t\right)\right)\,.\label{eq:w_ode_D}
\end{align}
\qed

\begin{lemma}
\label{lem:w_bound_D}
For $D>2$ and all $t$,
\[
\left\Vert \w\left(t\right)\right\Vert _{\infty}\leq\alpha^{D}h_{D}\left(\frac{\left(D-2\right)\bar{x}}{2D\gamma_{2}^{2}\alpha^{D}}\tilde{\gamma}\left(t\right)\right)\,.
\]
\end{lemma}
\proof
Note that
\[
\frac{d\mathcal{L}\left(t\right)}{dt}=\left(\nabla_{\w}\mathcal{L}\left(t\right)\right)^{\top}\frac{d\w\left(t\right)}{dt}=-\left(X\r\left(t\right)\right)^{\top}A_{D}\left(t\right)X\r\left(t\right)
\]
\begin{equation}
\Rightarrow\frac{d\tilde{\gamma}\left(t\right)}{dt}=-\frac{1}{\mathcal{L}\left(t\right)}\frac{d\mathcal{L}\left(t\right)}{dt}=\frac{\left(X\r\left(t\right)\right)^{\top}A_{D}\left(t\right)X\r\left(t\right)}{\left\Vert \r\left(t\right)\right\Vert _{1}}\,.\label{eq: gama_D}
\end{equation}
From eq. (\ref{eq:h_deriv}) we get a lower bound on the entries of
$A_{D}\left(t\right)$, $A_{D}\left(t\right)\geq2D^{2}\alpha^{2D-2}$.
Combining with eq. (\ref{eq: gama_D}) we get
\begin{equation}
\frac{d\tilde{\gamma}\left(t\right)}{dt}\geq\frac{2D^{2}\alpha^{2D-2}\left\Vert X\r\left(t\right)\right\Vert _{2}^{2}}{\left\Vert \r\left(t\right)\right\Vert _{1}}\,.\label{eq:gamma_tilda_bound-1}
\end{equation}

From eqs. (\ref{eq:gamma_tilda_bound-1}), (\ref{eq:xr_bound}) we
get
\begin{equation}
\frac{d\tilde{\gamma}\left(t\right)}{dt}\geq2D^{2}\alpha^{2D-2}\gamma_{2}^{2}\left\Vert \r\left(t\right)\right\Vert _{1}=2D^{2}\alpha^{2D-2}\gamma_{2}^{2}\exp\left(-\tilde{\gamma}\left(t\right)\right)\,.\label{eq: gama_bound_D}
\end{equation}
We employ the dynamics equation eq. (\ref{eq:w_ode_D}) and change
variables $t\rightarrow\tilde{\gamma}\left(t\right)$. Using eq. (\ref{eq: gama_bound_D})
we get that
\[
\left|\frac{dw_{i}\left(\tilde{\gamma}\right)}{d\tilde{\gamma}}\right|=\left|\frac{dw_{i}\left(t\right)}{dt}\frac{dt}{d\tilde{\gamma}}\right|\leq\alpha^{2D-2}D\left(D-2\right)h_{D}^{'}\left(h_{D}^{-1}\left(\frac{w_{i}\left(\tilde{\gamma}\right)}{\alpha^{D}}\right)\right)\frac{\left|\left[X\exp\left(-X^{\top}\w\left(\tilde{\gamma}\right)\right)\right]_{i}\right|}{2ND^{2}\alpha^{2D-2}\gamma_{2}^{2}\exp\left(-\tilde{\gamma}\right)}\,.
\]
Using $\exp\left(-\x_{k}^{\top}\w\left(\tilde{\gamma}\right)\right)\leq N\exp\left(-\tilde{\gamma}\right)$
which follows from eq. \eqref{tilda} we get
\begin{align*}
\left|\frac{dw_{i}\left(\tilde{\gamma}\right)}{d\tilde{\gamma}}\right| & \leq\frac{\left(D-2\right)\bar{x}_{i}}{2D\gamma_{2}^{2}}h_{D}^{'}\left(h_{D}^{-1}\left(\frac{w_{i}\left(\tilde{\gamma}\right)}{\alpha^{D}}\right)\right)
\end{align*}
and by the Grönwall's inequality we get the desired bound
\[
\left|w_{i}\left(\tilde{\gamma}\right)\right|\leq\alpha^{D}h_{D}\left(\frac{\left(D-2\right)\bar{x}_{i}}{2D\gamma_{2}^{2}\alpha^{D}}\tilde{\gamma}\right)\leq\alpha^{D}h_{D}\left(\frac{\left(D-2\right)\bar{x}}{2D\gamma_{2}^{2}\alpha^{D}}\tilde{\gamma}\right)\,.
\]
\qed

\section{Proof of \lemref{lemma:loss_bound}}
\label{loss_bound_proof}
We prove the loss bound for $D\geq 2$, any fixed $\alpha$, and  $\forall t$:
\[
\mathcal{L}\left(t\right)\leq\frac{1}{1+2D^{2}\alpha^{2D-2}\gamma_{2}^{2}t}.
\]

\proof
We employ the Grönwall's inequality. For $D=2$ from eq. \eqref{eq:gamma_tilda_bound1} we get
\begin{align}
\tilde{\gamma}\left(t\right)\geq\log\left(1+8\alpha^{2}\gamma_{2}^{2}t\right)
\label{r1}
\end{align}
and thus
\begin{align}
\mathcal{L}\left(t\right)\leq\frac{1}{1+8\alpha^{2}\gamma_{2}^{2}t}~.
\label{r2}
\end{align}
For $D>2$ from eq. \eqref{eq: gama_bound_D} we get
\begin{align}
\tilde{\gamma}\left(t\right)\geq\log\left(1+2D^{2}\alpha^{2D-2}\gamma_{2}^{2}t\right)
\label{r3}
\end{align}
and thus
\begin{align}
\mathcal{L}\left(t\right)\leq\frac{1}{1+2D^{2}\alpha^{2D-2}\gamma_{2}^{2}t}
~.
\label{r4}
\end{align}
Note that by substituting $D=2$ in eq. \eqref{r4} we get eq. \eqref{r2}, so \eqref{r4} is correct for $D\geq 2$.
\qed

\section{\condref{assmp:intermediate_depthD} holds for the linearized model}
\label{linearized_condition}

We show that \condref{assmp:intermediate_depthD_app}, which is equivalent to \condref{assmp:intermediate_depthD}, holds for the linearized model.
The linearized model is
\[
\bar{f}\left(\bar{\mathbf{u}},\mathbf{x}\right)=f\left(\mathbf{u}\left(0\right),\mathbf{x}\right)+\nabla_{\mathbf{u}}^{\top}f\left(\mathbf{u}\left(0\right),\mathbf{x}\right)\bar{\mathbf{u}}\,.
\]
For the diagonal linear network $f\left(\mathbf{u},\mathbf{x}\right)=\mathbf{x}^{\top}\left(\mathbf{u}_{+}^{D}-\mathbf{u}_{-}^{D}\right)$,
where $\mathbf{u}=\left[\begin{array}{c}
\mathbf{u}_{+}\\
\mathbf{u}_{-}
\end{array}\right]\in\mathbb{R}^{2d}$. We consider the initialization $\mathbf{u}\left(0\right)=\bar{\mathbf{u}}\left(0\right)=\alpha\mathbf{1}$,
thus $f\left(\mathbf{u}\left(0\right),\mathbf{x}\right)=0$, $\nabla_{\mathbf{u}}f\left(\mathbf{u}\left(0\right),\mathbf{x}\right)=D\alpha{}^{D-1}\left[\begin{array}{c}
\mathbf{x}\\
-\mathbf{x}
\end{array}\right]$ and
\[
\bar{f}\left(\bar{\mathbf{u}},\mathbf{x}\right)=D\alpha{}^{D-1}\mathbf{x}^{\top}\left(\bar{\mathbf{u}}_{+}-\bar{\mathbf{u}}_{-}\right)\,.
\]
 Let $\bar{\mathbf{w}}=D\alpha{}^{D-1}\left(\bar{\mathbf{u}}_{+}-\bar{\mathbf{u}}_{-}\right)$.
Then $\bar{f}\left(\bar{\mathbf{w}},\mathbf{x}\right)=\bar{\mathbf{w}}^{\top}\mathbf{x}$.
We consider gradient flow $\frac{d\bar{\mathbf{u}}\left(t\right)}{dt}=-\nabla\bar{\mathcal{L}}\left(\bar{\mathbf{u}}\left(t\right)\right)$
where
\[
\bar{\mathcal{L}}\left(\bar{\mathbf{u}}\left(t\right)\right)=\frac{1}{N}\sum_{n=1}^{N}\exp\left(-\bar{f}\left(\bar{\mathbf{u}}\left(t\right),\mathbf{x}_{n}\right)\right)\,.
\]
 Thus
\[
\frac{d\bar{\mathbf{u}}_{+}\left(t\right)}{dt}=\frac{1}{N}D\alpha{}^{D-1}\sum_{n=1}^{N}\exp\left(-D\alpha{}^{D-1}\mathbf{x}_{n}^{\top}\left(\bar{\mathbf{u}}_{+}-\bar{\mathbf{u}}_{-}\right)\right)\mathbf{x}_{n}
\]
\[
\frac{d\bar{\mathbf{u}}_{-}\left(t\right)}{dt}=-\frac{1}{N}D\alpha{}^{D-1}\sum_{n=1}^{N}\exp\left(-D\alpha{}^{D-1}\mathbf{x}_{n}^{\top}\left(\bar{\mathbf{u}}_{+}-\bar{\mathbf{u}}_{-}\right)\right)\mathbf{x}_{n}
\]
 and
\begin{align}
\frac{d\bar{\mathbf{w}}\left(t\right)}{dt} & =D\alpha{}^{D-1}\left(\frac{d\bar{\mathbf{u}}_{+}\left(t\right)}{dt}-\frac{d\bar{\mathbf{u}}_{-}\left(t\right)}{dt}\right)\nonumber \\
 & =\frac{2}{N}D^{2}\alpha{}^{2D-2}\sum_{n=1}^{N}\exp\left(-\mathbf{x}_{n}^{\top}\bar{\mathbf{w}}\left(t\right)\right)\mathbf{x}_{n}\,.\label{dw_bar}
\end{align}
It follows that
\begin{align*}
\frac{d\bar{\mathcal{L}}\left(t\right)}{dt} & =\left(\nabla_{\mathbf{\bar{\mathbf{w}}}}\bar{\mathcal{L}}\left(t\right)\right)^{\top}\frac{d\bar{\mathbf{w}}\left(t\right)}{dt}\\
 & =\left(-\frac{1}{N}\sum_{n=1}^{N}\exp\left(-\mathbf{x}_{n}^{\top}\bar{\mathbf{w}}\left(t\right)\right)\mathbf{x}_{n}\right)^{\top}\left(\frac{2}{N}D^{2}\alpha{}^{2D-2}\sum_{n=1}^{N}\exp\left(-\mathbf{x}_{n}^{\top}\bar{\mathbf{w}}\left(t\right)\right)\mathbf{x}_{n}\right)\\
 & =-2D^{2}\alpha{}^{2D-2}\left\Vert \frac{1}{N}\sum_{n=1}^{N}\exp\left(-\mathbf{x}_{n}^{\top}\bar{\mathbf{w}}\left(t\right)\right)\mathbf{x}_{n}\right\Vert _{2}^{2}\,.
\end{align*}
Let $\bar{\tilde{\gamma}}\left(t\right)=\log\frac{1}{\bar{\mathcal{L}}\left(t\right)}$.
Then
\begin{equation}
\frac{d\bar{\tilde{\gamma}}\left(t\right)}{dt}=-\frac{1}{\bar{\mathcal{L}}\left(t\right)}\frac{d\bar{\mathcal{L}}\left(t\right)}{dt}=\frac{1}{\bar{\mathcal{L}}\left(t\right)}2D^{2}\alpha{}^{2D-2}\left\Vert \frac{1}{N}\sum_{n=1}^{N}\exp\left(-\mathbf{x}_{n}^{\top}\bar{\mathbf{w}}\left(t\right)\right)\mathbf{x}_{n}\right\Vert _{2}^{2}\,.\label{dgamma_bar}
\end{equation}
From Lemma 2 of \cite{DBLP:conf/aistats/NacsonLGSSS19} we know that
\begin{equation}
\left\Vert \frac{1}{N}\sum_{n=1}^{N}\exp\left(-\mathbf{x}_{n}^{\top}\bar{\mathbf{w}}\left(t\right)\right)\mathbf{x}_{n}\right\Vert _{2}^{2}\geq\gamma_{2}^{2}\left(\frac{1}{N}\sum_{n=1}^{N}\exp\left(-\mathbf{x}_{n}^{\top}\bar{\mathbf{w}}\left(t\right)\right)\right)^{2}=\gamma_{2}^{2}\bar{\mathcal{L}}^{2}\left(t\right)\,.\label{z_ub}
\end{equation}
Combining eqs. (\ref{dgamma_bar}) and (\ref{z_ub}) we get
\begin{equation}
\frac{d\bar{\tilde{\gamma}}\left(t\right)}{dt}\geq2D^{2}\alpha{}^{2D-2}\gamma_{2}^{2}\bar{\mathcal{L}}\left(t\right)=2D^{2}\alpha{}^{2D-2}\gamma_{2}^{2}\exp\left(-\bar{\tilde{\gamma}}\left(t\right)\right)\,.\label{gamma_tb_lb}
\end{equation}
In addition,
\begin{equation}
\left\Vert \frac{1}{N}\sum_{n=1}^{N}\exp\left(-\mathbf{x}_{n}^{\top}\bar{\mathbf{w}}\left(t\right)\right)\mathbf{x}_{n}\right\Vert _{2}\leq\frac{1}{N}\sum_{n=1}^{N}\exp\left(-\mathbf{x}_{n}^{\top}\bar{\mathbf{w}}\left(t\right)\right)\left\Vert \mathbf{x}_{n}\right\Vert _{2}\leq x_{\max}\bar{\mathcal{L}}\left(t\right)\,.\label{z_lb}
\end{equation}
Combining eqs. (\ref{dgamma_bar}) and (\ref{z_lb}) we get
\[
\frac{d\bar{\tilde{\gamma}}\left(t\right)}{dt}\leq2D^{2}\alpha{}^{2D-2}x_{\max}^{2}\bar{\mathcal{L}}\left(t\right)=2D^{2}\alpha{}^{2D-2}x_{\max}^{2}\exp\left(-\bar{\tilde{\gamma}}\left(t\right)\right)
\]
and by the Grönwall\textquoteright s inequality we get
\[
\bar{\tilde{\gamma}}\left(t\right)\leq\log\left(1+2D^{2}\alpha{}^{2D-2}x_{\max}^{2}t\right)
\]
\begin{equation}
\Rightarrow t\geq\frac{\exp\left(\bar{\tilde{\gamma}}\right)-1}{2D^{2}\alpha{}^{2D-2}x_{\max}^{2}}\,.\label{t_bound}
\end{equation}
The $\ell_{2}$ max-margin solution is $\mathbf{w}_{\ell_{2}}=\sum_{n\in S_{2}}\nu_{n}\mathbf{x}_{n}$
where $S_{2}$ denotes the set of support vectors of $\mathbf{w}_{\ell_{2}}$.
Let $\tilde{\mathbf{w}}$ be a vector that satisfies $\exp\left(-\mathbf{x}_{n}^{\top}\tilde{\mathbf{w}}\right)=\nu_{n}$
for $n\in S_{2}$. Such $\tilde{\mathbf{w}}$ exists for almost all
datasets, where the support vectors of $\mathbf{w}_{\ell_{2}}$ are associated with positive dual variables $\nu_{n}$ \citep{soudry2018implicit}. Let
\begin{equation}
\mathbf{\kappa}\left(t\right)=\bar{\mathbf{w}}\left(t\right)-\log\left(\frac{2}{N}D^{2}\alpha{}^{2D-2}t\right)\mathbf{w}_{\ell_{2}}-\tilde{\mathbf{w}}\,.\label{kappa_def}
\end{equation}
Then
\[
\frac{d\mathbf{\kappa}\left(t\right)}{dt}=\frac{d\bar{\mathbf{w}}\left(t\right)}{dt}-\frac{1}{t}\mathbf{w}_{\ell_{2}}
\]
and thus
\begin{align}
\frac{1}{2}\frac{d}{dt}\left\Vert \mathbf{\kappa}\left(t\right)\right\Vert _{2}^{2}= & \left(\frac{d\mathbf{\kappa}\left(t\right)}{dt}\right)^{\top}\mathbf{\kappa}\left(t\right)\nonumber \\
= & \left(\frac{d\bar{\mathbf{w}}\left(t\right)}{dt}-\frac{1}{t}\mathbf{w}_{\ell_{2}}\right)^{\top}\mathbf{\kappa}\left(t\right)\nonumber \\
\overset{\eqref{dw_bar}}{=} & \frac{2}{N}D^{2}\alpha{}^{2D-2}\sum_{n=1}^{N}\exp\left(-\mathbf{x}_{n}^{\top}\bar{\mathbf{w}}\left(t\right)\right)\mathbf{x}_{n}^{\top}\mathbf{\kappa}\left(t\right)-\frac{1}{t}\mathbf{w}_{\ell_{2}}^{\top}\mathbf{\kappa}\left(t\right)\nonumber \\
= & \left[\frac{2}{N}D^{2}\alpha{}^{2D-2}\sum_{n\in S_{2}}\exp\left(-\mathbf{x}_{n}^{\top}\bar{\mathbf{w}}\left(t\right)\right)\mathbf{x}_{n}^{\top}\mathbf{\kappa}\left(t\right)-\frac{1}{t}\mathbf{w}_{\ell_{2}}^{\top}\mathbf{\kappa}\left(t\right)\right]\nonumber \\
 & +\left[\frac{2}{N}D^{2}\alpha{}^{2D-2}\sum_{n\notin S_{2}}\exp\left(-\mathbf{x}_{n}^{\top}\bar{\mathbf{w}}\left(t\right)\right)\mathbf{x}_{n}^{\top}\mathbf{\kappa}\left(t\right)\right]\label{dkappa}
\end{align}
For $n\in S_{2}$ we have that $\mathbf{x}_{n}^{\top}\mathbf{w}_{\ell_{2}}=1$,
thus
\begin{align*}
\exp\left(-\mathbf{x}_{n}^{\top}\bar{\mathbf{w}}\left(t\right)\right) & =\exp\left(-\mathbf{x}_{n}^{\top}\left(\log\left(\frac{2}{N}D^{2}\alpha{}^{2D-2}t\right)\mathbf{w}_{\ell_{2}}+\tilde{\mathbf{w}}+\mathbf{\kappa}\left(t\right)\right)\right)\\
 & =\frac{1}{\frac{2}{N}D^{2}\alpha{}^{2D-2}t}\exp\left(-\mathbf{x}_{n}^{\top}\tilde{\mathbf{w}}\right)\exp\left(-\mathbf{x}_{n}^{\top}\mathbf{\kappa}\left(t\right)\right)\\
 & =\frac{1}{\frac{2}{N}D^{2}\alpha{}^{2D-2}t}\nu_{n}\exp\left(-\mathbf{x}_{n}^{\top}\mathbf{\kappa}\left(t\right)\right)
\end{align*}
and the first bracketed term in eq. (\ref{dkappa}) can be written
as
\begin{align}
 & \frac{2}{N}D^{2}\alpha{}^{2D-2}\sum_{n\in S_{2}}\frac{1}{\frac{2}{N}D^{2}\alpha{}^{2D-2}t}\nu_{n}\exp\left(-\mathbf{x}_{n}^{\top}\mathbf{\kappa}\left(t\right)\right)\mathbf{x}_{n}^{\top}\mathbf{\kappa}\left(t\right)-\frac{1}{t}\sum_{n\in S_{2}}\nu_{n}\mathbf{x}_{n}^{\top}\mathbf{\kappa}\left(t\right)\nonumber \\
= & \frac{1}{t}\sum_{n\in S_{2}}\left[\nu_{n}\left(\exp\left(-\mathbf{x}_{n}^{\top}\mathbf{\kappa}\left(t\right)\right)-1\right)\mathbf{x}_{n}^{\top}\mathbf{\kappa}\left(t\right)\right]\nonumber \\
\leq & 0\label{dkappa_bound1}
\end{align}
since $\left(e^{-z}-1\right)z\leq0$ for all $z$.

Let $\theta=\min_{n\notin S_{2}}\left(\mathbf{x}_{n}^{\top}\mathbf{w}_{\ell_{2}}\right)>1$
and $c_{1}=\max_{n\notin S_{2}}\exp\left(-\mathbf{x}_{n}^{\top}\tilde{\mathbf{w}}\right)$.
For $n\notin S_{2}$ we have that
\begin{align*}
\exp\left(-\mathbf{x}_{n}^{\top}\bar{\mathbf{w}}\left(t\right)\right) & =\exp\left(-\mathbf{x}_{n}^{\top}\left(\log\left(\frac{2}{N}D^{2}\alpha{}^{2D-2}t\right)\mathbf{w}_{\ell_{2}}+\tilde{\mathbf{w}}+\mathbf{\kappa}\left(t\right)\right)\right)\\
 & \leq\frac{1}{\left(\frac{2}{N}D^{2}\alpha{}^{2D-2}t\right)^{\theta}}\exp\left(-\mathbf{x}_{n}^{\top}\tilde{\mathbf{w}}\right)\exp\left(-\mathbf{x}_{n}^{\top}\mathbf{\kappa}\left(t\right)\right)\\
 & \leq\frac{c_{1}}{\left(\frac{2}{N}D^{2}\alpha{}^{2D-2}t\right)^{\theta}}\exp\left(-\mathbf{x}_{n}^{\top}\mathbf{\kappa}\left(t\right)\right)
\end{align*}
and thus the second bracketed term in eq. (\ref{dkappa}) can be bounded
as following
\begin{align}
 & \frac{2}{N}D^{2}\alpha{}^{2D-2}\sum_{n\notin S_{2}}\exp\left(-\mathbf{x}_{n}^{\top}\bar{\mathbf{w}}\left(t\right)\right)\mathbf{x}_{n}^{\top}\mathbf{\kappa}\left(t\right)\nonumber \\
\leq & \frac{\frac{2}{N}D^{2}\alpha{}^{2D-2}c_{1}}{\left(\frac{2}{N}D^{2}\alpha{}^{2D-2}t\right)^{\theta}}\sum_{n\notin S_{2}}\exp\left(-\mathbf{x}_{n}^{\top}\mathbf{\kappa}\left(t\right)\right)\mathbf{x}_{n}^{\top}\mathbf{\kappa}\left(t\right)\nonumber \\
\leq & \frac{2D^{2}\alpha{}^{2D-2}c_{1}}{\left(\frac{2}{N}D^{2}\alpha{}^{2D-2}t\right)^{\theta}}\label{dkappa_bound2}
\end{align}
since $e^{-z}z\leq1$ for all $z$. Substituting eqs. (\ref{dkappa_bound1})
and (\ref{dkappa_bound2}) in eq. (\ref{dkappa}) we get
\[
\frac{1}{2}\frac{d}{dt}\left\Vert \mathbf{\kappa}\left(t\right)\right\Vert _{2}^{2}\leq\frac{2D^{2}\alpha{}^{2D-2}c_{1}}{\left(\frac{2}{N}D^{2}\alpha{}^{2D-2}t\right)^{\theta}}\,.
\]
Using (\ref{t_bound}) we get
\[
\frac{1}{2}\frac{d}{dt}\left\Vert \mathbf{\kappa}\left(t\right)\right\Vert _{2}^{2}\leq\frac{2D^{2}\alpha{}^{2D-2}c_{1}}{\left(\frac{2}{N}D^{2}\alpha{}^{2D-2}\frac{\exp\left(\bar{\tilde{\gamma}}\left(t\right)\right)-1}{2D^{2}\alpha{}^{2D-2}x_{\max}^{2}}\right)^{\theta}}=\frac{2D^{2}\alpha{}^{2D-2}c_{1}}{\left(\frac{1}{N}\frac{\exp\left(\bar{\tilde{\gamma}}\left(t\right)\right)-1}{x_{\max}^{2}}\right)^{\theta}}~.
\]
 We change variables $t\rightarrow\bar{\tilde{\gamma}}$ and get
\begin{align*}
\frac{1}{2}\frac{d}{d\bar{\tilde{\gamma}}}\left\Vert \mathbf{\kappa}\left(\bar{\tilde{\gamma}}\right)\right\Vert _{2}^{2} & =\frac{1}{2}\frac{d}{dt}\left\Vert \mathbf{\kappa}\left(\bar{\tilde{\gamma}}\left(t\right)\right)\right\Vert _{2}^{2}\frac{dt}{d\bar{\tilde{\gamma}}}\\
 & \overset{\eqref{gamma_tb_lb}}{\leq}\frac{2D^{2}\alpha{}^{2D-2}c_{1}}{\left(\frac{1}{N}\frac{\exp\left(\bar{\tilde{\gamma}}\right)-1}{x_{\max}^{2}}\right)^{\theta}}\frac{1}{2D^{2}\alpha{}^{2D-2}\gamma_{2}^{2}\exp\left(-\bar{\tilde{\gamma}}\right)}\\
 & =\frac{c_{1}\exp\left(-\left(\theta-1\right)\bar{\tilde{\gamma}}\right)}{\gamma_{2}^{2}\left(\frac{1}{N}\frac{1-\exp\left(-\bar{\tilde{\gamma}}\right)}{x_{\max}^{2}}\right)^{\theta}}\\
 & \leq C\frac{\exp\left(-\left(\theta-1\right)\bar{\tilde{\gamma}}\right)}{\left(1-\exp\left(-\bar{\tilde{\gamma}}\right)\right)^{\theta}}
\end{align*}
where $C$ is a constant. Integrating we have that for all $\bar{\tilde{\gamma}}_{0}>0,\bar{\tilde{\gamma}}>\bar{\tilde{\gamma}}_{0}$
\begin{align*}
\left\Vert \mathbf{\kappa}\left(\bar{\tilde{\gamma}}\right)\right\Vert _{2}^{2}-\left\Vert \mathbf{\kappa}\left(\bar{\tilde{\gamma}}_{0}\right)\right\Vert _{2}^{2} & \leq C\int_{\bar{\tilde{\gamma}}_{0}}^{\bar{\tilde{\gamma}}}\frac{\exp\left(-\left(\theta-1\right)\bar{\tilde{\gamma}}_{1}\right)}{\left(1-\exp\left(-\bar{\tilde{\gamma}}_{1}\right)\right)^{\theta}}d\bar{\tilde{\gamma}}_{1}\\
 & \leq C\int_{\bar{\tilde{\gamma}}_{0}}^{\infty}\frac{\exp\left(-\left(\theta-1\right)\bar{\tilde{\gamma}}_{1}\right)}{\left(1-\exp\left(-\bar{\tilde{\gamma}}_{1}\right)\right)^{\theta}}d\bar{\tilde{\gamma}}_{1}\\
 & =C\frac{1}{\left(\theta-1\right)\left(\exp\left(\bar{\tilde{\gamma}}_{0}\right)-1\right)^{\theta-1}}
\end{align*}
and thus
\begin{equation}
\left\Vert \mathbf{\kappa}\left(\bar{\tilde{\gamma}}\right)\right\Vert _{2}\leq C'\label{kappa_bound}
\end{equation}
where $C'$ is a constant. Finally, for $k\notin S_{2}$ we have that
$\mathbf{x}_{k}^{\top}\mathbf{w}_{\ell_{2}}\geq\theta>1$ and thus
\begin{align*}
\frac{\mathbf{x}_{k}^{\top}\bar{\mathbf{w}}\left(\bar{\tilde{\gamma}}\right)}{\bar{\tilde{\gamma}}} & \overset{\eqref{kappa_def}}{=}\frac{\mathbf{x}_{k}^{\top}\left(\log\left(\frac{2}{N}D^{2}\alpha{}^{2D-2}t\right)\mathbf{w}_{\ell_{2}}+\tilde{\mathbf{w}}+\mathbf{\kappa}\left(\bar{\tilde{\gamma}}\right)\right)}{\bar{\tilde{\gamma}}}\\
 & \overset{\eqref{t_bound}}{\geq}\frac{\log\left(\frac{2}{N}D^{2}\alpha{}^{2D-2}\frac{\exp\left(\bar{\tilde{\gamma}}\right)-1}{2D^{2}\alpha{}^{2D-2}x_{\max}^{2}}\right)\mathbf{x}_{k}^{\top}\mathbf{w}_{\ell_{2}}+\mathbf{x}_{k}^{\top}\tilde{\mathbf{w}}+\mathbf{x}_{k}^{\top}\mathbf{\kappa}\left(\bar{\tilde{\gamma}}\right)}{\bar{\tilde{\gamma}}}\\
 & \overset{\eqref{kappa_bound}}{\geq}\frac{\log\left(\frac{\exp\left(\bar{\tilde{\gamma}}\right)-1}{Nx_{\max}^{2}}\right)\theta-\log c_{1}-x_{\max}C'}{\bar{\tilde{\gamma}}}\\
 & =\frac{\log\left(\frac{\exp\left(\bar{\tilde{\gamma}}\right)-1}{Nx_{\max}^{2}}\right)}{\bar{\tilde{\gamma}}}\theta-\frac{\log c_{1}+x_{\max}C'}{\bar{\tilde{\gamma}}}~.
\end{align*}
Note that $\frac{\log\left(\frac{\exp\left(\bar{\tilde{\gamma}}\right)-1}{Nx_{\max}^{2}}\right)}{\bar{\tilde{\gamma}}}$
is monotonically increasing and
\[
\frac{\log\left(\frac{\exp\left(\bar{\tilde{\gamma}}\right)-1}{Nx_{\max}^{2}}\right)}{\bar{\tilde{\gamma}}}\overset{\bar{\tilde{\gamma}}\rightarrow\infty}{\rightarrow}1\,.
\]
Therefore there exists $\bar{\tilde{\gamma}}^{\star}$ (independent
of $\alpha$!) such that for $\tilde{\gamma}\geq\bar{\tilde{\gamma}}^{\star}$
\[
\frac{\log\left(\frac{\exp\left(\bar{\tilde{\gamma}}\right)-1}{Nx_{\max}^{2}}\right)}{\bar{\tilde{\gamma}}}\geq\frac{3\theta+1}{4\theta}
\]
 and
\[
\frac{\log c_{1}+x_{\max}C'}{\bar{\tilde{\gamma}}}\leq\frac{\theta-1}{4}\,.
\]
It follows that for $\tilde{\gamma}\geq\bar{\tilde{\gamma}}^{\star}$
\[
\frac{\mathbf{x}_{k}^{\top}\bar{\mathbf{w}}\left(\bar{\tilde{\gamma}}\right)}{\bar{\tilde{\gamma}}}\geq\frac{3\theta+1}{4\theta}\theta-\frac{\theta-1}{4}=\frac{\theta+1}{2}\doteq\rho_{0}>1~.
\]
From $\bar{\tilde{\gamma}}\geq\bar{\gamma}$ (where $\bar{\gamma}\left(t\right)=\min_{n}\left(\mathbf{x}_{n}^{\top}\bar{\mathbf{w}}\left(t\right)\right))$
we get
\[
\frac{\mathbf{x}_{k}^{\top}\bar{\mathbf{w}}\left(\bar{\tilde{\gamma}}\right)}{\bar{\gamma}}\geq\frac{\mathbf{x}_{k}^{\top}\bar{\mathbf{w}}\left(\bar{\tilde{\gamma}}\right)}{\bar{\tilde{\gamma}}}\geq\rho_{0}>1
\]
 for $\tilde{\gamma}\geq\bar{\tilde{\gamma}}^{\star}=o\left(\alpha^{D}\right)$
since $\bar{\tilde{\gamma}}^{\star}$ is independent of $\alpha$.

\section{Proofs for $D=2$}
\label{proofs_D2}

\subsection{Kernel Regime Proof}
\label{l2_limit_proof}

\begin{theorem}[\thmref{thm:l2_limit} for $D=2$]
\label{thm:l2_limit_app_D2}
For $D=2$, if $\tilde{\gamma}(\alpha)=o(\alpha^2)$
then
\[
\hat{\w}=\argmin_{\w}\left\Vert \w\right\Vert _{2}\,\,\,\,\mathrm{s.t.}\,\,\forall n:\,\x_{n}^{\top}\w\geq1\,.
\]
\end{theorem}

\proof
We show convergence of the $\ell_{2}$ margin $\gamma_{2}\left(T_{\alpha}\right)=\frac{\min_{n}\left(\x_{n}^{\top}\w\left(T_{\alpha}\right)\right)}{\left\Vert \w\left(T_{\alpha}\right)\right\Vert _{2}}$
to the max margin $\gamma_{2}$ when $\alpha\rightarrow\infty$.
Note that by definition $\gamma_{2}\left(T_{\alpha}\right)\leq\gamma_{2}$.
Next we show that $\gamma_{2}\left(T_{\alpha}\right)\geq\gamma_{2}$
when $\alpha\rightarrow\infty$. From eq. \eqref{tilda} we have that
\begin{align}
\gamma_{2}\left(t\right)=\frac{\min_{n}\left(\x_{n}^{\top}\w\left(t\right)\right)}{\left\Vert \w\left(t\right)\right\Vert _{2}}\geq\frac{\tilde{\gamma}\left(t\right)-\log(N)}{\left\Vert \w\left(t\right)\right\Vert _{2}}\,.
\label{l2_margin_D2}
\end{align}
In order to lower bound $\gamma_{2}\left(t\right)$ we derive a lower
bound on $\tilde{\gamma}\left(t\right)$ and an upper bound on $\left\Vert \w\left(t\right)\right\Vert _{2}$.
\paragraph{Lower bound on $\tilde{\gamma}\left(t\right)$:}
Combining (\ref{eq:gamma_tilda_bound}) and (\ref{eq:xr_bound}) we
get
\[
\frac{d\tilde{\gamma}\left(t\right)}{dt}\geq8\alpha^{2}\gamma_{2}\left\Vert X\r\left(t\right)\right\Vert _{2}
\]
\begin{equation}
\Rightarrow\tilde{\gamma}\left(t\right)\geq8\alpha^{2}\gamma_{2}\int_{0}^{t}\left\Vert X\r\left(\tau\right)\right\Vert _{2}d\tau\,.\label{eq: gamma2}
\end{equation}

\paragraph{Upper bound on $\left\Vert \w\left(t\right)\right\Vert _{2}$:}
We decompose $\dot{\w}\left(t\right)$ to two terms:
\begin{align*}
\dot{\w}\left(t\right) & =4\sqrt{\w^{2}\left(t\right)+4\alpha^{4}\mathbf{1}}\circ X\r\left(t\right)\\
 & =4\left(\sqrt{\w^{2}\left(t\right)+4\alpha^{4}\mathbf{1}}-2\alpha^{2}\mathbf{1}\right)\circ X\r\left(t\right)+8\alpha^{2}X\r\left(t\right)
\end{align*}
\[
\Rightarrow\left\Vert \dot{\w}\left(t\right)\right\Vert _{2}\leq4\left\Vert \left(\sqrt{\w^{2}\left(t\right)+4\alpha^{4}\mathbf{1}}-2\alpha^{2}\mathbf{1}\right)\circ X\r\left(t\right)\right\Vert _{2}+8\alpha^{2}\left\Vert X\r\left(t\right)\right\Vert _{2}
\]
\begin{equation}
\Rightarrow\left\Vert \w\left(t\right)\right\Vert _{2}\leq4\int_{0}^{t}\left\Vert \left(\sqrt{\w^{2}\left(\tau\right)+4\alpha^{4}\mathbf{1}}-2\alpha^{2}\mathbf{1}\right)\circ X\r\left(\tau\right)\right\Vert _{2}d\tau+8\alpha^{2}\int_{0}^{t}\left\Vert X\r\left(\tau\right)\right\Vert _{2}d\tau\,.\label{eq: w_b}
\end{equation}
 Let $v\left(t\right)=\left\Vert \left(\sqrt{\w^{2}\left(t\right)+4\alpha^{4}\mathbf{1}}-2\alpha^{2}\mathbf{1}\right)\circ X\r\left(t\right)\right\Vert _{2}$.
Then
\begin{align*}
v\left(t\right) & \leq\left\Vert \sqrt{\w^{2}\left(t\right)+4\alpha^{4}\mathbf{1}}-2\alpha^{2}\mathbf{1}\right\Vert _{\infty}x_{\max}\left\Vert \r\left(t\right)\right\Vert _{1}\\
 & =\left\Vert \sqrt{\w^{2}\left(t\right)+4\alpha^{4}\mathbf{1}}-2\alpha^{2}\mathbf{1}\right\Vert _{\infty}x_{\max}\exp\left(-\tilde{\gamma}\left(t\right)\right)\,.
\end{align*}
Using \lemref{lem:w_upper_bound} we get
\begin{align*}
v\left(t\right) & \leq\left(2\alpha^{2}\sqrt{\sinh^{2}\left(\frac{\bar{x}}{2\alpha^{2}\gamma_{2}^{2}}\tilde{\gamma}\left(t\right)\right)+1}-2\alpha^{2}\right)x_{\max}\exp\left(-\tilde{\gamma}\left(t\right)\right)\\
 & =2\alpha^{2}\left[\cosh\left(\frac{\bar{x}}{2\alpha^{2}\gamma_{2}^{2}}\tilde{\gamma}\left(t\right)\right)-1\right]x_{\max}\exp\left(-\tilde{\gamma}\left(t\right)\right)\,.
\end{align*}
We are interested in bounding $\int_{0}^{t}v\left(\tau\right)d\tau$.
We change variables $t\rightarrow\tilde{\gamma}\left(t\right)$ and
proceed using (\ref{eq:gamma_tilda_bound1}),
\begin{align}
\int_{0}^{t}v\left(\tau\right)d\tau & \leq\int_{0}^{\tilde{\gamma}\left(t\right)}2\alpha^{2}\left[\cosh\left(\frac{\bar{x}}{2\alpha^{2}\gamma_{2}^{2}}\tilde{\gamma}\right)-1\right]x_{\max}\exp\left(-\tilde{\gamma}\right)\frac{1}{8\alpha^{2}\gamma_{2}^{2}\exp\left(-\tilde{\gamma}\right)}d\tilde{\gamma}\nonumber \\
 & =\frac{x_{\max}}{4\gamma_{2}^{2}}\int_{0}^{\tilde{\gamma}\left(t\right)}\left[\cosh\left(\frac{\bar{x}}{2\alpha^{2}\gamma_{2}^{2}}\tilde{\gamma}\right)-1\right]d\tilde{\gamma}\nonumber \\
 & =\frac{x_{\max}}{4\gamma_{2}^{2}}\left[\frac{2\alpha^{2}\gamma_{2}^{2}}{\bar{x}}\sinh\left(\frac{\bar{x}}{2\alpha^{2}\gamma_{2}^{2}}\tilde{\gamma}\left(t\right)\right)-\tilde{\gamma}\left(t\right)\right]~.\label{eq: int_v}
\end{align}
Plugging eqs. (\ref{eq: int_v}) in (\ref{eq: w_b}) we get
\begin{align}
\left\Vert \w\left(t\right)\right\Vert _{2}\leq\frac{x_{\max}}{\gamma_{2}^{2}}\left[\frac{2\alpha^{2}\gamma_{2}^{2}}{\bar{x}}\sinh\left(\frac{\bar{x}}{2\alpha^{2}\gamma_{2}^{2}}\tilde{\gamma}\left(t\right)\right)-\tilde{\gamma}\left(t\right)\right]+8\alpha^{2}\int_{0}^{t}\left\Vert X\r\left(\tau\right)\right\Vert _{2}d\tau\,.
\label{a1}
\end{align}
\paragraph{Putting things together:}
From eqs. \eqref{l2_margin_D2} and \eqref{eq: w_lower_bound} we have
\begin{align}
\gamma_{2}\left(t\right)\geq\frac{\tilde{\gamma}\left(t\right)-\log(N)}{\left\Vert \w\left(t\right)\right\Vert _{2}}\geq \frac{\tilde{\gamma}(t)}{\left\Vert \w\left(t\right)\right\Vert _{2}}-\frac{\log(N)x_{\max}}{\tilde{\gamma}(t)} \,.
\label{a100}
\end{align}
Next we set $t=T_{\alpha}$ and take the limit $\alpha\rightarrow\infty$. Note that $\tilde{\gamma}(T_{\alpha})\overset{\alpha\rightarrow\infty}{\rightarrow}\infty$ since $\epsilon(T_{\alpha})\overset{\alpha\rightarrow\infty}{\rightarrow}0$, and thus the right term in eq. \eqref{a100} is vanishing.
Using eq. \eqref{a1} we get
\begin{align*}
\lim_{\alpha\rightarrow\infty}\frac{1}{\gamma_{2}\left(T_{\alpha}\right)}\leq&\lim_{\alpha\rightarrow\infty}\frac{\left\Vert \w\left(T_{\alpha}\right)\right\Vert _{2}}{\tilde{\gamma}\left(T_{\alpha}\right)}\\
\leq&\lim_{\alpha\rightarrow\infty}\Bigg[\frac{x_{\max}}{\gamma_{2}^{2}}\left[\frac{2\alpha^{2}\gamma_{2}^{2}}{\bar{x}\tilde{\gamma}\left(T_{\alpha}\right)}\sinh\left(\frac{\bar{x}}{2\alpha^{2}\gamma_{2}^{2}}\tilde{\gamma}\left(T_{\alpha}\right)\right)-1\right]+\frac{8\alpha^{2}}{\tilde{\gamma}\left(T_{\alpha}\right)}\int_{0}^{T_{\alpha}}\left\Vert X\r\left(\tau\right)\right\Vert _{2}d\tau\Bigg]\,.
\end{align*}
 We use $\frac{\tilde{\gamma}\left(T_{\alpha}\right)}{\alpha^{2}}\overset{\alpha\rightarrow\infty}{\rightarrow}0$, $\lim_{z\rightarrow0}\frac{\sinh z}{z}=1$ and eq. (\ref{eq: gamma2})
to get
\[
\lim_{\alpha\rightarrow\infty}\frac{1}{\gamma_{2}\left(T_{\alpha}\right)}\leq\frac{1}{\gamma_{2}}\,.
\]
 It follows that $\lim_{\alpha\rightarrow\infty}\gamma_{2}\left(T_{\alpha}\right)=\gamma_{2}$.
\qed

\subsection{Intermediate Regime Proof}
\label{q_alpha_proof}

\begin{theorem}[\thmref{thm:Q_alpha_D} for $D=2$]\label{thm:Q_alpha_D_app_D2}
Under \condref{assmp:intermediate_depthD_app}, for $D=2$ if
$\lim\limits_{\alpha\rightarrow\infty}\frac{\alpha^{2}}{\tilde{\gamma}(\alpha)}=\mu>0$,
then
\[
\hat{\w}=\argmin_{\w}Q_{\mu}^{2}\left(\w\right)\;\mathrm{ s.t. }\,\,\forall n,\,\x_{n}^{\top}\w\geq1
\]
where $Q_{\mu}^{2}\left(\w\right)=\sum_{i=1}^{d}q_{2}\left(\frac{w_{i}}{\mu}\right)$
and $q_{2}\left(s\right)=2-\sqrt{4+s^{2}}+s\cdot\arcsinh\left(\frac{s}{2}\right)$
.
\end{theorem}

\proof
We show that the KKT conditions hold in the limit $\alpha\rightarrow\infty$.
The KKT conditions are that there exists $\boldsymbol{\nu}\in\mathbb{R}_{\geq0}^{N}$ such that
\begin{align}
\nabla Q_{\mu}^2\left(\hat{\w}\right)  &=X\boldsymbol{\nu}\label{eq:stationarity}\\
\forall n:\,\x_{n}^{\top}\hat{\w}  &\geq1\label{eq: primal feasability}\\
\forall n:\,\nu_{n}\left(\x_{n}^{\top}\hat{\w}-1\right)  &=0.\label{eq: dual slackness}
\end{align}

\paragraph{Primal feasibility \eqref{eq: primal feasability}:}
The condition (\ref{eq: primal feasability})
follows by definition of $\hat{\w}$,
\begin{align}
\forall n:\,\,\x_{n}^{\top}\hat{\w}=\lim_{\alpha\rightarrow\infty}\frac{\x_{n}^{\top}\w\left(T_\alpha\right)}{\gamma\left(T_\alpha\right)}\geq\lim_{\alpha\rightarrow\infty}\frac{\min_n\left(\x_{n}^{\top}\w\left(T_\alpha\right)\right)}{\gamma\left(T_\alpha\right)}=1~.
\label{primal_proof}
\end{align}
\paragraph{Stationarity condition \eqref{eq:stationarity}:}
To show the condition (\ref{eq:stationarity}) let
\begin{equation}
\boldsymbol{\nu}=\frac{4}{\mu}\limsup_{\alpha\rightarrow\infty}\int_{0}^{T_{\alpha}}\r(s)ds\in\mathbb{R}_{\geq0}^{N}\,.\label{eq: nu}
\end{equation}
We need to show that
\[
\nabla Q_{\mu}\left(\hat{\w}\right)=\frac{1}{\mu}\arcsinh\left(\frac{\hat{\w}}{2\mu}\right)=X\boldsymbol{\nu}\,.
\]
Indeed from eqs. \eqref{w_sol} and \eqref{ratio1} we have
\begin{align*}
\hat{\w} & =\lim_{\alpha\rightarrow\infty}\frac{2\alpha^{2}\sinh\left(4X\int_{0}^{T_{\alpha}}\r(s)ds\right)}{\gamma\left(T_\alpha\right)}\\
 & =2\lim_{\alpha\rightarrow\infty}\frac{\tilde{\gamma}(T_{\alpha})}{\gamma\left(T_\alpha\right)}\lim_{\alpha\rightarrow\infty}\frac{\alpha^{2}}{\tilde{\gamma}\left(T_\alpha\right)}\limsup_{\alpha\rightarrow\infty}\sinh\left(4X\int_{0}^{T_{\alpha}}\r(s)ds\right)\\
 & =2\mu\sinh\left[\mu X\left(\frac{4}{\mu}\limsup_{\alpha\rightarrow\infty}\int_{0}^{T_{\alpha}}\r(s)ds\right)\right]\\
 & =2\mu\sinh\left(\mu X\boldsymbol{\nu}\right)
\end{align*}
and thus $\frac{1}{\mu}\arcsinh\left(\frac{\hat{\w}}{2\mu}\right)=X\boldsymbol{\nu}$,
as desired.\\
\paragraph{Complementary slackness \eqref{eq: dual slackness}:}
To show the condition (\ref{eq: dual slackness}) let $k\in\left[N\right]$
such that
\begin{equation}
\x_{k}^{\top}\hat{\w}>1\,.\label{eq: super-margin convergence}
\end{equation}
We need to show that $\nu_{k}=0$. We change variables $t\rightarrow\tilde{\gamma}\left(t\right)$
and using eq. (\ref{eq:gamma_tilda_bound1}) we get
\begin{equation}
\int_{0}^{T_{\alpha}}\exp\left(-\x_{k}^{\top}\w\left(s\right)\right)ds\leq\frac{1}{8\alpha^{2}\gamma_{2}^{2}}\int_{0}^{\tilde{\gamma}\left(T_\alpha\right)}\exp\left(-\x_{k}^{\top}\w\left(\tilde{\gamma}\right)+\tilde{\gamma}\right)d\tilde{\gamma}\label{eq:nu_bound}
~.
\end{equation}

From \condref{assmp:intermediate_depthD_app} we know that there exists $\tilde{\gamma}^{\star}\left(\alpha\right)=o\left(\alpha^{2}\right)$
and $\rho_{0}>1$ such that for large enough $\alpha$ and $\tilde{\gamma}\in\left[\tilde{\gamma}^{\star}\left(\alpha\right),\tilde{\gamma}\left(\alpha\right)\right]$,
$\frac{\mathbf{x}_{k}^{\top}\mathbf{w}\left(\tilde{\gamma}\right)}{\gamma\left(\tilde{\gamma}\right)}\ge\rho_{0}$.
Let
\[
\tilde{\gamma}_{1}^{\star}\left(\alpha\right)=\max\left(\frac{2\rho_{0}\log N}{\rho_{0}-1},\tilde{\gamma}^{\star}\left(\alpha\right)\right)=o\left(\alpha^{2}\right)
\]
 and $\rho_{1}=\frac{\rho_{0}+1}{2}>1$. Then for large enough $\alpha$
and $\tilde{\gamma}\in\left[\tilde{\gamma}_{1}^{\star}\left(\alpha\right),\tilde{\gamma}\left(\alpha\right)\right]$,
using $\tilde{\gamma}\leq\gamma+\log N$ we get
\begin{align*}
\frac{\mathbf{x}_{k}^{\top}\mathbf{w}\left(\tilde{\gamma}\right)}{\tilde{\gamma}} & =\frac{\mathbf{x}_{k}^{\top}\mathbf{w}\left(\tilde{\gamma}\right)}{\gamma}\frac{\gamma}{\tilde{\gamma}}\\
 & \geq\rho_{0}\frac{\tilde{\gamma}-\log N}{\tilde{\gamma}}\\
 & =\rho_{0}-\rho_{0}\frac{\log N}{\tilde{\gamma}}\\
 & \geq\rho_{0}-\rho_{0}\frac{\log N}{\frac{2\rho_{0}\log N}{\rho_{0}-1}}\\
 & =\rho_{1}\,.
\end{align*}

Next we decompose the RHS of eq. (\ref{eq:nu_bound})
as following
\begin{align}
\frac{1}{8\alpha^{2}\gamma_{2}^{2}}\int_{0}^{\tilde{\gamma}\left(T_\alpha\right)}\exp\left(-\x_{k}^{\top}\w\left(\tilde{\gamma}\right)+\tilde{\gamma}\right)d\tilde{\gamma}= & \frac{1}{8\alpha^{2}\gamma_{2}^{2}}\int_{0}^{\tilde{\gamma}^{\star}_1\left(\alpha\right)}\exp\left(-\x_{k}^{\top}\w\left(\tilde{\gamma}\right)+\tilde{\gamma}\right)d\tilde{\gamma}\nonumber \\
 & +\frac{1}{8\alpha^{2}\gamma_{2}^{2}}\int_{\tilde{\gamma}^{\star}_1\left(\alpha\right)}^{\tilde{\gamma}\left(T_\alpha\right)}\exp\left(-\x_{k}^{\top}\w\left(\tilde{\gamma}\right)+\tilde{\gamma}\right)d\tilde{\gamma}\nonumber \\
 & =(I)+(II)
 \label{eq: decomposition}
\end{align}
From eq. \eqref{tilda} we have that $\exp\left(-\x_{k}^{\top}\w\left(\tilde{\gamma}\right)+\tilde{\gamma}\right)\leq N$
and thus
\begin{equation}
(I)\leq\frac{1}{8\alpha^{2}\gamma_{2}^{2}}\int_{0}^{\tilde{\gamma}^{\star}_1\left(\alpha\right)}Nd\tilde{\gamma}=\frac{N\tilde{\gamma}^{\star}_1\left(\alpha\right)}{8\alpha^{2}\gamma_{2}^{2}}\overset{\alpha\rightarrow\infty}{\rightarrow}0\label{eq: part1}
\end{equation}
since $\tilde{\gamma}^{\star}_1(\alpha)=o(\alpha^2)$.
For the second term in eq. (\ref{eq: decomposition}) we have for large enough $\alpha$,
\begin{align}
(II) & =\frac{1}{8\alpha^{2}\gamma_{2}^{2}}\int_{\tilde{\gamma}^{\star}_1\left(\alpha\right)}^{\tilde{\gamma}\left(T_\alpha\right)}\exp\left[-\left(\x_{k}^{\top}\frac{\w\left(\tilde{\gamma}\right)}{\tilde{\gamma}}-1\right)\tilde{\gamma}\right]d\tilde{\gamma}\nonumber \\
 & \leq\frac{1}{8\alpha^{2}\gamma_{2}^{2}}\int_{\tilde{\gamma}^{\star}_1\left(\alpha\right)}^{\tilde{\gamma}\left(T_\alpha\right)}\exp\left[-\left(\rho_{1}-1\right)\tilde{\gamma}\right]d\tilde{\gamma}\nonumber \\
 & \leq\frac{1}{8\alpha^{2}\gamma_{2}^{2}}\int_{0}^{\infty}\exp\left[-\left(\rho_{1}-1\right)\tilde{\gamma}\right]d\tilde{\gamma}\nonumber \\
 & =\frac{1}{8\alpha^{2}\gamma_{2}^{2}\left(\rho_{1}-1\right)}\overset{\alpha\rightarrow\infty}{\rightarrow}0\,.\label{eq: part2}
\end{align}
 By substituting eqs. (\ref{eq: part1}) and (\ref{eq: part2}) in
eq. (\ref{eq: decomposition}) we get that $\nu_{k}=0$.
\qed

\subsection{Rich Regime Proof}
\label{L1_limit_2_proof}

\begin{theorem}[\thmref{thm:l1_limit_D} for $D=2$]\label{thm:l1_limit_D_app_D2}
Under \condref{assmp:intermediate_depthD_app_richD2}, for $D=2$ if
$\tilde{\gamma}(\alpha)=\omega(\alpha^2)$
then
\begin{align}
\hat{\w}=\argmin_{\w}\left\Vert \w\right\Vert _{1}\,\,\,\,\mathrm{s.t.}\,\,\forall n:\,\x_{n}^{\top}\w\geq1\,.
\label{l1_problem_D2}
\end{align}
\end{theorem}

\proof
We show that the KKT conditions for the $\ell_1$ max-margin problem \eqref{l1_problem_D2} hold in the limit $\alpha\rightarrow\infty$.
The KKT conditions are that there exists $\boldsymbol{\nu}^{\left(\ell_1\right)}\in\mathbb{R}_{\geq0}^{N}$ such that
\begin{align}
X\boldsymbol{\nu}^{\left(\ell_1\right)}&\in\partial^{\circ}\left\Vert \hat{\w}\right\Vert _{1}\label{eq:stationarity_l1}\\
\forall n:\,\x_{n}^{\top}\hat{\w}  &\geq1\label{eq: primal feasability_l1}\\
\forall n:\,\nu_{n}^{\left(\ell_1\right)}\left(\x_{n}^{\top}\hat{\w}-1\right)  &=0.\label{eq: dual slackness_l1}
\end{align}
To this end let
\begin{equation}
\boldsymbol{\nu}^{\left(\ell_{1}\right)}=\limsup_{\alpha\rightarrow\infty}\frac{4}{\log\frac{\tilde{\gamma}\left(T_\alpha\right)}{\alpha^{2}}}\int_{0}^{T_{\alpha}}\r\left(s\right)ds\in\mathbb{R}_{\geq0}^{N}\,.\label{eq: nu_L1}
\end{equation}
The proof for the primal feasibility condition \eqref{eq: primal feasability_l1} appears in eq. \eqref{primal_proof}.
\paragraph{Stationarity condition \eqref{eq:stationarity_l1}:}
\begin{align}
\hat{\w} & =\lim_{\alpha\rightarrow\infty}\frac{2\alpha^{2}\sinh\left(4X\int_{0}^{T_{\alpha}}\r\left(s\right)ds\right)}{\gamma\left(T_\alpha\right)}\nonumber \\
 & \overset{\eqref{ratio1}}{=}\lim_{\alpha\rightarrow\infty}\frac{2\alpha^{2}\sinh\left(\log\frac{\tilde{\gamma}\left(T_\alpha\right)}{\alpha^{2}}\frac{4X}{\log\frac{\tilde{\gamma}\left(T_\alpha\right)}{\alpha^{2}}}\int_{0}^{T_{\alpha}}\r\left(s\right)ds\right)}{\tilde{\gamma}\left(T_\alpha\right)}\nonumber \\
 & =\lim_{\alpha\rightarrow\infty}\frac{2\sinh\left(\log\left(\frac{\tilde{\gamma}\left(T_\alpha\right)}{\alpha^{2}}\right)^{\frac{4X}{\log\frac{\tilde{\gamma}\left(T_\alpha\right)}{\alpha^{2}}}\int_{0}^{T_{\alpha}}\r\left(s\right)ds}\right)}{\frac{\tilde{\gamma}\left(T_\alpha\right)}{\alpha^{2}}}\nonumber \\
 & =\lim_{\alpha\rightarrow\infty}\frac{2\sinh\left(\log\left(g\left(\alpha\right)\right)^{z\left(\alpha\right)}\right)}{g\left(\alpha\right)}\,,\label{eq: w_hat_L1-1}
\end{align}
where we defined
\[
g\left(\alpha\right)=\frac{\tilde{\gamma}\left(T_\alpha\right)}{\alpha^{2}}\in\mathbb{R}
\]
\[
z\left(\alpha\right)=\frac{4X}{\log\frac{\tilde{\gamma}\left(T_\alpha\right)}{\alpha^{2}}}\int_{0}^{T_{\alpha}}\r\left(s\right)ds\in\mathbb{R}^{d}\,.
\]
Note that from $\lim_{\alpha\rightarrow\infty}\frac{\alpha^{2}}{\tilde{\gamma}\left(T_\alpha\right)}=0$
we have $\lim_{\alpha\rightarrow\infty}g\left(\alpha\right)=\infty$
and from (\ref{eq: nu_L1}) we get $\limsup_{\alpha\rightarrow\infty}z\left(\alpha\right)=X\boldsymbol{\nu}^{\left(\ell_{1}\right)}$.
In addition, for some $f>0$ and $a\in\mathbb{R}$:
\[
\frac{2\sinh\left(\log f^{a}\right)}{f}=\frac{f^{a}-\frac{1}{f^{a}}}{f}=f^{a-1}-\frac{1}{f^{a+1}}\,.
\]
Therefore in (\ref{eq: w_hat_L1-1}) we have,
\[
\hat{\w}=\lim_{\alpha\rightarrow\infty}\left(g\left(\alpha\right)^{z\left(\alpha\right)-1}-\frac{1}{g\left(\alpha\right)^{z\left(\alpha\right)+1}}\right)\,.
\]
Next, it is easy to verify that for all $i=1,...,d$:
\[
\hat{w}_{i}>0\Rightarrow\limsup_{\alpha\rightarrow\infty}z_{i}\left(\alpha\right)=1
\]
\[
\hat{w}_{i}<0\Rightarrow\limsup_{\alpha\rightarrow\infty}z_{i}\left(\alpha\right)=-1
\]
\[
\hat{w}_{i}=0\Rightarrow-1\leq\limsup_{\alpha\rightarrow\infty}z_{i}\left(\alpha\right)\leq1
\]
and so $X\boldsymbol{\nu}^{\left(\ell_{1}\right)}\in\partial^{\circ}\left\Vert \hat{\w}\right\Vert _{1}$.

\paragraph{Complementary slackness \eqref{eq: dual slackness_l1}:}
We perform similar steps to the proof of the intermediate regime in Appendix \ref{q_alpha_proof}.
We change variables $t\rightarrow\tilde{\gamma}\left(t\right)$ and use the weaker \condref{assmp:intermediate_depthD_app_richD2}, where we replace $\tilde{\gamma}^{\star}(\alpha)$ with $\tilde{\gamma}^{\star}_1(\alpha)$ and $\rho_0$ with $\rho_1$ (see the proof of the intermediate regime in Appendix \ref{q_alpha_proof}). We get that
\begin{align}
\frac{4}{N\log\frac{\tilde{\gamma}\left(T_\alpha\right)}{\alpha^{2}}}\int_{0}^{T_{\alpha}}\exp\left(-\x_{k}^{\top}\w\left(s\right)\right)ds\le & \frac{1}{2N\alpha^{2}\gamma_{2}^{2}\log\frac{\tilde{\gamma}\left(T_\alpha\right)}{\alpha^{2}}}\int_{0}^{\tilde{\gamma}\left(T_\alpha\right)}\exp\left(-\x_{k}^{\top}\w\left(\tilde{\gamma}\right)+\tilde{\gamma}\right)d\tilde{\gamma}\nonumber \\
\leq & \frac{1}{2N\alpha^{2}\gamma_{2}^{2}\log\frac{\tilde{\gamma}\left(T_\alpha\right)}{\alpha^{2}}}\int_{0}^{\tilde{\gamma}^{\star}_1\left(\alpha\right)}\exp\left(-\x_{k}^{\top}\w\left(\tilde{\gamma}\right)+\tilde{\gamma}\right)d\tilde{\gamma}\nonumber \\
 & +\frac{1}{2N\alpha^{2}\gamma_{2}^{2}\log\frac{\tilde{\gamma}\left(T_\alpha\right)}{\alpha^{2}}}\int_{\tilde{\gamma}^{\star}_1\left(\alpha\right)}^{\tilde{\gamma}\left(T_\alpha\right)}\exp\left(-\x_{k}^{\top}\w\left(\tilde{\gamma}\right)+\tilde{\gamma}\right)d\tilde{\gamma}\nonumber \\
= & (I)+(II)\label{eq: L1_decom}
\end{align}
 Using $\tilde{\gamma}^{\star}_1\left(\alpha\right)=o\left(\alpha^{2}\log\frac{\tilde{\gamma}\left(T_\alpha\right)}{\alpha^{2}}\right)$
we bound the first term similarly to eq. \eqref{eq: part1}:
\begin{equation}
(I)\leq\frac{\tilde{\gamma}^{\star}_1\left(\alpha\right)}{2N\alpha^{2}\gamma_{2}^{2}\log\frac{\tilde{\gamma}\left(T_\alpha\right)}{\alpha^{2}}}\overset{\alpha\rightarrow\infty}{\rightarrow}0\,.\label{eq: y1}
\end{equation}
 The second term is bounded similarly to eq. \eqref{eq: part2}:
\begin{equation}
(II)\leq\frac{1}{2N\alpha^{2}\gamma_{2}^{2}\log\frac{\tilde{\gamma}\left(T_\alpha\right)}{\alpha^{2}}\left(\rho_{1}-1\right)}\overset{\alpha\rightarrow\infty}{\rightarrow}0\,.\label{eq: y2}
\end{equation}
By substituting eqs. (\ref{eq: y1}) and (\ref{eq: y2}) in eq. (\ref{eq: L1_decom})
we get that $\nu_{k}^{\left(\ell_{1}\right)}=0$.
\qed

\section{Proofs for $D>2$}
\label{proofs_D}

\subsection{Kernel Regime Proof}
\label{l2_limit_proof_D}

\begin{theorem}[\thmref{thm:l2_limit} for $D>2$]
\label{thm:l2_limit_app_D}
For $D>2$, if $\tilde{\gamma}(\alpha)=o(\alpha^D)$
then
\[
\hat{\w}=\argmin_{\w}\left\Vert \w\right\Vert _{2}\,\,\,\,\mathrm{s.t.}\,\,\forall n:\,\x_{n}^{\top}\w\geq1\,.
\]
\end{theorem}


The proof is similar in spirit to the proof for the case $D=2$ (see Appendix \ref{l2_limit_proof}).
\proof
We show convergence of the $\ell_{2}$ margin $\gamma_{2}\left(T_{\alpha}\right)=\frac{\min_{n}\left(\x_{n}^{\top}\w\left(T_{\alpha}\right)\right)}{\left\Vert \w\left(T_{\alpha}\right)\right\Vert _{2}}$
to the max margin $\gamma_{2}$ as $\alpha\rightarrow\infty$.
We have that
\begin{align}
\gamma_{2}\left(t\right)=\frac{\min_{n}\left(\x_{n}^{\top}\w\left(t\right)\right)}{\left\Vert \w\left(t\right)\right\Vert _{2}}\geq\frac{\tilde{\gamma}\left(t\right)-\log(N)}{\left\Vert \w\left(t\right)\right\Vert _{2}}\,.
\label{l2_margin_D}
\end{align}
\paragraph{Lower bound on $\tilde{\gamma}\left(t\right)$:}
Combining eqs. (\ref{eq:gamma_tilda_bound-1}) and (\ref{eq:xr_bound})
we get
\[
\frac{d\tilde{\gamma}\left(t\right)}{dt}\geq2D^{2}\alpha^{2D-2}\gamma_{2}\left\Vert X\r\left(t\right)\right\Vert _{2}
\]
\begin{equation}
\Rightarrow\tilde{\gamma}\left(t\right)\geq2D^{2}\alpha^{2D-2}\gamma_{2}\int_{0}^{t}\left\Vert X\r\left(\tau\right)\right\Vert _{2}d\tau\,.\label{eq: gamma2-1}
\end{equation}
\paragraph{Upper bound on $\left\Vert \w\left(t\right)\right\Vert _{2}$:}
We decompose $\dot{\w}\left(t\right)$ to two terms:
\begin{align*}
\dot{\w}\left(t\right) & =\alpha^{2D-2}D\left(D-2\right)h_{D}^{'}\left(h_{D}^{-1}\left(\frac{\w\left(t\right)}{\alpha^{D}}\right)\right)\circ X\r\left(t\right)\\
 & =\alpha^{2D-2}D^{2}\left[\frac{D-2}{D}h_{D}^{'}\left(h_{D}^{-1}\left(\frac{\w\left(t\right)}{\alpha^{D}}\right)\right)-2\cdot\mathbf{1}\right]\circ X\r\left(t\right)+2D^{2}\alpha^{2D-2}X\r\left(t\right)
\end{align*}
\[
\Rightarrow\left\Vert \dot{\w}\left(t\right)\right\Vert _{2}\leq\alpha^{2D-2}D^{2}\left\Vert \left[\frac{D-2}{D}h_{D}^{'}\left(h_{D}^{-1}\left(\frac{\w\left(t\right)}{\alpha^{D}}\right)\right)-2\cdot\mathbf{1}\right]\circ X\r\left(t\right)\right\Vert _{2}+2D^{2}\alpha^{2D-2}\left\Vert X\r\left(t\right)\right\Vert _{2}
\]
\begin{align}
\Rightarrow\left\Vert \w\left(t\right)\right\Vert _{2}\leq & \alpha^{2D-2}D^{2}\int_{0}^{t}\left\Vert \left[\frac{D-2}{D}h_{D}^{'}\left(h_{D}^{-1}\left(\frac{\w\left(\tau\right)}{\alpha^{D}}\right)\right)-2\cdot\mathbf{1}\right]\circ X\r\left(\tau\right)\right\Vert _{2}d\tau\nonumber \\
 & +2D^{2}\alpha^{2D-2}\int_{0}^{t}\left\Vert X\r\left(\tau\right)\right\Vert _{2}d\tau\,.\label{eq: w_b-1}
\end{align}
 Let $v\left(t\right)=\left\Vert \left[\frac{D-2}{D}h_{D}^{'}\left(h_{D}^{-1}\left(\frac{\w\left(t\right)}{\alpha^{D}}\right)\right)-2\cdot\mathbf{1}\right]\circ X\r\left(t\right)\right\Vert _{2}$.
Then
\begin{align*}
v\left(t\right) & \leq\left\Vert \frac{D-2}{D}h_{D}^{'}\left(h_{D}^{-1}\left(\frac{\w\left(t\right)}{\alpha^{D}}\right)\right)-2\cdot\mathbf{1}\right\Vert _{\infty}x_{\max}\left\Vert \r\left(t\right)\right\Vert _{1}\\
 & =\left\Vert \frac{D-2}{D}h_{D}^{'}\left(h_{D}^{-1}\left(\frac{\w\left(t\right)}{\alpha^{D}}\right)\right)-2\cdot\mathbf{1}\right\Vert _{\infty}x_{\max}\exp\left(-\tilde{\gamma}\left(t\right)\right)\,.
\end{align*}
 Using \lemref{lem:w_bound_D} we get
\[
v\left(t\right)\leq\left[\frac{D-2}{D}h_{D}^{'}\left(\frac{\left(D-2\right)\bar{x}}{2D\gamma_{2}^{2}\alpha^{D}}\tilde{\gamma}\left(t\right)\right)-2\right]x_{\max}\exp\left(-\tilde{\gamma}\left(t\right)\right)\,.
\]
We are interested in bounding $\int_{0}^{t}v\left(\tau\right)d\tau$.
We change variables $t\rightarrow\tilde{\gamma}\left(t\right)$ and
proceed using eq. (\ref{eq: gama_bound_D}),
\begin{align}
\int_{0}^{t}v\left(\tau\right)d\tau & \leq\int_{0}^{\tilde{\gamma}\left(t\right)}\left[\frac{D-2}{D}h_{D}^{'}\left(\frac{\left(D-2\right)\bar{x}}{2D\gamma_{2}^{2}\alpha^{D}}\tilde{\gamma}\right)-2\right]\frac{x_{\max}\exp\left(-\tilde{\gamma}\right)}{2\alpha^{2D-2}D^{2}\gamma_{2}^{2}\exp\left(-\tilde{\gamma}\right)}d\tilde{\gamma}\nonumber \\
 & =\frac{x_{\max}}{2\alpha^{2D-2}D^{2}\gamma_{2}^{2}}\int_{0}^{\tilde{\gamma}\left(t\right)}\left[\frac{D-2}{D}h_{D}^{'}\left(\frac{\left(D-2\right)\bar{x}}{2D\gamma_{2}^{2}\alpha^{D}}\tilde{\gamma}\right)-2\right]d\tilde{\gamma}\nonumber \\
 & =\frac{x_{\max}}{2\alpha^{2D-2}D^{2}\gamma_{2}^{2}}\left[\frac{2\gamma_{2}^{2}\alpha^{D}}{\bar{x}}h_{D}\left(\frac{\left(D-2\right)\bar{x}}{2D\gamma_{2}^{2}\alpha^{D}}\tilde{\gamma}\left(t\right)\right)-2\tilde{\gamma}\left(t\right)\right]~.\label{eq: int_v_D}
\end{align}
Plugging eq. (\ref{eq: int_v_D})
in eq. (\ref{eq: w_b-1}) we get
\begin{align}
\left\Vert \w\left(t\right)\right\Vert _{2}\leq\frac{x_{\max}}{2\gamma_{2}^{2}}\left[\frac{2\gamma_{2}^{2}\alpha^{D}}{\bar{x}}h_{D}\left(\frac{\left(D-2\right)\bar{x}}{2D\gamma_{2}^{2}\alpha^{D}}\tilde{\gamma}\left(t\right)\right)-2\tilde{\gamma}\left(t\right)\right]+2D^{2}\alpha^{2D-2}\int_{0}^{t}\left\Vert X\r\left(\tau\right)\right\Vert _{2}d\tau\,.
\label{a200}
\end{align}
\paragraph{Putting things together:}
From eqs. \eqref{l2_margin_D} and \eqref{eq: w_lower_bound} we have
\begin{align}
\gamma_{2}\left(t\right)\geq\frac{\tilde{\gamma}\left(t\right)-\log(N)}{\left\Vert \w\left(t\right)\right\Vert _{2}}\geq \frac{\tilde{\gamma}(t)}{\left\Vert \w\left(t\right)\right\Vert _{2}}-\frac{\log(N)x_{\max}}{\tilde{\gamma}(t)} \,.
\label{a101}
\end{align}
Next we set $t=T_{\alpha}$ and take the limit $\alpha\rightarrow\infty$. Note that $\tilde{\gamma}(T_{\alpha})\overset{\alpha\rightarrow\infty}{\rightarrow}\infty$ since $\epsilon(T_{\alpha})\overset{\alpha\rightarrow\infty}{\rightarrow}0$, and thus the right term in eq. \eqref{a101} is vanishing.
Using eq. \eqref{a200} we get
\begin{align*}
\lim_{\alpha\rightarrow\infty}\frac{1}{\gamma_{2}\left(T_{\alpha}\right)}\leq&\lim_{\alpha\rightarrow\infty}\frac{\left\Vert \w\left(T_{\alpha}\right)\right\Vert _{2}}{\tilde{\gamma}\left(T_{\alpha}\right)}\\
\leq&\lim_{\alpha\rightarrow\infty}\Bigg[\frac{x_{\max}}{2\gamma_{2}^{2}}\left[\frac{2\gamma_{2}^{2}\alpha^{D}}{\bar{x}\tilde{\gamma}\left(T_{\alpha}\right)}h_{D}\left(\frac{\left(D-2\right)\bar{x}}{2D\gamma_{2}^{2}\alpha^{D}}\tilde{\gamma}\left(T_{\alpha}\right)\right)-2\right]+\frac{2D^{2}\alpha^{2D-2}}{\tilde{\gamma}\left(T_{\alpha}\right)}\int_{0}^{T_{\alpha}}\left\Vert X\r\left(\tau\right)\right\Vert _{2}d\tau\Bigg]\,.
\end{align*}
 We use $\frac{\tilde{\gamma}\left(T_{\alpha}\right)}{\alpha^{2}}\overset{\alpha\rightarrow\infty}{\rightarrow}0$, eq. (\ref{eq: h_0}) and eq. (\ref{eq: gamma2-1})
to get
\[
\lim_{\alpha\rightarrow\infty}\frac{1}{\gamma_{2}\left(T_{\alpha}\right)}\leq\frac{1}{\gamma_{2}}\,.
\]
 It follows that $\lim_{\alpha\rightarrow\infty}\gamma_{2}\left(T_{\alpha}\right)=\gamma_{2}$.
\qed

\subsection{Intermediate Regime Proof}
\label{q_alpha_D_proof}

\begin{theorem}[\thmref{thm:Q_alpha_D} for $D>2$]\label{thm:Q_alpha_D_app}
Under \condref{assmp:intermediate_depthD_app}, for $D>2$ if
$\lim\limits_{\alpha\rightarrow\infty}\frac{\alpha^{D}}{\tilde{\gamma}(\alpha)}=\mu>0$,
then
\[
\hat{\w}=\argmin_{\w}Q_{\mu}^{D}\left(\w\right)\;\mathrm{ s.t. }\,\,\forall n,\,\x_{n}^{\top}\w\geq1
\]
where $Q_{\mu}^{D}\left(\w\right)=\sum_{i=1}^{d}q_{D}\left(\frac{w_{i}}{\mu}\right)$
and $q_{D}\left(s\right)=\int_{0}^{s}h_{D}^{-1}\left(z\right)dz$
for $h_{D}\left(z\right)=\left(1-z\right)^{-\frac{D}{D-2}}-\left(1+z\right)^{-\frac{D}{D-2}}$.
\end{theorem}

\proof
The proof is similar in spirit to the proof for the case $D=2$ (see Appendix \ref{q_alpha_proof}).
We show that the KKT conditions hold in the limit $\alpha\rightarrow\infty$.
The KKT conditions are that there exists $\boldsymbol{\nu}\in\mathbb{R}_{\geq0}^{N}$ such that
\begin{align}
\nabla Q_{\mu}^D\left(\hat{\w}\right)  &=X\boldsymbol{\nu}\label{eq:stationarity_D}\\
\forall n:\,\x_{n}^{\top}\hat{\w}  &\geq1\label{eq: primal feasability_D}\\
\forall n:\,\nu_{n}\left(\x_{n}^{\top}\hat{\w}-1\right)  &=0.\label{eq: dual slackness_D}
\end{align}
The proof of primal feasibility (\ref{eq: primal feasability_D}) for $D=2$ applies also here.
\paragraph{Stationarity condition \eqref{eq:stationarity_D}:}
Let
\begin{equation}
\boldsymbol{\nu}=\frac{D\left(D-2\right)}{\mu}\limsup_{\alpha\rightarrow\infty}\left(\alpha^{D-2}\int_{0}^{T_{\alpha}}\r\left(s\right)ds\right)\in\mathbb{R}_{\geq0}^{N}\,.\label{eq: nu-1}
\end{equation}
We need to show that
\[
\nabla Q_{\mu}^{D}\left(\w\right)=\frac{1}{\mu}h_{D}^{-1}\left(\frac{\w}{\mu}\right)=X\boldsymbol{\nu}\,.
\]
Indeed using \lemref{lem:w_D_dynamics} and eq. \eqref{ratio1} we have
\begin{align*}
\hat{\w} & =\lim_{\alpha\rightarrow\infty}\frac{\alpha^{D}h_{D}\left(\alpha^{D-2}D\left(D-2\right)X\int_{0}^{T_{\alpha}}\r\left(s\right)ds\right)}{\gamma\left(T_\alpha\right)}\\
 & =\lim_{\alpha\rightarrow\infty}\frac{\tilde{\gamma}\left(T_\alpha\right)}{\gamma\left(T_\alpha\right)}\lim_{\alpha\rightarrow\infty}\frac{\alpha^{D}}{\tilde{\gamma}\left(T_\alpha\right)}\limsup_{\alpha\rightarrow\infty}h_{D}\left(\alpha^{D-2}D\left(D-2\right)X\int_{0}^{T_{\alpha}}\r\left(s\right)ds\right)\\
 & =\mu h_{D}\left[\mu X\left(\frac{D\left(D-2\right)}{\mu}\limsup_{\alpha\rightarrow\infty}\left(\alpha^{D-2}\int_{0}^{T_{\alpha}}\r\left(s\right)ds\right)\right)\right]\\
 & =\mu h_{D}\left(\mu X\boldsymbol{\nu}\right)
\end{align*}
and thus $\frac{1}{\mu}h_{D}^{-1}\left(\frac{\hat{\w}}{\mu}\right)=X\boldsymbol{\nu}$,
as desired.
\paragraph{Complementary slackness \eqref{eq: dual slackness_D}:}
Let $k\in\left[N\right]$ such that
\begin{equation}
\x_{k}^{\top}\hat{\w}>1\,.\label{eq: super-margin convergence-1}
\end{equation}
We have to show that $\nu_{k}=0$. We change variables $t\rightarrow\tilde{\gamma}\left(t\right)$
and using eq. (\ref{eq: gama_bound_D}) we get
\begin{equation}
\alpha^{D-2}\int_{0}^{T_{\alpha}}\exp\left(-\x_{k}^{\top}\w\left(s\right)\right)ds\leq\frac{1}{2D^{2}\alpha^{D}\gamma_{2}^{2}}\int_{0}^{\tilde{\gamma}\left(T_\alpha\right)}\exp\left(-\x_{k}^{\top}\w\left(\tilde{\gamma}\right)+\tilde{\gamma}\right)d\tilde{\gamma}~.\label{eq:nu_bound-1}
\end{equation}
 Next we decompose the RHS of eq. (\ref{eq:nu_bound-1}) and employ \condref{assmp:intermediate_depthD_app}, where similarly to the case $D=2$ we replace $\tilde{\gamma}^{\star}(\alpha)$ with $\tilde{\gamma}^{\star}_1(\alpha)$ and $\rho_0$ with $\rho_1$ (see the proof of the intermediate regime for $D=2$ in Appendix \ref{q_alpha_proof}). We get that
\begin{align}
\frac{1}{2D^{2}\alpha^{D}\gamma_{2}^{2}}\int_{0}^{\tilde{\gamma}\left(T_\alpha\right)}\exp\left(-\x_{k}^{\top}\w\left(\tilde{\gamma}\right)+\tilde{\gamma}\right)d\tilde{\gamma}= & \frac{1}{2D^{2}\alpha^{D}\gamma_{2}^{2}}\int_{0}^{\tilde{\gamma}^{\star}_1\left(\alpha\right)}\exp\left(-\x_{k}^{\top}\w\left(\tilde{\gamma}\right)+\tilde{\gamma}\right)d\tilde{\gamma}\nonumber \\
 & +\frac{1}{2D^{2}\alpha^{D}\gamma_{2}^{2}}\int_{\tilde{\gamma}^{\star}_1\left(\alpha\right)}^{\tilde{\gamma}\left(T_\alpha\right)}\exp\left(-\x_{k}^{\top}\w\left(\tilde{\gamma}\right)+\tilde{\gamma}\right)d\tilde{\gamma}\nonumber \\
 & =(I)+(II)\label{eq: decomposition-1}
\end{align}
From eq. \eqref{tilda} we have that
$\exp\left(-\x_{k}^{\top}\w\left(\tilde{\gamma}\right)+\tilde{\gamma}\right)\leq N$
and thus
\begin{equation}
(I)\leq\frac{1}{2D^{2}\alpha^{D}\gamma_{2}^{2}}\int_{0}^{\tilde{\gamma}^{\star}_1\left(\alpha\right)}Nd\tilde{\gamma}=\frac{N\tilde{\gamma}^{\star}_1\left(\alpha\right)}{2D^{2}\alpha^{D}\gamma_{2}^{2}}\overset{\alpha\rightarrow\infty}{\rightarrow}0\label{eq: part1-1}
\end{equation}
since $\tilde{\gamma}^{\star}_1\left(\alpha\right)=o(\alpha^D)$.
For the second term in eq. (\ref{eq: decomposition-1}) we get for large enough $\alpha$,
\begin{align}
(II) & =\frac{1}{2D^{2}\alpha^{D}\gamma_{2}^{2}}\int_{\tilde{\gamma}^{\star}_1\left(\alpha\right)}^{\tilde{\gamma}\left(T_\alpha\right)}\exp\left[-\left(\x_{k}^{\top}\frac{\w\left(\tilde{\gamma}\right)}{\tilde{\gamma}}-1\right)\tilde{\gamma}\right]d\tilde{\gamma}\nonumber \\
 & \leq\frac{1}{2D^{2}\alpha^{D}\gamma_{2}^{2}}\int_{\tilde{\gamma}^{\star}_1\left(\alpha\right)}^{\tilde{\gamma}\left(T_\alpha\right)}\exp\left[-\left(\rho_{1}-1\right)\tilde{\gamma}\right]d\tilde{\gamma}\nonumber \\
 & \leq\frac{1}{2D^{2}\alpha^{D}\gamma_{2}^{2}}\int_{0}^{\infty}\exp\left[-\left(\rho_{1}-1\right)\tilde{\gamma}\right]d\tilde{\gamma}\nonumber \\
 & =\frac{1}{2D^{2}\alpha^{D}\gamma_{2}^{2}\left(\rho_{1}-1\right)}\overset{\alpha\rightarrow\infty}{\rightarrow}0\,.\label{eq: part2-1}
\end{align}
 By substituting eqs. (\ref{eq: part1-1}) and (\ref{eq: part2-1})
in eq. (\ref{eq: decomposition-1}) and back in eq. \eqref{eq:nu_bound-1} we get that $\nu_{k}=0$.
\qed

\subsection{Rich Regime Proof}
\label{L1_limit_D_proof}

\begin{theorem}[\thmref{thm:l1_limit_D} for $D>2$]\label{thm:l1_limit_D_app}
Under \condref{assmp:intermediate_depthD_app}, for $D>2$ if
$\tilde{\gamma}(\alpha)=\omega(\alpha^D)$
then
\begin{align}
\hat{\w}=\argmin_{\w}\left\Vert \w\right\Vert _{1}\,\,\,\,\mathrm{s.t.}\,\,\forall n:\,\x_{n}^{\top}\w\geq1\,.
\label{l1_problem}
\end{align}
\end{theorem}

\proof
We show that the KKT conditions \eqref{eq:stationarity_l1}, \eqref{eq: primal feasability_l1}, \eqref{eq: dual slackness_l1} for the $\ell_1$ max-margin problem \eqref{l1_problem} hold in the limit $\alpha\rightarrow\infty$.
To this end let
\begin{equation}
\boldsymbol{\nu}^{\left(\ell_1\right)}=D\left(D-2\right)\limsup_{\alpha\rightarrow\infty}\left(\alpha^{D-2}\int_{0}^{T_{\alpha}}\r\left(s\right)ds\right)\in\mathbb{R}_{\geq0}^{N}\,.\label{eq: nu_l1}
\end{equation}
Note that this definition is similar to eq. \eqref{eq: nu-1}, and if $\nu_k=0$ for some $k$ then also $\nu^{\left(\ell_1\right)}_k=0$.
Therefore it is left to show the stationarity condition $X\boldsymbol{\nu}^{\left(\ell_1\right)}\in\partial^{\circ}\left\Vert \hat{\w}\right\Vert _{1}$.
Indeed, from eq. (\ref{eq:domain_bounds}) we know that $-1\leq\left[X\boldsymbol{\nu}^{\left(\ell_{1}\right)}\right]_{i}\leq1$
for all $i$. In addition
\begin{align*}
\hat{\w} & =\lim_{\alpha\rightarrow\infty}\frac{\alpha^{D}h_{D}\left(\alpha^{D-2}D\left(D-2\right)X\int_{0}^{T_{\alpha}}\r\left(s\right)ds\right)}{\gamma\left(T_\alpha\right)}\\
 & \overset{\eqref{ratio1}}{=}\lim_{\alpha\rightarrow\infty}\left(\frac{\alpha^{D}}{\tilde{\gamma}\left(T_\alpha\right)}h_{D}\left(\alpha^{D-2}D\left(D-2\right)X\int_{0}^{T_{\alpha}}\r\left(s\right)ds\right)\right)\,.
\end{align*}
Assume that $\hat{w}_{i}>0$. As $\lim_{\alpha\rightarrow\infty}\left(\frac{\alpha^{D}}{\tilde{\gamma}\left(T_\alpha\right)}\right)=0$
we must have that
\[
\left[\alpha^{D-2}D\left(D-2\right)X\int_{0}^{T_{\alpha}}\r\left(s\right)ds\right]_{i}\overset{\alpha\rightarrow\infty}{\rightarrow}1
\]
 and thus $\left[X\boldsymbol{\nu}^{\left(\ell_{1}\right)}\right]_{i}\overset{\alpha\rightarrow\infty}{\rightarrow}1$.
Similarly, if $\hat{w}_{i}<0$ we get $\left[X\boldsymbol{\nu}^{\left(\ell_{1}\right)}\right]_{i}\overset{\alpha\rightarrow\infty}{\rightarrow}-1.$
It follows that $X\boldsymbol{\nu}^{\left(\ell_{1}\right)}\in\partial^{\circ}\left\Vert \hat{\w}\right\Vert _{1}$.
\qed

\section{Additional Simulation Results and Details}
\label{app:simulations}

\subsection{Optimization trajectories with $\tilde{\gamma}$ indicators}

\begin{figure*}[t!]
	\includegraphics[width=0.99\textwidth]{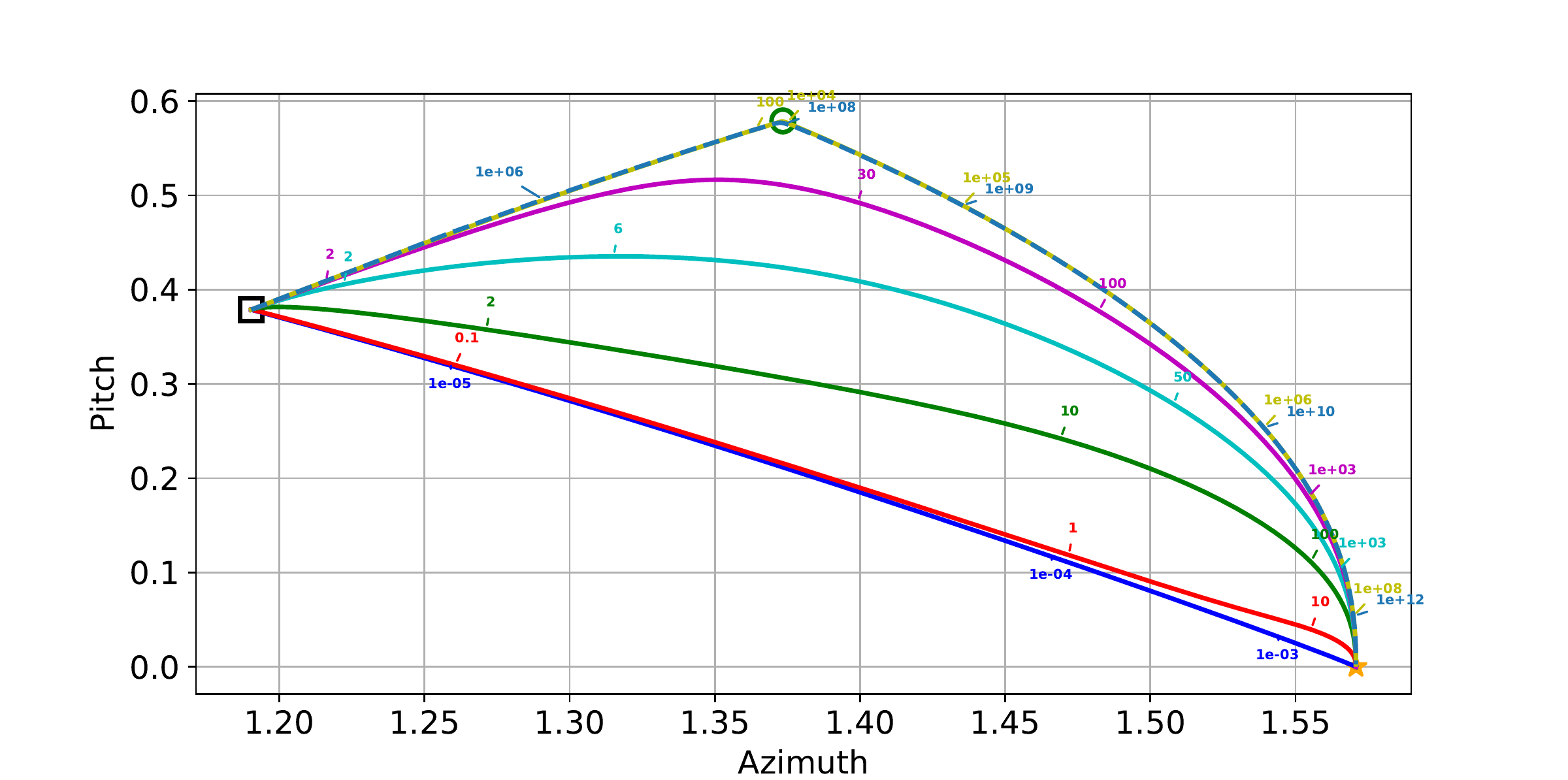}
	\includegraphics[width=0.99\textwidth]{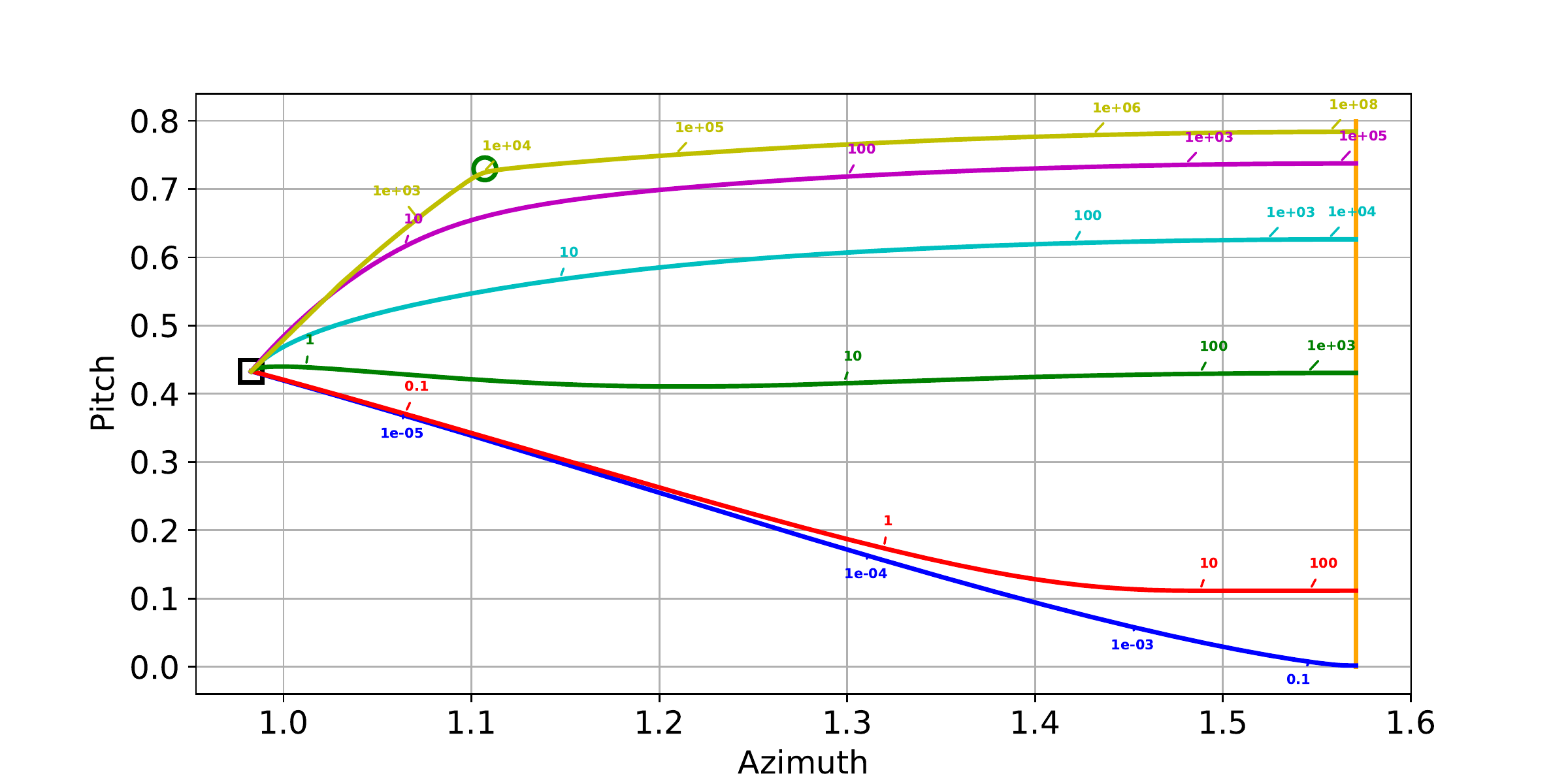}
	\includegraphics[width=0.99\textwidth]{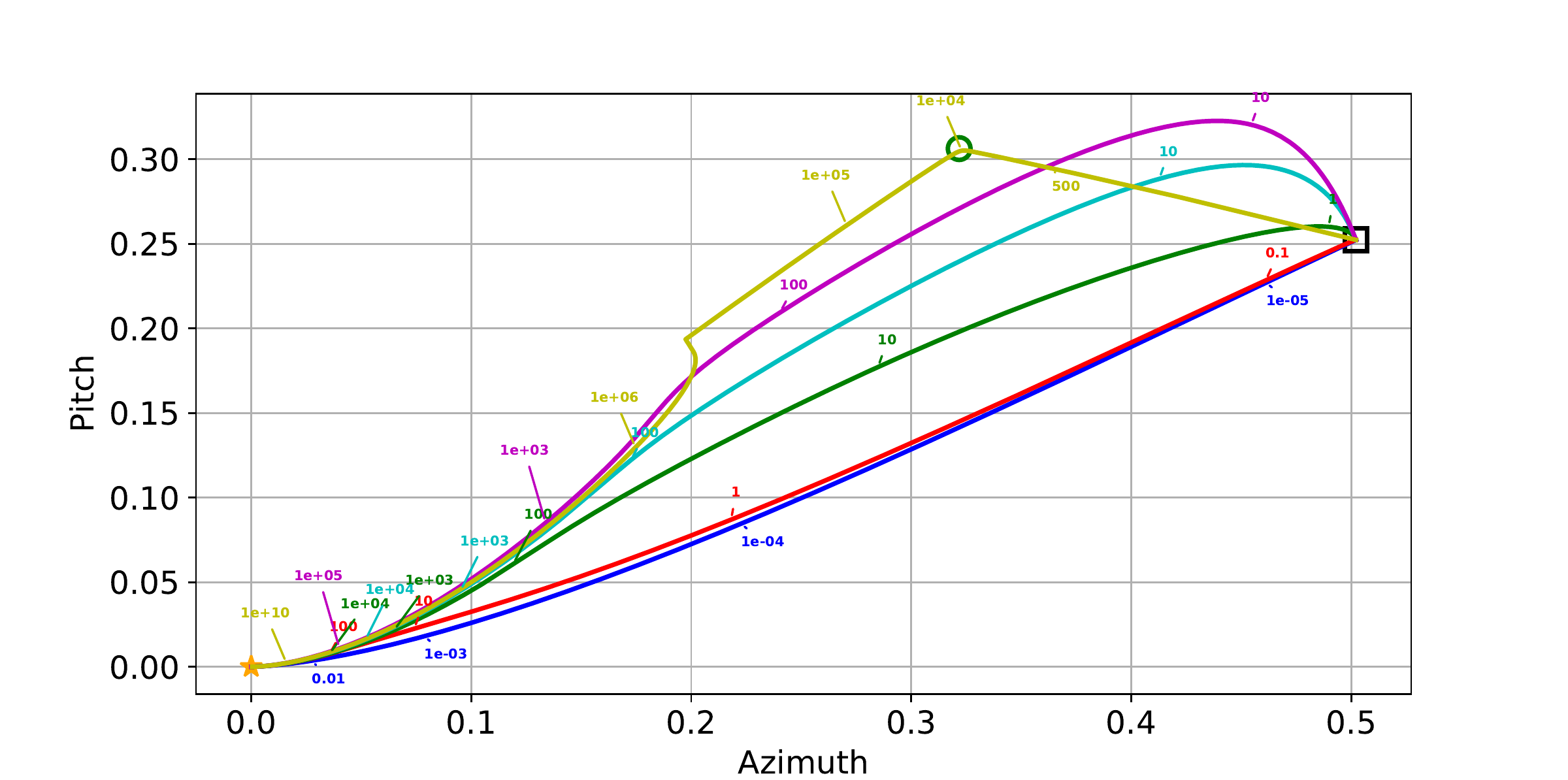}
	\includegraphics[width=0.99\textwidth]{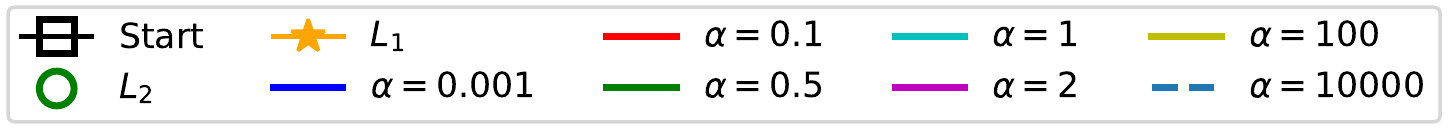}
	\caption{The same optimization trajectories from \figref{fig:depth2_dim3} with $\tilde{\gamma}$ values indications.}
	\label{fig:depth2_with_gamma_values}
\end{figure*}

In \figref{fig:depth2_with_gamma_values} we repeat the optimization trajectories from \figref{fig:depth2_dim3}, but we add indicators that indicate the value of $\tilde{\gamma}$ along the path. Recall that $\tilde{\gamma}(t)=-\log\epsilon(t)$. For example, a number $10$ near some point on the path means that the loss at this point is $\exp(-10)$.

In all three examples we observe that for $\alpha=100$, where the trajectory first visits the $\ell_2$ predictor, around the $\ell_2$ predictor we have $\tilde{\gamma}=10^4=\alpha^2$, as suggested by our theoretical results. In the top figure we also plot the path for $\alpha=10000$, and again around the $\ell_2$ predictor it holds that $\tilde{\gamma}=10^8=\alpha^2$.

In addition we can see that in order to be rather close to $\ell_1$ with large initialization, we need very large $\tilde{\gamma}$, corresponding to extremely small loss $\epsilon$. For example, consider the center plot. For $\alpha=100$ to be close to $\ell_1$ direction we need $\tilde{\gamma}=10^8$, or $\epsilon=\exp(-10^8)$ ! However, with small initialization, e.g. $\alpha=0.001$, $\tilde{\gamma}$ can be as small as $0.1$, or $\epsilon=\exp(-0.1)\approx 0.9$, and we are close to $\ell_1$.

\subsection{Understanding the non-unique $\ell_1$ case}
In \figref{fig:manyL1} we showed an example of optimization trajectories for data with non-unique $\ell_1$ predictor. We can observe that for different initializations the selected $\ell_1$ direction, and thus the implicit bias, is different.

It is interesting to understand what are the properties of different $\ell_1$ solutions. To this end, in \figref{fig:data2_excess} we plot the optimization trajectory in a different way. Instead of looking at the direction of the predictor (as in \figref{fig:manyL1}), we consider the excess $\ell_1$ and $\ell_2$ norms along the path,
defined as $\Vert \w(t)\Vert_1/\Vert \w_{\ell_1}\Vert_1-1$  and $\Vert \w(t)\Vert_2/\Vert \w_{\ell_2}\Vert_2-1$ where $\w_{\ell_1}$ and $\w_{\ell_2}$ are the $\ell_1$ and $\ell_2$ max-margin (minimum norm) solutions accordingly.

We can observe that for large initialization, where we follow the $Q_\mu$ path, the selected $\ell_1$ predictor has the smallest $\ell_2$ norm. Moreover, for small initialization the selected $\ell_1$ predictor has the largest $\ell_2$ norm. Thus, in this case we see an example where the asymptotic (at long time/small loss) implicit bias is affected by the initialization. This in contrast to previous results for exp-tailed losses  (e.g. \cite{soudry2018implicit,ji2019gradient,gunasekar2018implicit,nacson2019lexicographic,lyu2020gradient}), where the asymptotic bias was independent of the initialization.

\subsection{Local minima in high dimension}

\begin{figure*}[t!]
	\includegraphics[width=0.99\textwidth]{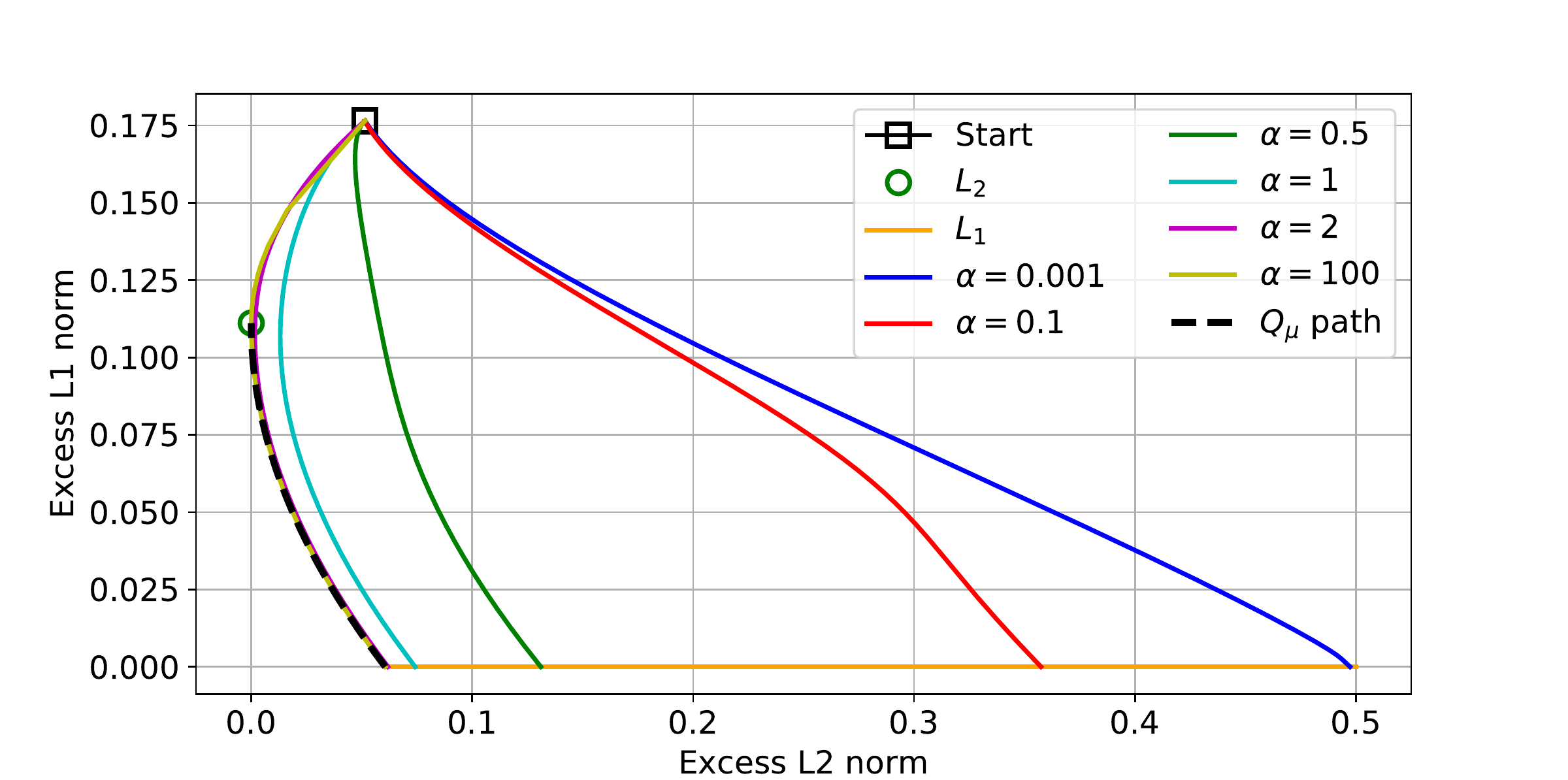}
	\caption{Optimization trajectories for data in \figref{fig:manyL1} in excess $\ell_1$-norm - excess $\ell_2$-norm plane.}
	\label{fig:data2_excess}
\end{figure*}

\begin{figure*}[t!]
	\subfigure[Random data]{\label{fig:dim10_local_min}\includegraphics[width=0.94\textwidth]{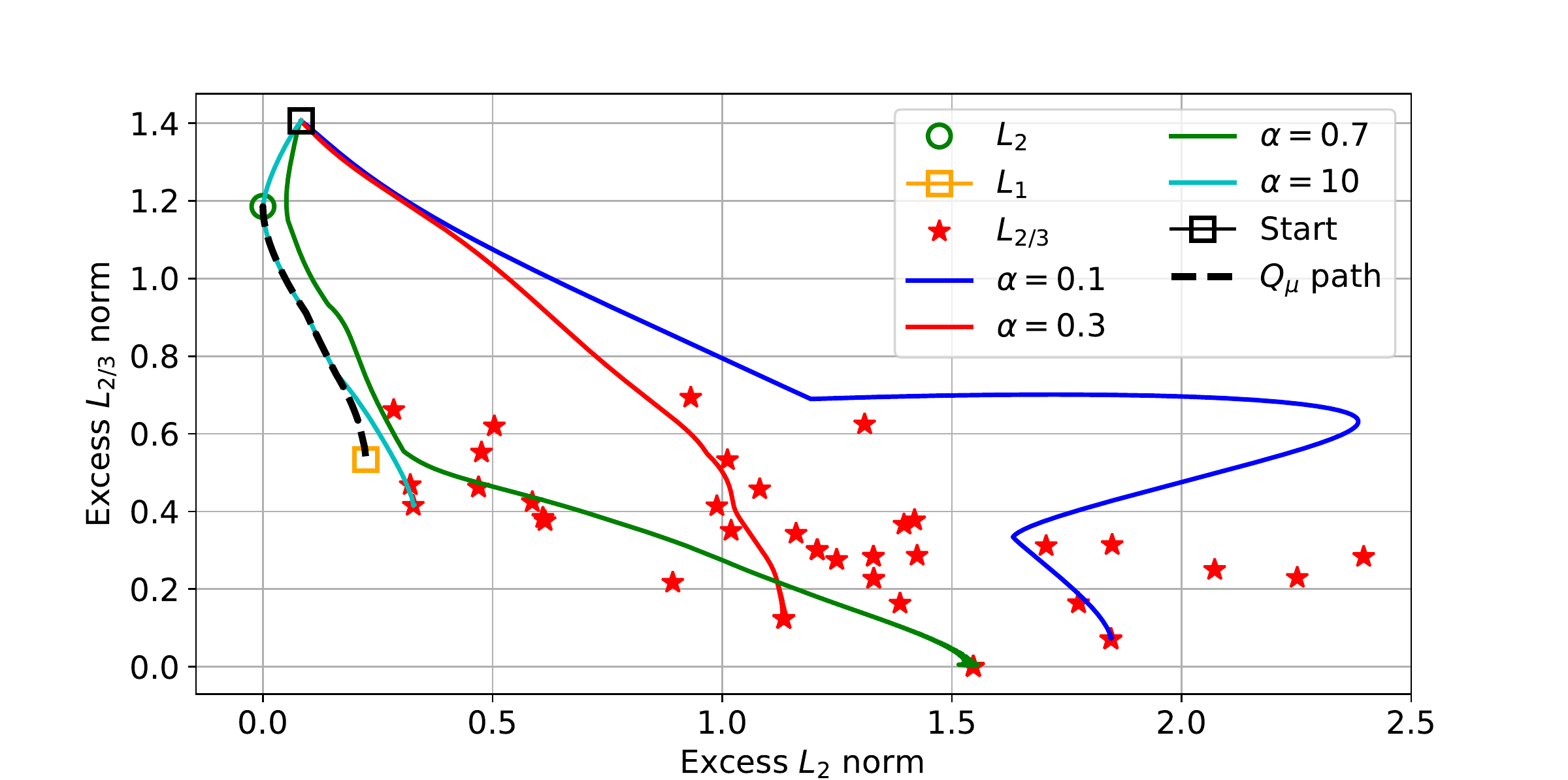}}
	\subfigure[Sparse data]{\label{fig:dim10_sparse_local_min}\includegraphics[width=0.94\textwidth]{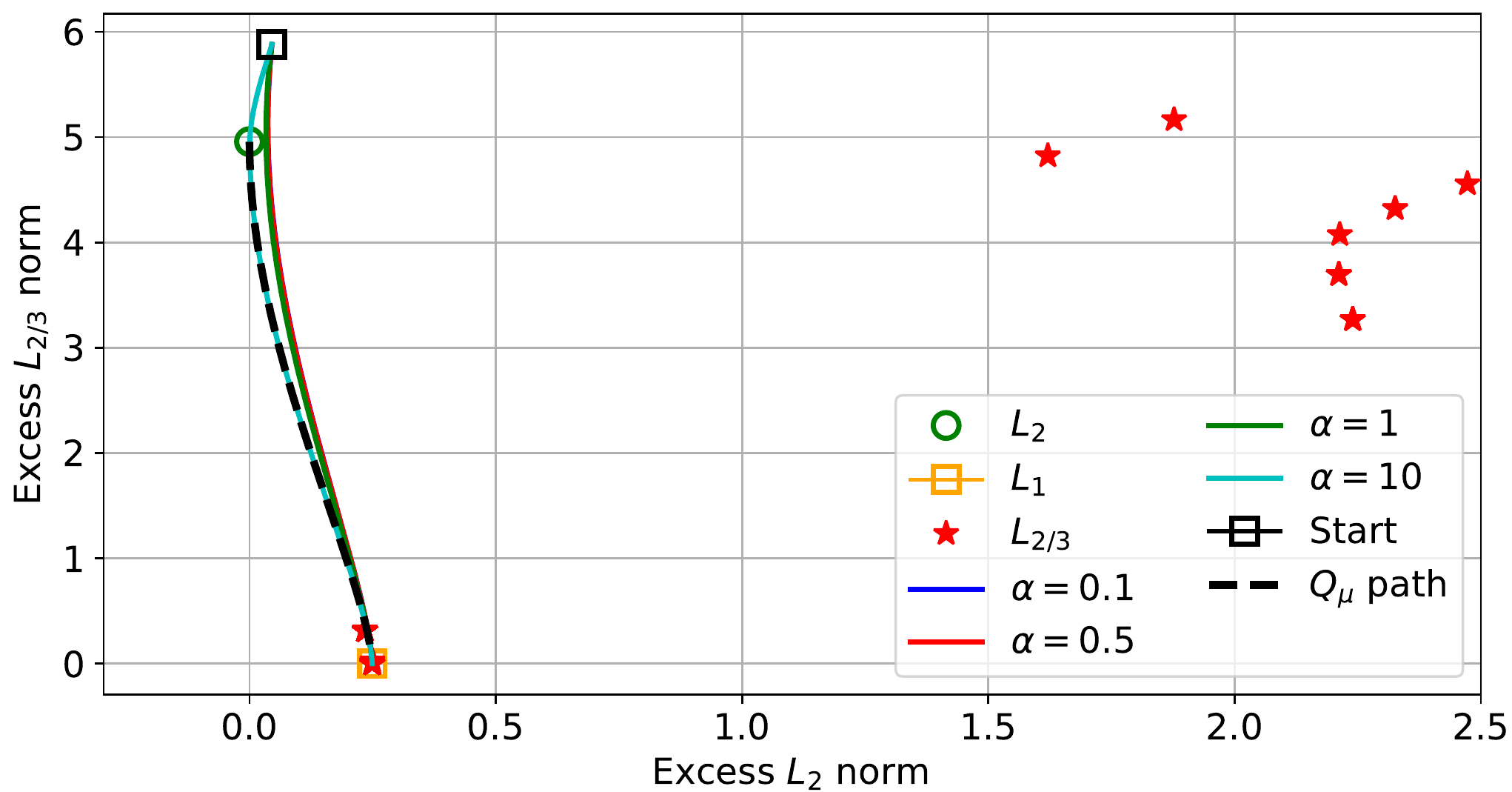}}
	\subfigure[Zoom-in of  \figref{fig:dim10_sparse_local_min}]{\label{fig:dim10_sparse_local_min_zoom}\includegraphics[width=0.94\textwidth]{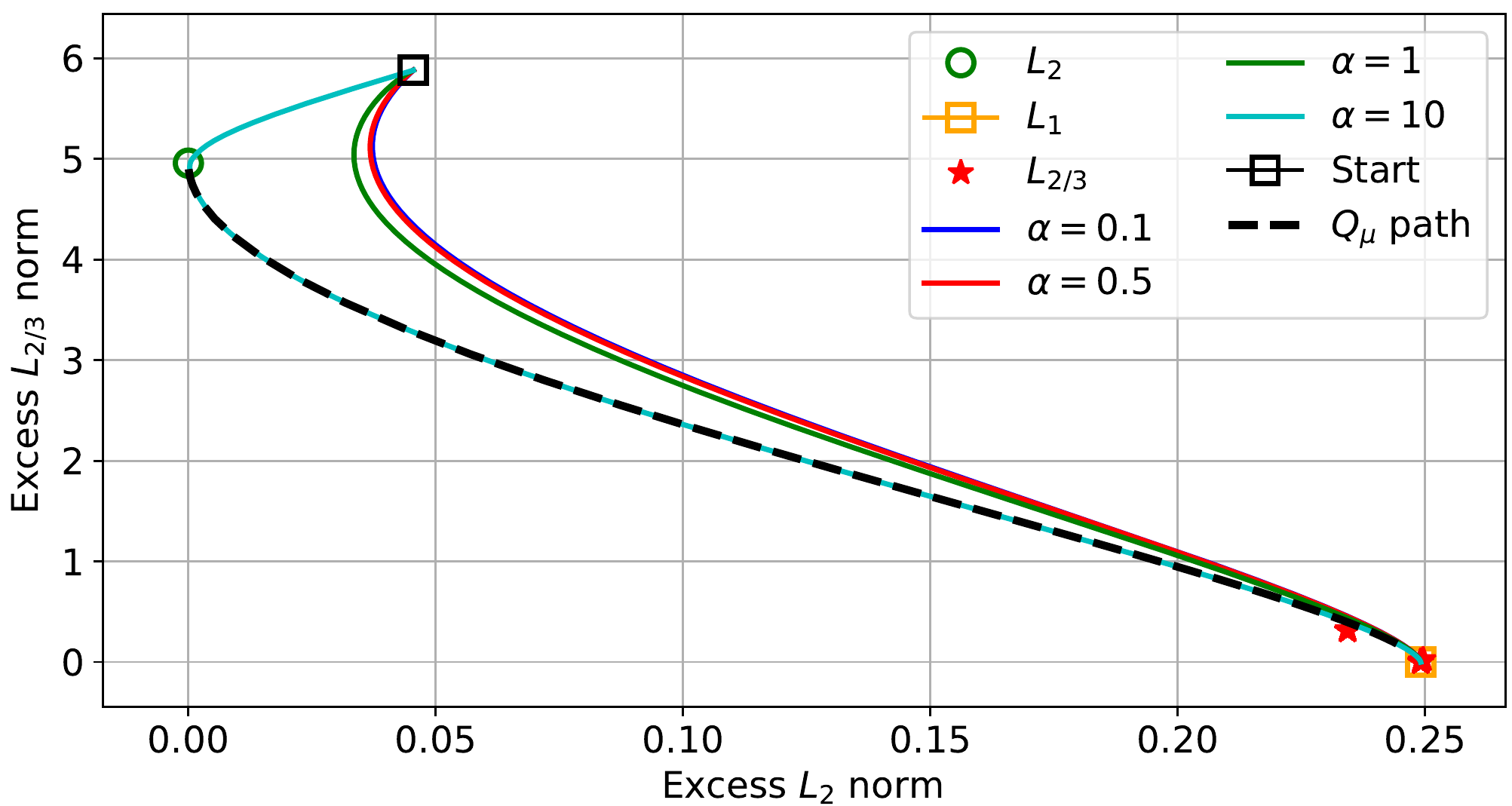}}
	\caption{Optimization trajectories for data in dimension 10.}
	\label{fig:local_minima}
\end{figure*}

In \figref{fig:local_minima} we consider optimization trajectories for data in dimension $10$. In this case we cannot show the direction of the predictor $\frac{\w(t)}{\Vert\w(t)\Vert_2}$ on a sphere, as we did for data in dimension $3$. Instead, we take the approach similar to \figref{fig:data2_excess}, where we show the excess margins.

We consider two datasets in dimension $10$. The first is a random, yet separable, data composed of $10$ points where the coordinates are drawn from $\sim \mathcal{U}(0,1)$.
The second dataset is a sparse dataset of $4$ points, where the first coordinate is $1$ and the other $9$ coordinates are random noise $\sim \mathcal{U}(0,0.5)$. This dataset allows a large separation between the $\ell_2$ max-margin and $\ell_1$ max-margin solutions.

We train depth-3 linear diagonal network and plot the optimization trajectories in $\ell_{2/3}$-$\ell_2$ plane. We observe that for the random data (\figref{fig:dim10_local_min}) there are many local minima of the max $\ell_{2/3}$ margin, and depending on initialization we are biased towards different local-minima points. However, with a large initialization we converge to a local point, quite close to the $Q_\mu$ path and $\ell_1$.

For the sparse data (\figref{fig:dim10_sparse_local_min}, and a zoom-in shown in \figref{fig:dim10_sparse_local_min_zoom}) the local minima are quite far away from the paths. Also in this case the $\ell_1$ and $\ell_{2/3}$ max-margin solutions are the same, and with a large initialization we converge to them, along the $Q_\mu$ path.
This seem to suggest that for certain structure data, like sparse data, we tend to converge to the \textit{global} max $\ell_{2/3}$-margin predictor.

\subsection{Tangent kernel during training}

In \figref{fig:D1} we showed how the excess $\ell_1$-norm depends on $\alpha$ and depth $D$, and measured closeness to the rich limit by excess $\ell_1$-norm.
An alternative and complementary approach is to look at the tangent kernel
$K_t(\x,\x')=\innerprod{\nabla_\u f(\u(t),\x)}{\nabla_\u f(\u(t),\x')}$, which is directly related to closeness to the kernel regime.
As discussed in \secref{sec:kernel-rich-background}, the tangent kernel is almost fixed in the kernel regime, yet can change significantly when we exit the kernel regime.

\begin{figure*}[t!]
	\includegraphics[width=0.99\textwidth]{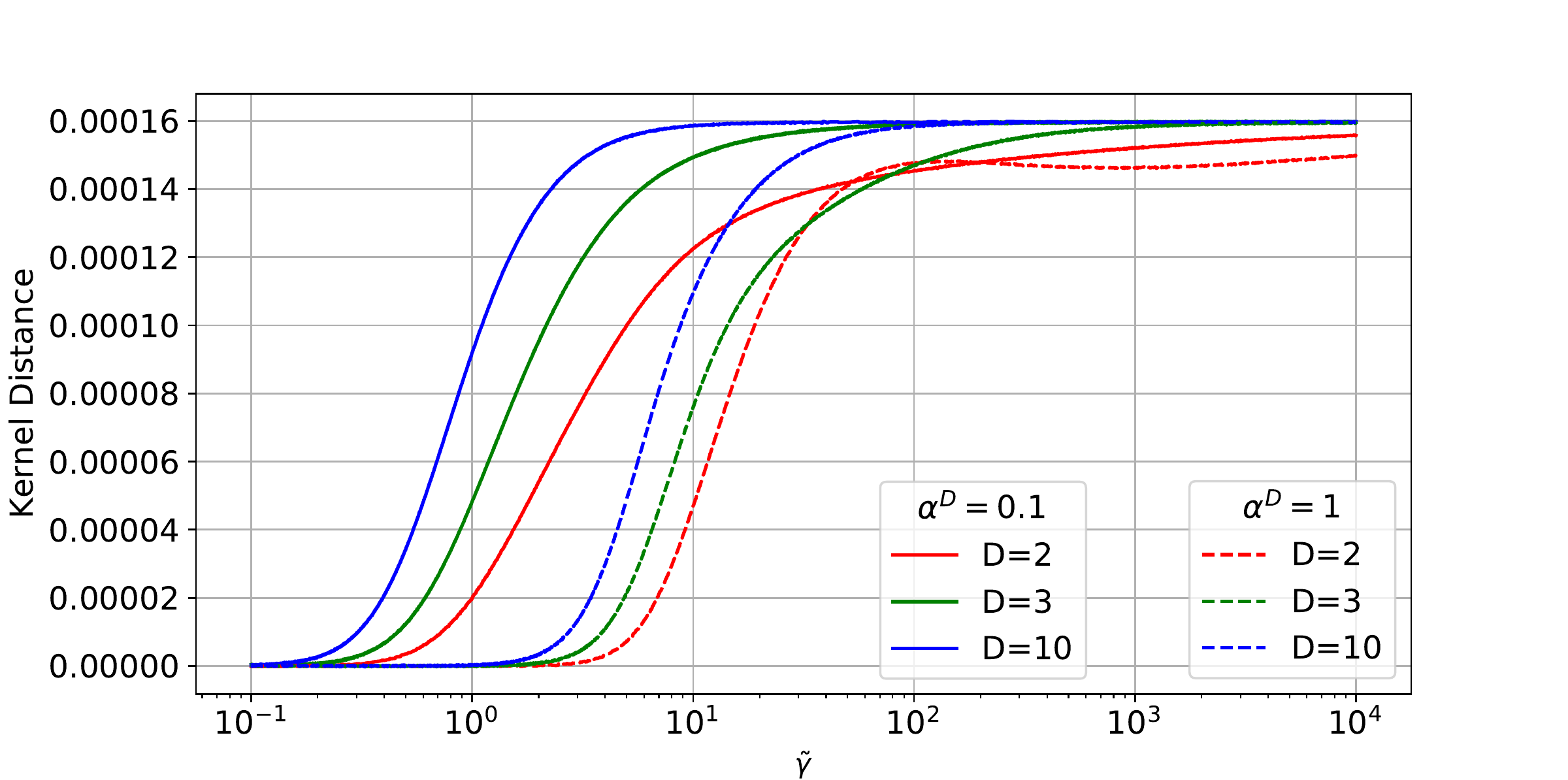}
	\caption{The Kernel distance is defined as $1-\textrm{CosineSimilarity}\left(K(t),K(0)\right)$ where $K(0)$ is the tangent kernel at initialization and $K(t)$ is the tangent kernel at time $t$.}
	\label{fig:kernel_distance}
\end{figure*}

In \figref{fig:kernel_distance} we show the kernel distance during optimization for the same data and network (depth 2) as in \figref{fig:D1}. The kernel distance is defined as $1-\textrm{CosineSimilarity}\left(K(t),K(0)\right)$ where $K(0)\in\mathbb{R}^{N\times N}$ is the tangent kernel at initialization, $K(t)\in\mathbb{R}^{N\times N}$ is the tangent kernel at time $t$, and
\begin{align*}
    \textrm{CosineSimilarity}\left(K(t),K(0)\right) = \frac{\innerprod{K(t)}{K(0)}}{\Vert K(t)\Vert_2 \Vert K(0)\Vert_2}=
    \frac{\textrm{Tr}\left(K(t)K^\top(0)\right)}{\Vert K(t)\Vert_2 \Vert K(0)\Vert_2}~.
\end{align*}
Here we focus on exiting the kernel regime, rather than closeness to the rich regime. We observe that increasing depth will help to exit the kernel regime (where $distance\approx 0$) earlier, at a larger loss value $\epsilon$. Decreasing the initialization has a similar effect, and this is consistent with \figref{fig:D1}.

\subsection{Addressing Numerical Issues}
In our simulations we employ the normalized gradient descent update rule, given by
\[
\u\left(t+1\right)=\u\left(t\right)-\eta\frac{\nabla\mathcal{L}\left(\u\left(t\right)\right)}{\mathcal{L}\left(\u\left(t\right)\right)}
\]
where $\u\in\mathbb{R}^{2d}$ is the vector of parameters and
\[
\mathcal{L}\left(\u\left(t\right)\right)=\frac{1}{N}\sum_{n=1}^{N}\exp\left(-\tilde{\x}_{n}^{\top}\u^{D}\left(t\right)\right)\,.
\]
This algorithm effectively enlarges the learning rate according to
the current loss, and for single layer linear models \citet{DBLP:conf/aistats/NacsonLGSSS19} showed that  the loss decreases exponentially faster.

Let $G\left(\u\left(t\right)\right)=\frac{\nabla\mathcal{L}\left(\u\left(t\right)\right)}{\mathcal{L}\left(\u\left(t\right)\right)}$.
During training the loss can become extremely small, e.g. well beyond $10^{-1000}$,
and in this case also the gradient is very small. This can cause numerical
issues in calculating $G.$ In order to have a numerically stable
evaluation of $G$, and avoid cases like $0/0$, we take an approach
similar to \cite{lyu2020gradient}. Specifically, let
\[
\bar{\gamma}_{n}(t)=\tilde{\x}_{n}^{\top}\u^{D}\left(t\right)\,\,\,\,\,\,,\,\,\,\,\,\,\,\,\,\,\,\gamma(t)=\min_{n}\bar{\gamma}_{n}(t)\,
\]
Then we have that
\[
\nabla\mathcal{L}\left(\u\left(t\right)\right)=-\frac{D}{N}\u^{D-1}\left(t\right)\circ\sum_{n=1}^{N}\exp\left(-\bar{\gamma}_{n}(t)\right)\tilde{\x}_{n}
\]
and
\begin{align}
G\left(\u\left(t\right)\right) & =-D\u^{D-1}\left(t\right)\circ\frac{\sum_{n=1}^{N}\exp\left(-\bar{\gamma}_{n}(t)\right)\tilde{\x}_{n}}{\sum_{n=1}^{N}\exp\left(-\bar{\gamma}_{n}(t)\right)}\nonumber \\
 & =-D\u^{D-1}\left(t\right)\circ\frac{\sum_{n=1}^{N}\exp\left(-\left(\bar{\gamma}_{n}(t)-\gamma(t)\right)\right)\tilde{\x}_{n}}{\sum_{n=1}^{N}\exp\left(-\left(\bar{\gamma}_{n}(t)-\gamma(t)\right)\right)}~.
 \label{eq: G_u_t}
\end{align}
We calculate $G$ according to (\ref{eq: G_u_t}). Note that $\max_{n}\exp\left(-\left(\bar{\gamma}_{n}(t)-\gamma(t)\right)\right)=1$
so the denominator is at least $1$ and the sum in the numerator will
contain at least one support vector $\tilde{\x}_{n:\bar{\gamma}_{n}(t)=\gamma(t)}$.

It is important to note that we never represent the loss values, but only the parameters $\u$. Thus, as long as $\u$ can be represented by \texttt{float64} precision the simulation can continue and we get extremely large parameters corresponding to extremely small loss.

\end{document}